\documentclass{article}

\usepackage[main, final]{neurips_2026}

\usepackage[utf8]{inputenc} %
\usepackage[T1]{fontenc}    %
\usepackage{nicefrac}       %
\usepackage{microtype}      %
\usepackage{xcolor}         %
\usepackage{wrapfig}
\usepackage[]{graphicx}
\usepackage{mypackages}

\usepackage{booktabs}   
\usepackage{multirow}   
\usepackage{pifont}     
\newcommand{\cmark}{\ding{51}}
\newcommand{\xmark}{\ding{55}}

\usepackage{xspace}
\newcommand{\rehg}{\textsc{RE-HG}\xspace}
\newcommand{\oracle}{\textsc{Oracle}\xspace}
\newcommand{\noinf}{\textsc{No-Influence}\xspace}
\newcommand{\rehgsingle}{\textsc{RE-HG(Single)}\xspace}

\title{Reward-Estimated Hypergradient for Bilevel Reinforcement Learning with Black-Box Follower}

\author{%
  Shigeki Kusaka\footnotemark[1]\,\,\footnotemark[2]\,\,\footnotemark[3]
  \And
  Mikoto Kudo\footnotemark[1]\,\,\footnotemark[2] 
  \And
  Takumi Tanabe\footnotemark[4] 
  \AND
  Akifumi Wachi\footnotemark[4]
  \And Youhei Akimoto\footnotemark[1]\,\,\footnotemark[2]\,\,\footnotemark[3]
  \AND 
  \textnormal{
  \footnotemark[1] \ University of Tsukuba \space 
  \footnotemark[2] \ RIKEN AIP \space
  \footnotemark[3] \ Institute of Science Tokyo \space
  \footnotemark[4] \ LY Corporation} \\
  \texttt{\{shigeki.kusaka, mikoto\}@bbo.cs.tsukuba.ac.jp} \\
  \texttt{\{takumi.tanabe, akifumi.wachi\}@lycorp.co.jp} \\
  \texttt{akimoto@cs.tsukuba.ac.jp}
}

\begin{document}

\maketitle

\begin{abstract}

Hypergradient (HG) methods for bilevel reinforcement learning (RL) typically assume a white-box follower whose reward is known. In practice, however, the leader rarely knows the follower’s reward function and can only observe its trajectories. We address this black-box setting and propose Reward-Estimated Hypergradients (RE-HG), which recovers the follower’s reward from these trajectories via inverse RL (IRL). Yet IRL identifies the reward only up to potential-based shaping, a transformation that leaves the follower’s behavior unchanged but biases the HG. We prove that, for linear rewards, this bias cancels when IRL estimates are aggregated across sufficiently diverse dynamics, which the leader’s own updates naturally supply. To handle finite samples, RE-HG truncates small eigenvalues of the aggregated feature covariance, restoring the identifiable subspace and stabilizing the estimation. Experiments show that RE-HG estimates gradients aligned with a white-box Oracle and achieves comparable performance; empirically, the truncation also suppresses HG explosions, letting RE-HG avoid local optima that trap the Oracle in sharp reward landscapes.
\end{abstract}

\section{Introduction}
Reinforcement Learning (RL)~\citep{Sutton2018-jf} has achieved remarkable success in games~\citep{DBLP:journals/nature/SilverHMGSDSAPL16,10.5555/3306127.3332052}, robotics~\citep{10.5555/2946645.2946684}, and the fine-tuning of large language models~\citep{ouyang2022training,guo2025deepseek}.
Many real-world deployments, however, involve multiple agents whose interaction is hierarchical: one agent, the \emph{leader}, shapes the environment in which another agent, the \emph{follower}, learns.
Bilevel RL~\citep{DBLP:journals/siamjo/HongWWY23} models this structure as a nested optimization problem, in which the leader optimizes its objective while anticipating the follower's best response:
\begin{align*}
    &\max_{\theta \in \Theta}\ J_L(\theta, g^{\theta\dagger}) &\text{(Leader)}\\
    &\mathrm{s.t.}\ g^{\theta\dagger} = \argmax_{g \in \mathcal{G}} J_{\theta}(g). &\text{(Follower)}
\end{align*}
Here, $\theta$ parameterizes the leader, $J_L$ is the leader's objective, and $g^{\theta\dagger}$ is the follower's best response, which maximizes the follower's own objective $J_{\theta}$ defined by its reward function.
The standard approach to optimizing the leader is to compute the hypergradient (HG), the gradient of $J_L$ with respect to $\theta$ that accounts for how the follower's best response $g^{\theta\dagger}$ changes with $\theta$.
 
Existing HG methods, however, assume a white-box follower: they require the follower's reward function and, in some cases, its policy or value function (\Cref{tab:assumptions_comparison}).
In practice, this assumption rarely holds: the follower's preferences are often implicit, as with human arm movements~\citep{e918480ee3244d87910100b47661a7c3} or human drivers in mixed traffic~\citep{11105015}, and manually engineering a reward function that reflects them is notoriously difficult.
To the best of our knowledge, no established method computes the HG in this black-box follower setting, where the leader can observe only the follower's trajectories.
 
A natural approach is to recover the follower's reward from its trajectories by inverse RL (IRL) and to compute the HG with the recovered reward.
However, IRL identifies the reward only up to potential-based shaping~\citep{Ng1999icml}: shaped rewards induce the same optimal policy, so the follower's trajectories cannot distinguish between them.
In single-agent RL, this ambiguity is harmless, since shaping adds no bias to the policy gradient.
The HG, in contrast, depends directly on the follower's value function, which shaping shifts by an arbitrary potential, so the ambiguity left by IRL becomes an arbitrary bias in the HG.
 
Our key observation is that the bilevel structure itself supplies the information needed to resolve this ambiguity.
Prior work has shown that a reward can be identified up to a constant when expert behavior is observed under multiple transition dynamics~\citep{Cao2021neurips,Rolland2022neurips}.
In bilevel RL, such observations arise naturally: each leader update changes the dynamics that the follower faces, and the follower responds with a new best response.
We show that, under a linear reward assumption, aggregating the IRL estimates obtained across these iterations cancels the shaping bias and recovers the reward up to a constant offset, which leaves the HG unchanged.
To make this aggregation reliable with finite samples, we truncate the small eigenvalues of the aggregated feature covariance, which restores the identifiable subspace and keeps the estimation numerically stable.
Combining these components, we propose \emph{Reward-Estimated Hypergradient} (\rehg), which computes the HG from observed trajectories alone.
 
The main contributions of this work are as follows:
 
($i$) We identify reward-shaping ambiguity as the fundamental obstacle to HG estimation with a black-box follower.
 
($ii$) We prove that, under a linear reward assumption, aggregating IRL estimates across the dynamics induced by the leader's updates cancels this bias (\Cref{thm:identifiability}), and that the remaining reward error controls the HG error (\Cref{prop:bound}).
 
($iii$) We propose \rehg, which combines this aggregation with eigenvalue truncation to compute the HG from trajectories alone with finite samples.
 
($iv$) Experiments on discrete and continuous bilevel tasks show that \rehg estimates gradients aligned with a white-box \oracle and achieves comparable performance.
Empirically, eigenvalue truncation also suppresses HG explosions, letting \rehg avoid local optima that trap the \oracle in sharp reward landscapes.

Our code is available at \url{https://github.com/akimotolab/RE-HG}.

\section{Problem Formulation}
 
\paragraph{Entropy Regularized MDPs}
In this work, we assume that a follower solves a Markov decision process (MDP) with entropy regularization.
An MDP is defined by the tuple $\langle\mathcal{S}, \mathcal{B}, p, \rho_0, r_F, \gamma_F\rangle$, where $\mathcal{S}$ represents the state space, $\mathcal{B}$ denotes the follower's action space, and $p: \mathcal{S}\times\mathcal{B}\times\mathcal{S}\to[0,1]$ is the state transition probability.
The initial state distribution is given by $\rho_0: \mathcal{S}\to[0,1]$, the follower's reward function is $r_F:\mathcal{S}\times\mathcal{B}\to\mathbb{R}$, and $\gamma_F \in [0,1)$ is the discount factor.
 
The follower aims to identify a stochastic policy $g \in \mathcal{G}: \mathcal{S} \times \mathcal{B} \to [0,1]$ that maximizes the expected discounted return plus an entropy bonus:
\begin{equation}
\max_{g \in \mathcal{G}} J(g) :=\mathbb{E}_{\tau}\left[ \sum_{t=0}^\infty \gamma_F^t \left( r_F(s_t, b_t) + \beta H(g;s_t) \right)\right],\label{eq:lower}
\end{equation}
where $\mathcal{G}$ is the set of all valid policies on $\mathcal{S} \times \mathcal{B}$, and $\beta > 0$ determines the regularization strength.
The term $H(g; s) := -\mathbb{E}_{b \sim g(\cdot \mid s)}[\log g(b \mid s)]$ represents the entropy of policy $g$ at state $s \in \mathcal{S}$.
The expectation is taken over trajectories $\tau := (s_0, b_0, s_1, b_1, \dots)$ distributed according to $p(\tau \mid g) = \rho_0(s_0) \prod_{t=0}^{\infty} p(s_{t+1} \mid s_{t}, b_{t}) g(b_{t} \mid s_{t})$.
 
The primary motivation for employing entropy regularization is to ensure that the optimal policy is both unique and stochastic. This framework is essential in inverse reinforcement learning (IRL), where it serves as a natural mechanism for modeling the inherent stochasticity of expert behavior.
Under this setting, we define the ``soft'' value function $V_F^g$ and action-value function $Q_F^g$ as:
\begin{equation*}
V_F^{g}(s) := \mathbb{E}_\tau \left[ \sum_{t=0}^\infty \gamma_F^t \left( r_F(s_t, b_t) + \beta H(g; s_t) \right) \mid s_0 = s \right]\end{equation*}
and $Q_F^{g}(s,b) := r_F(s, b) + \gamma_F \mathbb{E}_{s' \sim p(\cdot \mid s, b)}[ V_F^{g}(s') ]$.
Given a sufficiently expressive policy class, the unique optimal policy $g^\star$ is characterized by the Boltzmann distribution relative to the optimal value functions $V_F^\star(s) = \max_{g\in \mathcal{G}} V_F^{g}(s)$ and $Q_F^\star(s,b) = \max_{g\in \mathcal{G}} Q_F^{g}(s, b)$ \citep{softQ}:
\begin{equation*}g^\star(b \mid s) = \exp \left( \beta^{-1} \left( Q_F^\star(s, b) - V_F^\star(s) \right) \right).
\end{equation*}
This uniqueness is a critical property for the stability and well-definedness of bi-level RL problems.
 
\paragraph{Configurable Markov Decision Process}
A Configurable Markov Decision Process (Conf-MDP) extends the standard MDP by allowing certain environmental dynamics to be tuned via a configuration parameter. We denote a Conf-MDP as $\mathcal{M}_{\theta}$, where $\theta \in \Theta$ parameterizes the transition probability $p^{\theta}$ and the initial state distribution $\rho_0^{\theta}$.
Thus, the environment is defined by the tuple $\langle \mathcal{S}$, $\mathcal{B}$, $p^{\theta}$, $\rho_0^{\theta}$, $r_F$, $\gamma_F \rangle$. For any fixed configuration $\theta \in \Theta$, the follower solves the entropy-regularized objective $J_{\theta}$ defined by \eqref{eq:lower} using the parameterized dynamics $p^{\theta}$ and $\rho_0^{\theta}$.
 
The leader's goal is to optimize their own utility by selecting a configuration $\theta$ that accounts for the follower's subsequent best response.
This results in the following bi-level optimization problem:
\begin{equation}
\max_{\theta \in \Theta} J_L(\theta, g^{\theta\dagger}), \quad \text{where} \quad g^{\theta\dagger} := \argmax_{g \in \mathcal{G}} J_{\theta}(g). \label{eq:upper}
\end{equation}
Here, $J_L$ is the leader's utility and $g^{\theta\dagger}$ is the follower's best response to $\theta$.
Because of the entropy regularization, $g^{\theta\dagger}$ is unique for every $\theta$ and is given by the $\theta$-dependent Boltzmann distribution:
\begin{equation*}
g^{\theta\dagger}(b \mid s) = \exp \left( \frac{Q_F^{\theta\dagger}(s, b) - V_F^{\theta\dagger}(s)}{\beta} \right),
\end{equation*}
where $Q_F^{\theta\dagger}$ and $V_F^{\theta\dagger}$ are the optimal soft value functions.
We assume the follower always reacts with its optimal policy.
The leader's utility is defined as the expected discounted cumulative reward:
\begin{equation}
J_L(\theta, g^{\theta\dagger}) := \mathbb{E}_\tau \left[ \sum_{t=0}^\infty \gamma_L^t r_L^{\theta}(s_t, b_t) \right], \label{eq:leader-utility}
\end{equation}
where $\gamma_L$ is the leader's discount factor and $r_L^{\theta}$ is the leader's reward function.
Because the leader has control over the environment, we assume that the leader knows $p^\theta$ and $\rho_0^\theta$. Moreover, we assume that the gradients $\nabla_{\theta} r_L^{\theta}$, $\nabla_{\theta} \log \rho_0^{\theta}$, and $\nabla_{\theta} \log p^{\theta}$ exists and the leader can compute them explicitly.

\paragraph{2-Player Markov Game}
A 2-player Markov Game (MG) further generalizes the Conf-MDP by treating the leader as an active agent.
In this setting, the leader follows a policy $f_{\theta}$ parameterized by $\theta \in \Theta$, indirectly influencing the environment through their actions.
An MG is defined by the tuple $\langle \mathcal{S}, (\mathcal{A}, \mathcal{B}), p, \rho_0, (r_L, r_F), (\gamma_L, \gamma_F)\rangle$, where $\mathcal{A}$ and $\mathcal{B}$ are the action spaces for the leader and follower, respectively. Transitions $p(s' \mid s, a, b)$ and rewards $r_L, r_F$ now depend on both agents' actions.
The policies are $f_{\theta}: \mathcal{S} \times \mathcal{A} \to [0,1]$ and $g: \mathcal{S} \times \mathcal{A} \times \mathcal{B} \to [0,1]$.
 
As in the Conf-MDP, the follower optimizes the entropy-regularized objective (\ref{eq:lower}), but the dynamics are now induced by the leader's policy: $p^{\theta}(s', a' \mid s, a, b) = f_{\theta}(a' \mid s') p(s' \mid s, a, b)$.
The leader's optimization remains consistent with \eqref{eq:upper}, with the utility defined as:
\begin{equation}
J_L(\theta, g^{\theta\dagger}) := \mathbb{E}_\tau \left[ \sum_{t=0}^\infty \gamma_L^t r_L(s_t, a_t, b_t) \right], \label{eq:leader-utility-MG}
\end{equation}
where $\tau = (s_0, a_0, b_0, s_1, \dots)$ consists of state, leader action, and follower action. We define the follower's soft value and action-value functions under a specific leader policy $f_{\theta}$ as:
\begin{equation*}
V_F^{\theta,g}(s, a) := \mathbb{E}_\tau \left[ \sum_{t=0}^\infty \gamma_F^t \left( r_F(s_t, a_t, b_t) + \beta H(g; s_t, a_t) \right) \middle|\begin{array}{l}s_0=s\\a_0=a\end{array} \right]
\end{equation*}
and $Q_F^{\theta,g}(s, a, b) := r_F(s, a, b) + \gamma_F \mathbb{E}_{s' \sim p} \mathbb{E}_{a' \sim f_\theta} [ V_F^{\theta,g}(s', a') ]$. Here, $V_F^{\theta,g}$ takes the leader's action $a$ as an input to reflect the follower's observation of the leader's move.
The leader's action-value function is defined without entropy regularization:
\begin{equation*}Q_L^{\theta,g}(s, a, b) := \mathbb{E}_\tau \left[ \sum_{t=0}^\infty \gamma_L^t r_L(s_t, a_t, b_t)~\middle|~\begin{array}{l}s_0=s\\a_0=a\\b_0=b\end{array} \right].\end{equation*}
For brevity, we use the notation $Q_L^{\theta \dagger}$, $V_F^{\theta \dagger}$, and $Q_F^{\theta \dagger}$ to denote the respective value functions when the follower plays their unique best response $g^{\theta\dagger}$ to the leader's policy $f_{\theta}$.
Unlike Conf-MDP settings, the leader may not know the transition probability. We consider two scenarios: the leader (1) knows $p$ and $\rho_0$ explicitly; and (2) does not know them explicitly, but can simulate the interaction with $p$ and $\rho_0$.

\paragraph{Inverse Reinforcement Learning}
 
Inverse Reinforcement Learning (IRL) is the process of recovering an underlying reward function from an observed policy, assuming that the policy is optimal with respect to that reward. In our framework, we employ the Maximum Discounted Causal Entropy (MDCE) IRL framework \citep{Zhou2018tac} to estimate the follower's reward function based on their best response $g^{\theta\dagger}$ under a given leader policy $f_{\theta}$.
To formalize this, let $J_\theta(g; r)$ represent the follower's objective as defined in \eqref{eq:lower}, but with the true reward $r_F$ replaced by a surrogate reward $r$, and the regularization constant set to $\beta = 1$. This simplification is without loss of generality, as the scaling factor $\beta$ is effectively absorbed into the surrogate reward; specifically, maximizing $J_\theta(g; r_F / \beta)$ with $\beta=1$ is mathematically equivalent to the original formulation in \eqref{eq:lower}. We denote the solution to the forward reinforcement learning problem under a surrogate reward $r$ as:
$\textsc{RL}_\theta(r) = \argmax_g J_\theta(g; r)$.
Given the follower's best response $g^{\theta\dagger}$ as the ``expert'' policy, the IRL problem is formulated as finding a reward function $r$ that minimizes the difference between the maximum achievable value and the value attained by the expert:
\begin{equation}
\textsc{IRL}_\theta(g^{\theta\dagger}) = \argmin_{r} \left( \max_{g} J_\theta(g; r) - \mathbb{E}_{\tau \sim (f_\theta, g^{\theta\dagger})} \left[ \sum_{t=0}^{\infty} \gamma_F^t r(s_t, a_t, b_t) \right] \right), \label{eq:irl}
\end{equation}
where the expectation is taken over trajectories $\tau$ generated by the leader $f_\theta$ and the follower $g^{\theta\dagger}$.
As established in \citet[Corollary 3.2.1]{GAIL} and \citet[Theorem 2]{Syed2008icml}, if a policy $\hat{g}$ is obtained such that $\hat{g} \in \textsc{RL}_\theta \circ \textsc{IRL}_\theta(g^{\theta\dagger})$, then we successfully recover the follower's behavior, i.e., $\hat{g} = g^{\theta\dagger}$. This implies that the reward function $\hat{r}$ resulting from the \textsc{IRL} process effectively characterizes the optimal behavior of the follower under the leader parameter $\theta$.

\section{Related Work}\label{sec:related}
 
\paragraph{Hypergradient for Bilevel RL}

In Bilevel RL, computing the HG to optimize the leader's policy has become a standard approach. While existing methods have evolved significantly in terms of convergence rates and the relaxation of theoretical assumptions, they ultimately rely on the ``white-box'' nature of the follower to varying degrees. For instance, PARL~\citep{DBLP:conf/iclr/ChakrabortyBKWM24} and PBRL~\citep{10.5555/3692070.3693893} assume deterministic updates that require full knowledge of the HG, while HPGD~\citep{DBLP:conf/nips/ThomaP0RH24} and SoBiRL~\citep{DBLP:conf/aistats/YangGY25} are bottlenecked by the requirement for a sampler capable of starting from arbitrary state-action pairs or allowing repeated visits to those pairs. Although BC-HG circumvents the need for repeated visits, it still necessitates direct access to the follower’s policy and value functions.
As summarized in \Cref{tab:assumptions_comparison}, all these approaches, regardless of their estimation technique, rely on explicit knowledge of the follower's true reward function, which \rehg does not require.

\providecommand{\myxmark}{\textcolor{red}{\xmark}}
\begin{table}[t]
    \centering
    \caption{Comparison of requirements for HG estimation methods in Bilevel RL.}
    \label{tab:assumptions_comparison}
    \resizebox{\textwidth}{!}{
    \begin{tabular}{lccccc}
        \toprule
        \multirow{2}{*}{\textbf{Method}} & \textbf{Requires Follower's} & \textbf{Update Method} & \textbf{Requires Hessian}  & \textbf{Requires Repeated} \\
        & \textbf{True Reward} & \textbf{(HG Info)} & \textbf{(2nd Order Deriv.)}  & \textbf{State Visits} \\
        \midrule
        PARL~\citep{DBLP:conf/iclr/ChakrabortyBKWM24} & \cmark & Deterministic & \cmark  & \myxmark \\
        PBRL~\citep{10.5555/3692070.3693893}        & \cmark & Deterministic & \myxmark  & \myxmark \\
        SoBiRL~\citep{DBLP:conf/aistats/YangGY25}      & \cmark & Stochastic & \myxmark  & \cmark \\
        HPGD \citep{DBLP:conf/nips/ThomaP0RH24}       & \cmark & Stochastic    & \myxmark  & \cmark \\
        BC-HG~\citep{Kudo2026icaps}       & \cmark & Stochastic    & \myxmark  & \myxmark \\
        \midrule
        \textbf{RE-HG (Ours)}        & \myxmark & Stochastic & \myxmark & \myxmark \\
        \bottomrule
    \end{tabular}
    }
\end{table}
 
\paragraph{Inverse Reinforcement Learning}
While the optimal policy for a given reward function is uniquely identified under entropy regularization, the reward function leading to the given policy is not uniquely determined. \citet{Ng1999icml} shows that for an arbitrary function $\Phi: \mathcal{S} \to \R$, the shaped reward function $r'(s, b) = r(s, b) + \gamma_F \E_{s' \sim p(\cdot\mid s, b)}[\Phi(s')] - \Phi(s)$ admits the optimal policy equivalent to that for the original reward function $r$.
It has been argued that a shaped reward function is insufficient to obtain the optimal policy when the transition probability used for the policy training is different from the one used for the expert trajectory generation.
Several works propose approaches to recover the original reward function (up to a constant shift, which is unavoidable) under different assumptions.
For example, \citet{AIRL,firl2020corl} assume a state-only reward, along with other conditions on the transition.
\citet{Cao2021neurips,Rolland2022neurips} assume access to the expert trajectories on two different MDPs sharing the same reward function.
We develop a method based on the latter idea. A direct comparison against the joint multi-environment formulation of \citet{Cao2021neurips} is provided in \Cref{apdx:multi-env-irl}.

\section{Approach}
 
We derive a novel estimation of the HG of the leader's objective.
As a baseline, we employ BC-HG~\citep{Kudo2026icaps} for the HG estimate that assumes the knowledge of the optimal value function of the follower.
Our approach, RE-HG, incorporates an inverse reinforcement learning technique to mitigate the necessity of the optimal value function of the follower.
 
\paragraph{Baseline: Hypergradient estimation with Known Follower's Value}
 
If the follower's optimal value function is accessible, the HG of the leader's objective can be computed both for Conf-MDP and 2-player MG settings.
The HG for the Conf-MDP setting is derived as follows.
Let $Q_L^{\theta,g}(s, b)$ represent the leader's expected discounted return given that the follower chooses action $b$ at state $s$:
\begin{equation*}
    Q_L^{{\theta} {g}}(s, b) := \mathbb{E}_\tau \left[ \sum_{t=0}^\infty \gamma_L^t r_L(s_t, b_t)  \middle|\begin{array}{l}s_0=s\\b_0=b\end{array} \right].
\end{equation*}
Distinct from standard action-value functions, this definition treats the follower's action $b$ as the primary input alongside the state $s$, rather than the leader's action. We denote the leader's value function under the follower's optimal response as $Q_L^{\theta \dagger} := Q_L^{\theta,g^{\theta\dagger}}$.
Then, a concept of the \emph{benefit} is introduced. For a given state $s$, the benefit $B_L^{\theta\dagger}(s, b)$ quantifies the relative advantage the leader gains when the follower selects action $b$ compared to the follower's average behavior under $g^{\theta\dagger}$:
\begin{equation*}
B_L^{{\theta}\dagger}(s, b) := Q_L^{{\theta}\dagger}(s,b) - \mathbb{E}_{b' \sim g^{{\theta}\dagger}(\cdot \mid s)} \left[ Q_L^{{\theta}\dagger}(s, b') \right].
\end{equation*}
\citet{Kudo2026icaps} show that the HG of $J_L$ in \eqref{eq:leader-utility} can be computed in the case of Conf-MDPs as
\begin{equation}
    \nabla_{{\theta}}J_L({\theta}, g^{\theta\dagger})
    =  \frac{\partial J_L({\theta}, g^{\theta\dagger})}{\partial \theta} + \mathbb{E}_\tau \Bigg[ \sum_{t=0}^\infty \gamma_L^t \frac{B_L^{{\theta}\dagger}(s_t, b_t)}{\beta} \nabla_{\theta} Q_F^{{\theta} \dagger}(s_t, b_t)  \Bigg],
\label{eq:hg_configurable}
\end{equation}
where $\frac{\partial J_L({\theta}, g^{\theta\dagger})}{\partial \theta}$ is the partial derivative of $J_L(\theta, g)$ with respect to $\theta$ evaluated at $(\theta, g^{\theta\dagger})$, whose explicit form is provided in \Cref{apdx:hg}. The gradient of the follower's soft action-value function is:
\begin{equation}
\nabla_{{\theta}}Q_F^{{\theta}\dagger}(s,b) = \mathbb{E}_\tau \Bigg[ \sum_{t=0}^\infty \gamma_F^{t+1} V_F^{{\theta}\dagger}(s_{t+1}) \nabla_{{\theta}} \log p^{{\theta}}(s_{t+1} \mid s_t, b_t) \Bigg|\begin{array}{l}s_0=s\\b_0=b\end{array} \Bigg].
\label{eq:nablaQF}
\end{equation}
The HG can be computed similarly in 2-player MGs, described in \Cref{apdx:hg}. 
Importantly, in both cases, the term $\frac{B_{L}}{\beta} \nabla_{\theta} Q_{F}$ appears. We refer to it as the \emph{influence term}, as it accounts for the change in the follower's best response induced by the leader's configuration.

\paragraph{Follower's Value Estimation via Repeated Inverse Reinforcement Learning}
 
The challenge in estimating the HG without knowing the follower's reward function is to estimate the value function of the follower, $V_F^{\theta\dag}$.
As mentioned earlier, an IRL returns a reward function $\hat{r} \in \textsc{IRL}_\theta(g^{\theta\dagger})$ that admits the same optimal policy $g^{\theta\dagger}$.
In Conf-MDP settings, it includes any function $r_\Phi(s, b) = r(s, b) + \gamma_F\E[\Phi(s')\mid s, b] - \Phi(s)$ with an arbitrary state-only potential function $\Phi: \mathcal{S} \to \R$ \citep{Ng1999icml}.
The soft value function under the shaped reward $r_\Phi$ is then $V_{\Phi}^{\theta\dagger} = V_F^{\theta\dagger} - \Phi$.
Therefore, if we use a shaped reward to estimate the follower's value function, the term $\nabla_\theta Q_F^{\theta\dagger}$ will be arbitrarily affected by $\Phi$, resulting in biased estimation of the HG $\nabla_\theta J_L(\theta, g^{\theta\dagger})$.
The same argument holds for 2-player MG settings.
In this case, the leader's action is regarded as a part of the state from the follower's perspective. Hence, the estimated reward is shaped as $r_\Phi(s, a, b) = r(s, a, b) + \gamma_F\E[\Phi(s', a')\mid s, a, b] - \Phi(s, a)$ with an arbitrary potential function $\Phi: \mathcal{S} \times \mathcal{A} \to \R$. Consequently, the follower's value function is misestimated as $V_{\Phi}^{\theta\dagger} = V_F^{\theta\dagger} - \Phi$, yielding a biased HG due to the additional term 
$-\mathbb{E}_{\tau} \left[ \sum \gamma_F^{t+1} \Phi(s_{t+1}) \nabla_{\theta} \log p^{\theta}(s_{t+1} \mid s_t, b_t) \right]$ arising from substituting $V_{\Phi}^{\theta\dagger}$ in \eqref{eq:nablaQF}.
Different from the single-agent MDPs, where the reward shaping adds no bias in policy gradient, the shaping term is crucial for HG. 
The defect of the shaping term is demonstrated in \Cref{sec:experiments} as well as in \Cref{demo:syn}.
 
To circumvent this hurdle, we pose the assumption on the follower's reward function.
In what follows, we consider Conf-MDP settings for conciseness.
To treat 2-player MGs, we can simply regard a state $s \in \mathcal{S}$ as a pair $(s, a) \in \mathcal{S} \times \mathcal{A}$ of the state and the leader's action at state $s$, the transition probability as $p(s'\mid s, a, b) f_\theta(a' \mid s')$, and the initial state distribution as $\rho_0(s)f_\theta(a \mid s)$.
\begin{assumption}[Follower's Reward Linearity]\label{asm:linear}
For a given feature extractor $\phi: \mathcal{S} \times \mathcal{B} \to \R^d$, the follower's reward function is given by $r_F(s, b) = \inner{\omega^\star}{\phi(s, b)}$ with $\omega^\star \in \R^d$. Moreover, the features have full rank, i.e., $\textsc{Span}(\{\phi(s, b): (s, b) \in \mathcal{S} \times \mathcal{B}\}) = \R^d$.
\end{assumption}
 
Reward linearity is not sufficient to identify the true reward function up to a constant offset \citep{Cao2021neurips,Rolland2022neurips}. However, the reward parameter obtained by $\textsc{IRL}_\theta(g^{\theta\dagger})$ is in a subspace defined as follows.
\begin{proposition}\label{prop:irl}
Given the best response follower $g^{\theta\dagger}$,
let $\psi^{\theta\dagger}$ be the expected discounted features under the leader $f_{\theta}$ and its best response follower $g^{\theta\dagger}$ defined as
\begin{equation}
\psi^{\theta\dagger}(s, b) = \E_{p^\theta, g^{\theta\dagger}} \left[\sum_{t=0}^\infty \gamma_F^t \phi(s_t, b_t) \Bigg|\begin{array}{l}s_0=s\\b_0=b\end{array}\right].
\end{equation}
Let $\Sigma^{\theta\dagger} = \E_{d_{\gamma_F}^{\theta,g^{\theta\dagger}}}\left[\Cov_{g^{\theta\dagger}}[\psi^{\theta\dagger}(s, b) ]\right]$.
Then, the solution to the IRL lies in the kernel of $\Sigma^{\theta\dagger}$ centered at the ground truth parameter $\omega^\star$, i.e.,
$\textsc{IRL}_\theta(g^{\theta\dagger}) = \{ \omega \in \R^d :  \Sigma^{\theta\dagger} (\omega - \omega^\star) = 0\}$.
[The proof is provided in \Cref{apdx:prop:irl}.]
\end{proposition}
 
Since the follower's transition probability depends on $\theta$ and changes at each outer iteration, we naturally obtain expert policies on different MDPs sharing the same reward, the setting in which \citet{Cao2021neurips,Rolland2022neurips} identify the reward up to a constant offset.
Consequently, we obtain the convergence of the estimated reward parameter up to a constant offset of the reward.
\begin{theorem}\label{thm:identifiability}
Let $\{(\theta_k, g^{\theta_k \dagger})\}_{k=1}^{K}$ be a sequence of leader parameters and their corresponding best-response followers. For each $k \in [K]$, let $\omega_k \in \textsc{IRL}_{\theta_k}(g^{\theta_k \dagger})$ be an optimal reward parameter, and define the expected feature covariance matrix as
$\Sigma_k = \E_{d_{\gamma_F}^{\theta_k, g^{\theta_k \dagger}}}\left[\Cov_{g^{\theta_k \dagger}}[\psi^{\theta_k \dagger}(s, b)]\right]$, and let $\bar{\Sigma}_K = \sum_{k=1}^K \Sigma_k$.
Then, any solution $\bar{\omega}_K$ to $\sum_{k=1}^K \Sigma_k \omega = \sum_{k=1}^K \Sigma_k \omega_k$ satisfies $\bar{\omega}_K \in \bigcap_{k=1}^{K} \textsc{IRL}_{\theta_k}(g^{\theta_k \dagger})$.
Furthermore, let $\Delta_\mathrm{const} \in \R^d$ be a vector satisfying $\inner{\phi(s, b)}{\Delta_\mathrm{const}} = 1$ for all $(s, b) \in \mathcal{S} \times \mathcal{B}$ (if such a vector exists, and $0$ otherwise). If the aggregated covariance matrix satisfies the rank condition $\textsc{Rank}\left( \bar{\Sigma}_K \right) = d - \ind{\Delta_\mathrm{const} \neq 0}$,
then we have $\bar{\omega}_K = \omega^\star + c \Delta_\mathrm{const}$ for some $c \in \R$.
Consequently, the recovered reward function satisfies $r_{\bar{\omega}_K}(s, b) = r_{\omega^\star}(s, b) + c$.
[The proof is provided in \Cref{apdx:thm:identifiability}.]
\end{theorem}
 
To solve the system $\sum_{k=1}^K \Sigma_k \omega = \sum_{k=1}^K \Sigma_k \omega_k$ reliably with finite samples, we standardize the covariance matrix so that the eigenvalues are either $1$ or $0$ by dropping eigenvalues smaller than a threshold $\epsilon$, then solve $\sum_{k=1}^K U_k \omega = \sum_{k=1}^K U_k \omega_k$ with the standardized covariance matrix $U$, whose solution space is the same as the original system. The algorithmic detail is provided in \Cref{alg:bilevel_reward_identification}. 
The resulting reward parameter $\bar{\omega}_K$ is guaranteed to converge to the ground-truth reward parameter up to a shift in the direction of $\Delta_\mathrm{const}$ under mild conditions, shown in \Cref{cor:conv}. 
 
The eigenvalue truncation plays three distinct roles, which we separate here to avoid conflating them.
($i$) \emph{Support for bias cancellation.} With finitely many trajectories, the empirical covariance matrix is generically of full rank, and the directions that form the kernel of the exact $\Sigma_k$ appear as small non-zero eigenvalues. Truncation discards these directions and restores the kernel characterized in \Cref{prop:irl}. It does not alter the identifiability limit itself, which is dictated by \Cref{thm:identifiability}.
($ii$) \emph{Numerical stabilization.} In tasks whose feature distribution is heavily skewed, such as the regulator tasks considered here, truncation is what keeps the aggregated linear system solvable to an acceptable accuracy.
($iii$) \emph{Implicit regularization.} By avoiding division by near-zero eigenvalues, truncation bounds the norm of the recovered reward, which in turn bounds the influence term and prevents the HG norm from exploding.
Roles ($i$) and ($ii$) are what \Cref{alg:bilevel_reward_identification} and \Cref{cor:conv} are designed for; role ($iii$) is an empirical effect that we observe consistently in \Cref{sec:experiments}, and we do not claim it as a theoretical guarantee.
 
\begin{corollary}\label{cor:conv}
Let $\{U_k\}_{k=1}^{K}$ be the sequence defined in \Cref{alg:bilevel_reward_identification}, let $\bar{U}_K = \sum_{k=1}^K U_k$, and let $\widehat{\omega}_k$ be the sequence of estimated reward vectors. 
Suppose that $\widehat{\omega}_k$ are independently distributed, each $\widehat{\omega}_k$ satisfies 
$\E[U_k \widehat{\omega}_k] = U_k \omega_k$ and $\Cov[U_k \widehat{\omega}_k] \preccurlyeq C \cdot I$ for some universal constant $C \geq 0$, and $\textsc{Rank}\left(\sum_{k=1}^{K}U_k \right) = d - \ind{\Delta_\mathrm{const} \neq 0}$ for all $K \geq \bar{K}$ for some $\bar{K} \geq 1$.
Let $W = I - \norm{\Delta_\mathrm{const}}_2^{-2} \Delta_\mathrm{const}\Delta_\mathrm{const}^\top$ (which must be regarded as $W = I$ if $\Delta_\mathrm{const} = 0$), and let $\Delta_\omega^K = W (\bar{\omega}_K - \omega^\star)$.
Then, the output of \Cref{alg:bilevel_reward_identification} satisfies
$\E[\Delta_\omega^K] = 0$ and $\Cov[\Delta_\omega^K] \preccurlyeq C \cdot K \cdot (\bar{U}_K^{-})^2$, where $\bar{U}_K^{-}$ is the Moore-Penrose pseudo-inverse of $\bar{U}_K$. 
In particular, if the greatest eigenvalue of $\bar{U}_K^{-}$ scales in $o(K^{-1/2})$, we have $\lim_{K\to\infty} \Delta_\omega^K  = 0$ in probability, implying that the true reward is recovered up to a constant offset.
[The proof is provided in \Cref{apdx:cor:conv}.]
\end{corollary}
 
\Cref{cor:conv} is a conditional statement about the aggregation step: it treats the sequence $\{U_k\}$ as fixed in advance and assumes the estimates $\widehat{\omega}_k$ to be independent. Neither holds verbatim in \Cref{algo:proposal}, where $\theta_k$ is selected adaptively and $U_k$ and $\widehat{\omega}_k$ are computed from the same finite batch. \Cref{apdx:adaptive} states the assumptions under which the conclusion survives in the adaptive case, and makes explicit which variant of the algorithm the resulting guarantee covers.

The ground-truth reward vector recovery translates to the true HG recovery.
\begin{proposition}\label{prop:bound}
Let $\omega^\star$ be the ground truth reward vector and $\widehat{\omega}$ be the estimated reward vector. 
Let $c = \inner{\widehat{\omega} - \omega^\star}{\Delta_\mathrm{const}}/\norm{\Delta_\mathrm{const}}_2^2$ and 
$\Delta_\omega = \widehat{\omega} - \omega^\star - c \Delta_\mathrm{const}$. 
Then, the optimal soft-value functions $V^\star$ and $\widehat{V}$ corresponding to $\omega^\star$ and $\widehat{\omega}$ satisfy $\norm{\widehat{V} - V^\star - c(1 - \gamma_F)^{-1}}_{\infty} \leq L(1 - \gamma_F)^{-1} \norm{\Delta_\omega}_2$, where $L \geq \norm{\phi(s, b)}_2$ for all $(s, b) \in \mathcal{S} \times \mathcal{B}$. Consequently, the error in the HG computation due to the error of the reward estimation is $O(\norm{\Delta_\omega}_2)$. [The proof is provided in \Cref{apdx:prop:bound}.]
\end{proposition}
 
\paragraph{Implementation}\label{sec:implementation}
 
\Cref{algo:proposal} shows the overall framework of the proposed approach, \emph{Reward-Estimated Hypergradient} \emph{RE-HG}.
At each upper-level iteration $k$, the follower first learns its best response under $\theta_k$. In our setting, the leader cannot intervene in this process.
The trajectories, $\mathcal{D}_k$, are then collected for the following IRL steps (Lines~\ref{line:irl0}--\ref{line:irl1}). The HG is computed by using the trajectories, $\mathcal{R}_k$, obtained in simulation with the surrogate follower $\widehat{g}_k$ (Lines~\ref{line:HG0}--\ref{line:HG1}). For the implementation, we remark on two poihttps://www.overleaf.com/projectnts.
First, we use the trajectories $\mathcal{R}_k$ rather than $\mathcal{D}_k$ for HG estimation. As we explicitly utilize $\widehat{g}_k$ itself for the HG estimation, using $\mathcal{R}_k$ will relax the bias due to the inconsistent use of $\widehat{g}_k$ and $\mathcal{D}_k$. 
Second, while the overall framework does not change, environment-specific implementation, for example, the use of dynamic programming when the transition dynamics are known, can simplify and stabilize the resulting algorithm. The details are discussed in \Cref{apdx:exp} and summarized in \Cref{tab:implementation_details}.
 
\begin{algorithm}[t]
\caption{The RE-HG Algorithm}
\label{algo:proposal}
\begin{algorithmic}[1]\small
\STATE{Initialize leader parameters $\theta^0$ and leader's critic $Q_L^0$}
\FOR{$k=0$ to $N-1$}
    \STATE{(Follower) Learn the best response $g^{\theta_k\dagger}$ for $\mathcal{M}_{\theta_k}$}\label{line:interact}
    \STATE{Collect the trajectories $\mathcal{D}_k = \{\tau_k^n\}$, compute feature covariance $\Sigma_k$}
    \STATE{Perform IRL to obtain reward vector $\omega_k$, surrogate policy $\widehat{g}_k$}\label{line:irl0}
    \STATE{Update the aggregated reward vector $\bar{\omega}_k$ by \Cref{alg:bilevel_reward_identification}}\label{line:irl1}
    \STATE{Re-derive surrogate policy $\widehat{g}_k$, action value function $Q_F^k$, and state value function $V_F^k$ using %
    $\bar{\omega}_k$}\label{line:QF}
    \STATE{Collect the trajectories $\mathcal{R}_k = \{\tau_k^n\}$ by simulating $\widehat{g}_k$ on $\mathcal{M}_{\theta_k}$}
    \STATE{$Q_L^{k+1}\gets \texttt{CriticUpdate}(Q_L^{k},\mathcal{R}_k)$ \COMMENT{by \Cref{algo:critic_update_expected_sarsa} or \Cref{algo:critic_update_td_target}}}\label{line:HG0}
    \FOR{$(\bar{s},b)\in\tau\in \mathcal{R}_k$ (where $\bar{s} = s \in \mathcal{S}$ for Conf-MDPs, $\bar{s} = (s, a) \in \mathcal{S}\times\mathcal{A}$ for MGs)}\label{line:for}
        \STATE{$V_L^{k+1}(\bar{s})\gets\E_{b\sim \widehat{g}_k(\cdot\mid \bar{s})}\left[Q_L^{k+1}(\bar{s},b)\right]$}
        \STATE{$B_L^{k+1}(\bar{s},b)\gets Q_L^{k+1}(\bar{s},b)-V_L^{k+1}(\bar{s})$}\label{line:benefit}
        \STATE{$\widehat{\nabla_{\theta_k}Q_F^k}(\bar{s},b)\gets\texttt{FollowerQGrad}(\bar{s},b,\mathcal{R}_k,V_F^k)$ \label{line:FG}\COMMENT{by \Cref{alg:QF-grad}}}
    \ENDFOR
    \STATE{$\widehat{\partial_{\theta_k}J_L^k}\gets\texttt{PartialDerivative}(\mathcal{R}_k,\theta^k,V_L^{k+1},Q_L^{k+1})$ \COMMENT{by \Cref{algo:estimate_partial_grad_Conf-MDP} or \Cref{algo:estimate_partial_grad_2mdp}}}
    \STATE{$\widehat{\nabla_{\theta_k}J_L^k}\gets \widehat{\partial_{\theta_k}J_L^k} + \text{InfluenceTerm}(B_L^{k+1}, \widehat{\nabla_{\theta_k}Q_F^k})$ \label{line:infterm}\COMMENT{by \Cref{algo:influence} or \Cref{algo:baseline_influence}}}
    \STATE{$\theta^{k+1}\gets \theta^k+\alpha\widehat{\nabla_{\theta_k}J_L^k}$}\label{line:HG1}
\ENDFOR
\end{algorithmic}
\end{algorithm}
 
\section{Experiments}
\label{sec:experiments}
We evaluate the robustness of \rehg against an \oracle baseline that computes HGs with full access to the ground-truth rewards and transition dynamics. This baseline relies on stronger assumptions than the standard white-box follower assumption utilized in \citet{DBLP:conf/nips/ThomaP0RH24,Kudo2026icaps}, as it requires access to the transition dynamics. As an idealized model, it represents the empirical upper bound and serves as our primary benchmark for gradient accuracy and overall performance.
 
\paragraph{2-Player Markov Games (Discrete Case)}\label{sec:2mg-dis}
 
\begin{figure}[t]
  \centering
  \begin{subfigure}[b]{0.20\textwidth}
    \centering
\begin{tikzpicture}[
        scale=0.6, transform shape, 
        show background rectangle,
        background rectangle/.style={fill=none, draw=none},
        inner frame sep=1.5mm, 
        ->, >=stealth, auto,
        node distance=5em, 
        every state/.style={fill=black!15, draw=black, text=black},
        every node/.style={align=center, font=\small}
    ]
 
        \node[state] (S1) {S1};
        \node[state, below=of S1] (S0) {S0};
        \node[state, below=of S0] (S2) {S2};
 
        \path
        (S1) edge [loop left]  node {$A_0|B_1$} (S1)
        (S0) edge [loop left]  node {$*|B_0$} (S0)
        (S2) edge [loop left]  node {$*|B_1$,\\$*|B_2$} (S2)
 
        (S1) edge [bend right=20, left] node {$A_0|B_0$,\\ $A_1|*$~\colorbox{gray!30}{0$\mid$-1}} (S0)
        (S1) edge [bend left=60, left] node {$A_0|B_2$\\ \colorbox{gray!30}{0$\mid$1}} (S2)
        (S0) edge [bend right=20, left] node {$*|B_2$\\ \colorbox{gray!30}{0$\mid$1}} (S2)
        (S0) edge [bend right=20, right] node {$*|B_1$\\ \colorbox{gray!30}{1$\mid$1}} (S1)
        (S2) edge [bend right=20, right] node {$*|B_0$} (S0);
\end{tikzpicture}
    \caption{Environment}
    \label{fig:state_machine_sub}
  \end{subfigure}%
  \begin{subfigure}{0.8\textwidth}
    \centering
    \includegraphics[width=0.9\textwidth,clip,trim=0 0 0 0]{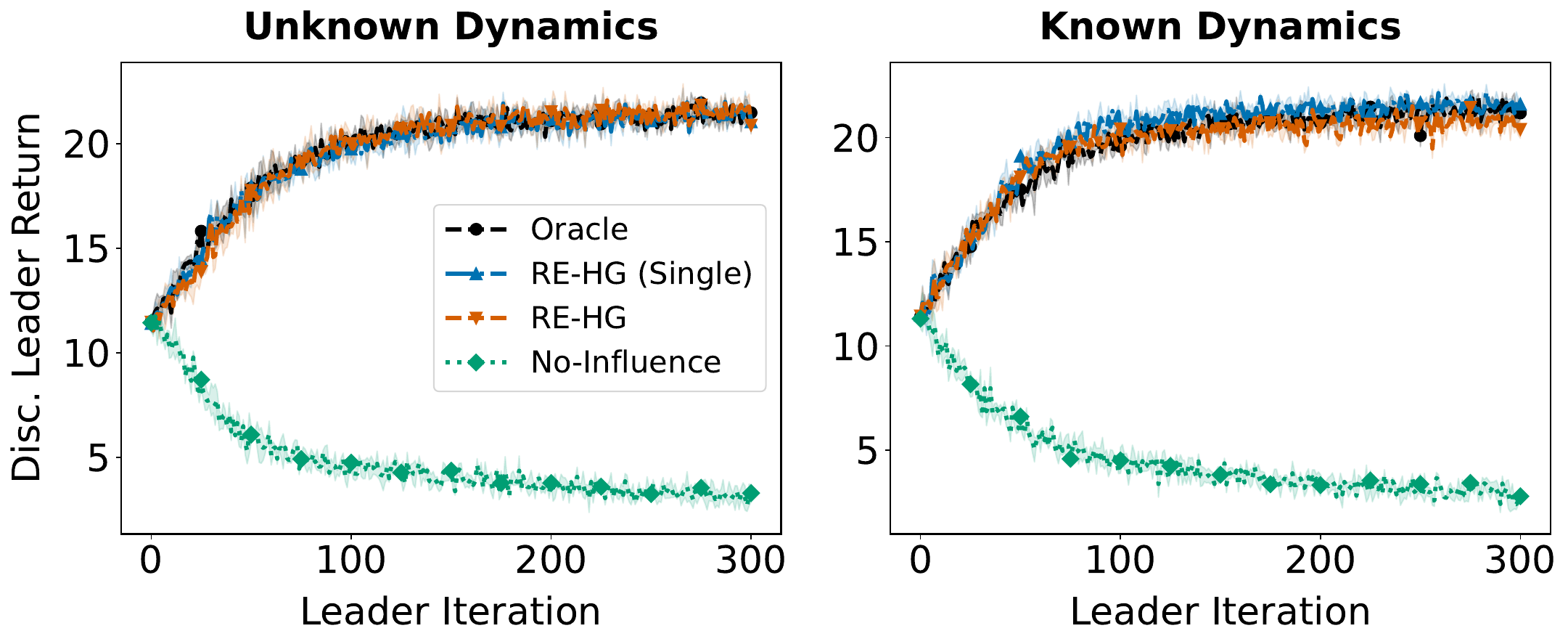}
    \caption{Discounted leader return under unknown dynamics (left) and known dynamics (right)}
    \label{fig:comparison_curves_sub}
  \end{subfigure}%
  \caption{Discrete toy task. (a) State transition diagram. \colorbox{gray!20}{$r_L$ | $r_F$} represents the rewards. (b) 
  Leader's discounted return across 3 independent training runs. For each run, performance is evaluated as the mean of 10 test rollouts. Avg.\ (line) and std.\ (shaded area) across the 3 runs are displayed.
  }
  \label{fig:integrated_results}
\end{figure}
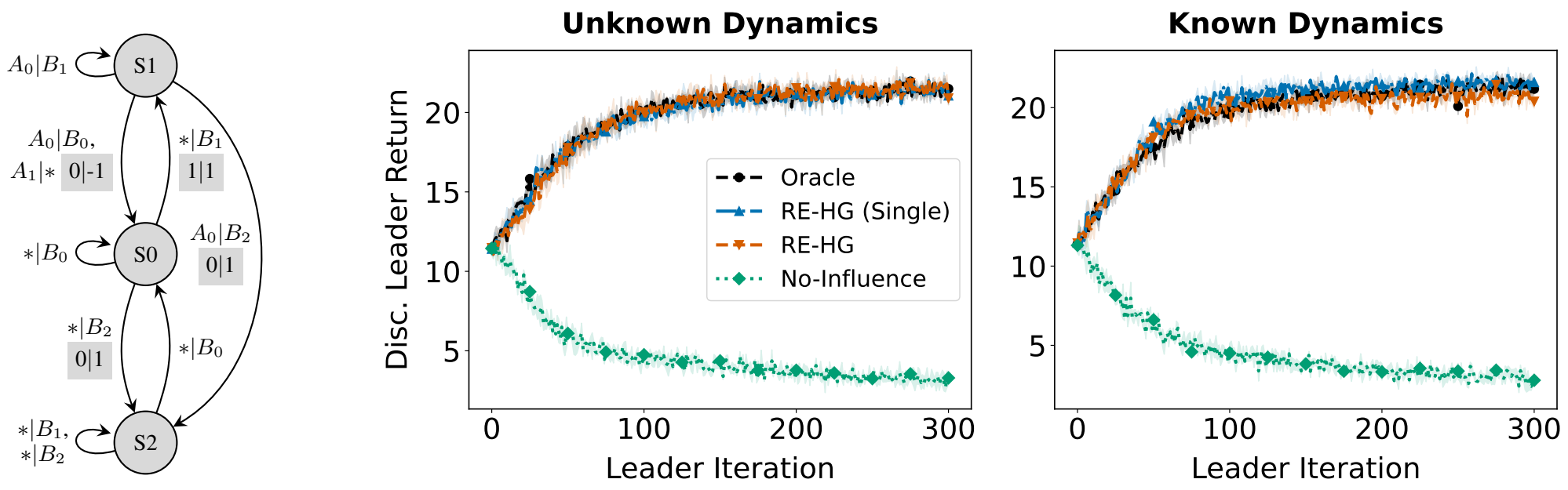

To assess the necessity of the influence term and the HG accuracy of \rehg, we design an environment (\Cref{fig:state_machine_sub}) with an inherent leader-follower conflict.
 
The leader earns $r_L = +1$ only on $S_0 \to S_1$, favoring an $S_0 \leftrightarrow S_1$ loop. Conversely, the follower prefers $S_0 \to S_1 \to S_2$ to maximize $r_F = +1$. At $S_1$, the leader can choose $A_1$ to force the loop, but this imposes a penalty $r_F = -1$. If the leader greedily persists with $A_1$, the follower will adapt by transitioning $S_0 \to S_2$ to avoid the penalty, eventually depriving the leader of all rewards. Thus, the leader must strategically guide the follower to $S_1$ while preventing deflection to $S_2$ by choosing $A_0$ at appropriate intervals.

As observed in \Cref{fig:comparison_curves_sub}, \rehg and \oracle significantly outperform \noinf, a variant ignoring the influence term. 
It clearly indicates the need to estimate the influence term. Specifically, the leader policies in \rehg and \oracle converge to $f(A_0 | S_1) \approx 1.0$. In contrast, \noinf leader policy converges to $f(A_0 | S_1) \approx 0.0$, resulting in the breakdown described above. The relatively low Pearson correlation stems from estimation errors in unvisited regions, which do not affect the gradient direction of the HG. In fact, \rehg maintains a high cosine similarity. Essentially, \rehg successfully recovers the global reward structure necessary for optimization. Further analysis of the Pearson and cosine similarities and experimental details are provided in \Cref{apdx:2mg-discrete}.

\begin{figure}[t]
    \centering
    \begin{subfigure}{0.28\textwidth}
    \includegraphics[width=\hsize]{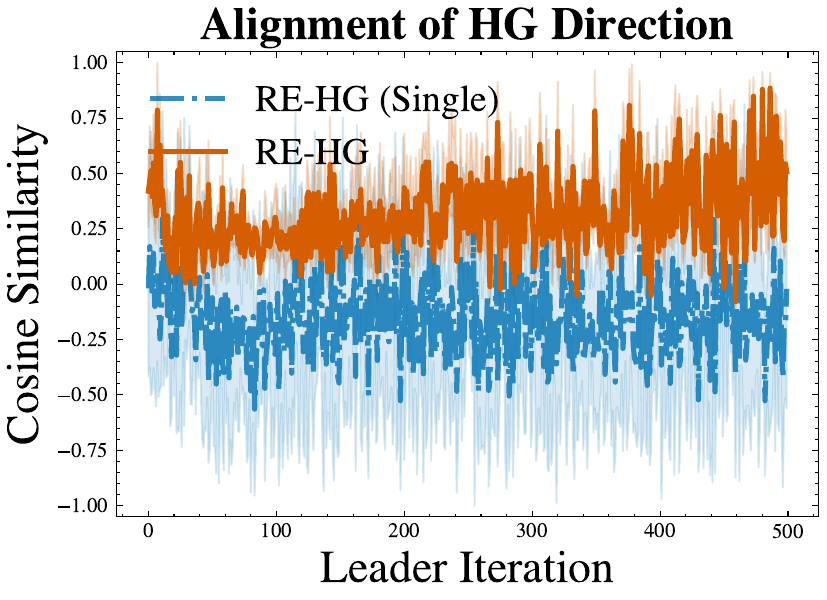}
    \caption{HG Similarity}\label{fig:hg_cosine_similarity}
    \end{subfigure}%
    \begin{subfigure}{0.28\textwidth}
    \includegraphics[width=\hsize]{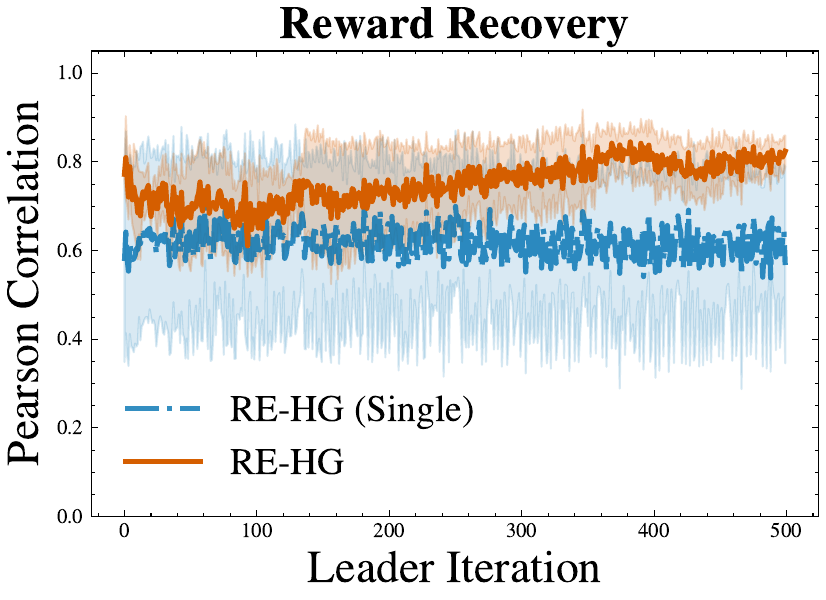}
    \caption{Reward Correlation}\label{fig:reward_pearson}
    \end{subfigure}%
    \begin{subfigure}{0.44\textwidth}
    \includegraphics[width=\hsize]{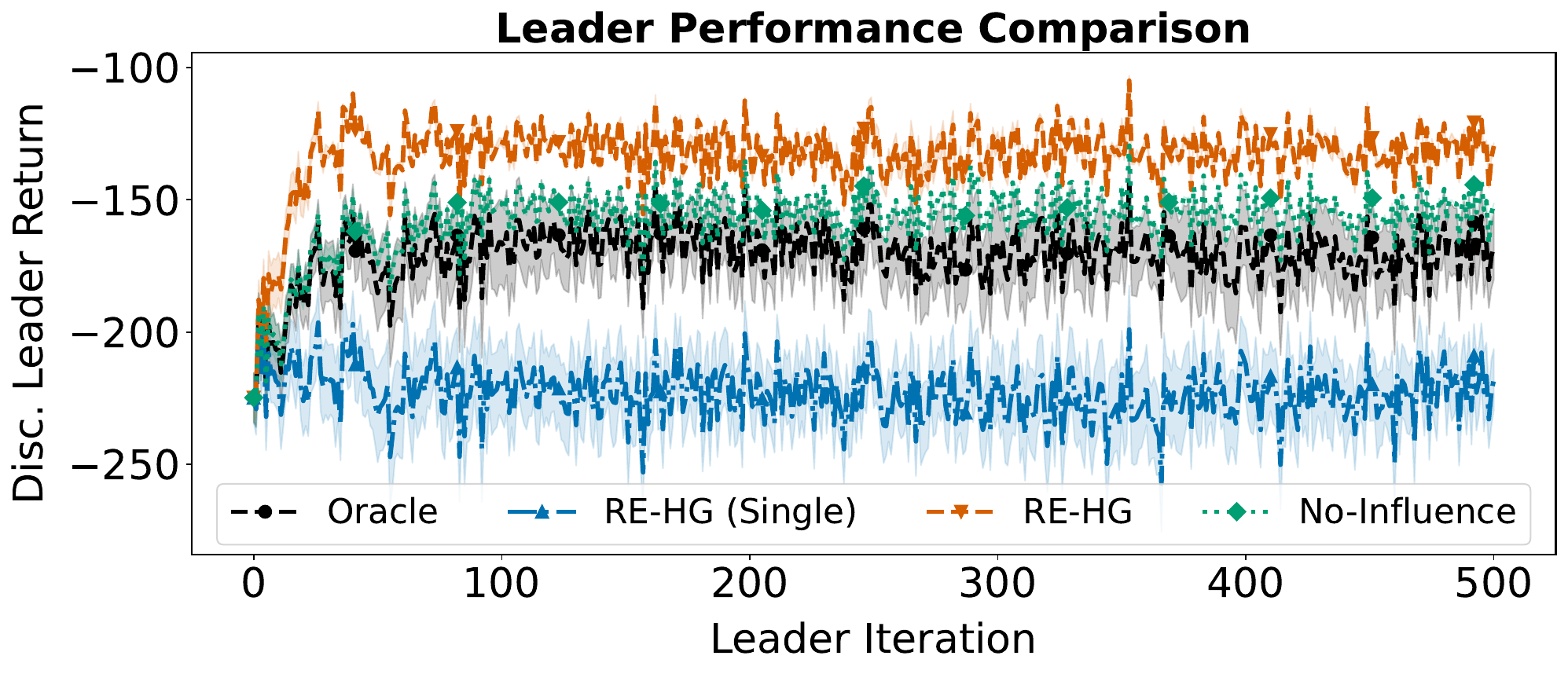}
    \caption{Leader's Discounted Return}
    \label{fig:confpafo}
    \end{subfigure}
    \caption{Results on Building Thermal Control Task. (a) Cosine similarity of the HG between the estimated and Oracle models. (b) Pearson correlation between the recovered and true rewards, weighted by the undiscounted state-action occupancy measure of the surrogate follower policy. (c) Leader's discounted return. 
    For each run, performance is evaluated as the mean of 50 test rollouts. Avg.\ (line) and std.\ (shaded area) across the 5 independent training runs are displayed.
    }
\end{figure}

\paragraph{Configurable MDP (Continuous Case)}\label{sec:c-continuous}
We evaluate the effectiveness of the aggregation mechanism in \rehg for mitigating estimation instability by comparing it against \rehgsingle, a variant performing IRL only once at the beginning.
The task is under-actuated temperature regulation in a building with $n=8$ zones. The state $s_t \in \mathbb{R}^n$ is the temperature deviation, the follower's action $b \in \mathbb{R}^m$ ($m=4$) is the HVAC control, and the leader configures the insulation and airflow $\theta = (\alpha, \nu) \in [0,1]^{2n}$. The system follows linear-Gaussian dynamics $s^\prime = A_\theta s + B b + w$, where $w \sim \mathcal{N}(\bm{0}, \Sigma_w)$ represents unobservable disturbances. The follower maximizes an entropy-regularized quadratic reward $r_F(s, b) = -s^\top W_F s - b^\top U_F b$, while the leader minimizes insulation costs and thermal instability. Crucially, because the most costly zone lacks direct HVAC control, the follower must regulate it indirectly via heat transfer influenced by the leader's configuration. In this experiment, identical gradient norm clipping with a maximum norm of 1.0 is applied to every method.
 
Comparisons in \Cref{fig:confpafo} confirm that the proposed method, \rehg, overcomes the estimation instability of \rehgsingle and clearly outperforms it. Lacking bias cancellation and eigenvalue truncation, \rehgsingle recovers the excessively steep reward, causing the gradient norm to reach the order of $10^7$. Because estimation errors propagate directly into the gradient, the optimization is misled into an inappropriate local minimum, resulting in significant performance degradation. In contrast, \rehg suppresses the reward norm through the aforementioned two implicit regularization effects of the aggregation process, keeping the gradient magnitude around $10^1$. Although the values are not perfectly identical due to the scale suppression effect, \Cref{fig:hg_cosine_similarity,fig:reward_pearson} show that \rehg achieves a higher cosine similarity with \oracle and a higher Pearson correlation with the true reward weighted by the surrogate follower policy compared to \rehgsingle. This indicates that the global gradient direction and reward structure are appropriately recovered.
 
As a secondary effect, this smoothing leads \rehg to attain a higher return than \oracle. We stress that this is an optimization effect and not an indication that \rehg estimates the HG more accurately. While \oracle computes the exact true gradient, the follower's skewed cost matrices ($W_F$, $U_F$) create a steep reward landscape, causing the gradient scale to explode (around $10^3$). Consequently, terms other than the influence term cannot be properly accounted for, severely distorting the overall gradient direction and trapping \oracle in a suboptimal local minimum (behavior similar to \noinf converging to a local minimum). On the other hand, by recovering a smoothed reward bounded within a moderate scale ($0.2$ to $3.2$) rather than the steep true reward, \rehg obtains a gradient landscape on which all terms are appropriately balanced, avoiding the sharp local optima that trap \oracle. This gap is not explained by the magnitude of the update alone: \Cref{apdx:oracle-ablation} reports \oracle variants with a learning rate, gradient clipping, and a scaled influence term, none of which reaches the performance of \rehg.
 
Further details, including the improvement of cosine similarity and Pearson correlation when the steep reward setting is relaxed, and experimental setups, are provided in \Cref{apdx:Conf-MDP}.
 
\paragraph{2-Player Markov Games (Continuous Case)}\label{sec:2mg-continuous}
 
\begin{wrapfigure}{R}{0.65\textwidth}
    \centering
    \includegraphics[width=\hsize,clip,trim=8 7 6 8]{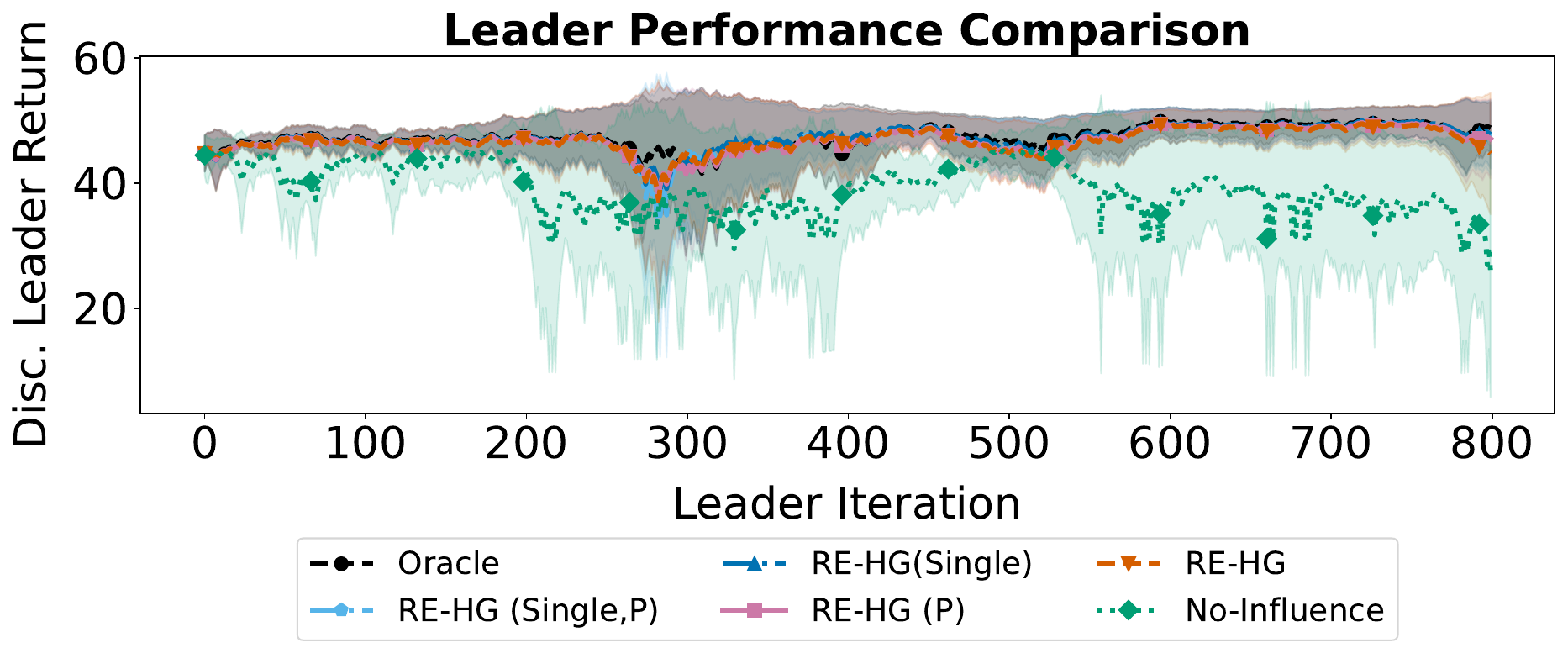}
    \caption{Results on 2-player MG, averaged over 3 independent training runs (10 test rollouts per run). Avg.\ (line) and std.\ (shaded area) are displayed. ``P'' indicates the known transition probability scenario.}
    \label{fig:2pmgt.pdf}
\end{wrapfigure}
 
We evaluate the HG accuracy of \rehg on a continuous MG formulated as a Linear Quadratic Regulator (LQR) task. The dynamics are $s_{t+1} = A s_t + B b_t + C a_t$, where $s_t \in \mathbb{R}^2$ is the state (starting at $(1,0)$), $b_t \in \mathbb{R}^1$ is the follower's control, and $a_t \in \mathbb{R}^1$ is the leader's interference. Both agents employ Gaussian policies.
The follower aims to stabilize at $s^\star_F = (0,0)$ by maximizing a quadratic reward $r_F(s, b) = -s^\top W_F s - b^\top U_F b$, where counteracting the leader's interference increases control costs. Conversely, the leader seeks to induce oscillation around $s^\star_L = (-0.8, 0)$ by maximizing $r_L(s, a, b) := \exp( -\frac{1}{2} (s - s^\star_L)^\top M (s - s^\star_L) )$. 
This explicit conflict of interest tests whether the leader can strategically guide the follower by accounting for its cost structure via HG, rather than relying on inefficient brute-force interference.

\Cref{fig:2pmgt.pdf} shows that \rehg achieves comparable performance to \oracle, effectively navigating the true gradient direction despite the presence of estimation errors. Furthermore, it is more stable than \noinf. Both the HG cosine similarity relative to \oracle and the weighted Pearson correlation consistently remained near $1.0$ across the majority of the iterations for \rehg. Figures illustrating these metrics, along with full experimental settings, are provided in \Cref{apdx:2mg-continuous}.

\section{Conclusion}
We introduced RE-HG, an algorithm for optimizing the leader's parameters via HG in a bilevel RL setting with a black-box follower. Our theoretical analysis revealed that HG estimation is fundamentally vulnerable to reward shaping, an unavoidable issue in single-environment IRL. To resolve this, we established a bias cancellation theory showing that, under a linear reward assumption, the here-proposed aggregated IRL mathematically neutralizes errors caused by potential-based shifts. Based on this theory, we developed a reward recovery algorithm enhanced by a norm suppression smoothing effect. Empirically, despite operating without the restrictive white-box follower assumption, RE-HG achieved performance competitive with a white-box Oracle baseline by yielding accurate gradient direction estimates.
 
\paragraph{Limitations}
RE-HG has several limitations.
First, our theoretical framework relies on the assumption of an optimal follower. Whether RE-HG operates stably when the follower follows sub-optimal learning trajectories remains open for discussion; \Cref{apdx:suboptimal-follower} reports an empirical study under two types of suboptimal followers, but our analysis does not cover this case.
Second, formulating the IRL as an inner-loop min-max optimization inherently introduces estimation variance that directly propagates to the outer-loop HG. Stabilizing the influence term remains an open challenge.
Third, our bias cancellation theory is built upon the linear reward assumption with a known feature extractor. Robustness to a misspecified feature basis is examined empirically in \Cref{apdx:misspec}, but extending the theory to non-linear reward structures requires further theoretical and empirical investigation; integrating multi-environment IRL techniques designed for non-linear representations is a natural direction.
Finally, RE-HG removes the need for the follower's reward function, but not the need for environment access: computing and rolling out the surrogate best response requires either the transition model or a simulator. Integrating offline IRL is a natural direction for removing this dependence.
 
\begin{ack}
This work was partially supported by JSPS KAKENHI Grant Number 23H00483 and 26K02993.
\end{ack}

\bibliographystyle{apalike}
\bibliography{ref}

\clearpage
\appendix
\section{Proofs}
\subsection{\Cref{prop:irl}}\label{apdx:prop:irl}
\begin{proof}
As is claimed in \citet{Zeng2022neurips}, the process of $\textsc{IRL}_{\theta}$ can be interpreted as a maximum likelihood estimation of $g^{\theta\dagger}$ by the policy $g_{r}^\star$ optimal for the reward model $r$, where the reward parameter is considered as the parameter of the policy itself. Indeed, we observe
\begin{align}
\textsc{IRL}_\theta(g^{\theta\dag})
&= \argmin_{r} \max_{g} J_\theta(g; r) - \E_{\rho_0^{\theta}, p^\theta, g^{\theta\dagger}} \left[ \sum_{t=0}^{\infty} \gamma^t r(s_t, b_t) \right] \\
&= \argmin_{r} \max_{g} \E_{\rho_0^{\theta}}\left[ V_{r, g}^\mathrm{soft}(s_0) \right] - \E_{\rho_0^{\theta}, p^\theta, g^{\theta\dagger}} \left[ \sum_{t=0}^{\infty} \gamma^t r(s_t, b_t) \right] \\
&= \argmin_{r} \E_{\rho_0^{\theta}}\left[ V_{r, g_r^\star}^\mathrm{soft}(s_0) \right] - \E_{\rho_0^{\theta}, p^\theta, g^{\theta\dagger}} \left[ \sum_{t=0}^{\infty} \gamma^t r(s_t, b_t) \right] \\
&= \argmin_{r} \E_{\rho_0^{\theta}, p^\theta, g^{\theta\dagger}}\left[ \sum_{t=0}^{\infty} \gamma^t \left( V_{r, g_r^\star}^\mathrm{soft}(s_t) - \gamma V_{r, g_r^\star}^\mathrm{soft}(s_{t+1}) \right) \right] - \E_{\rho_0^{\theta}, p^\theta, g^{\theta\dagger}} \left[ \sum_{t=0}^{\infty} \gamma^t r(s_t, b_t) \right] \\
&= \argmax_{r} \E_{\rho_0^{\theta}, p^\theta, g^{\theta\dagger}}\left[ \sum_{t=0}^{\infty} \gamma^t \left( r(s_t, b_t) + \gamma V_{r, g_r^\star}^\mathrm{soft}(s_{t+1}) - V_{r, g_r^\star}^\mathrm{soft}(s_t)\right) \right] \\
&= \argmax_{r} \E_{\rho_0^{\theta}, p^\theta, g^{\theta\dagger}}\left[ \sum_{t=0}^{\infty} \gamma^t \left( Q_{r, g_r^\star}^\mathrm{soft}(s_{t}, b_{t}) - V_{r, g_r^\star}^\mathrm{soft}(s_t) \right)\right]\\
&= \argmax_{r} \E_{d_{\gamma_F}^{\theta,g^{\theta\dagger}}}\left[\E_{g^{\theta\dagger}}\left[  \log g_r^\star(s, b) \right] \right] .
\end{align}

Let $\ell(s, b; \omega) = \log g_{r_\omega}^\star(s, b)$.
Let $V_\omega^\mathrm{soft}$ and $Q_\omega^\mathrm{soft}$ be the optimal soft value functions under the reward model $r_\omega$.
Then, we have
\begin{align}
\nabla_\omega \ell(s, b; \omega)
&= \nabla_\omega Q_\omega^\mathrm{soft}(s, b) - \nabla_\omega V_\omega^\mathrm{soft}(s)
\\
&= \nabla_\omega Q_\omega^\mathrm{soft}(s, b) - \E_{g_\omega^\star}[\nabla_\omega Q_\omega^\mathrm{soft}(s, b) ],
\end{align}
where we used the soft-Bellman equation for value function, $V_\omega^\mathrm{soft}(s) = \log \int \exp( Q_\omega^\mathrm{soft}(s, b)) \mathrm{d}b$.
On the other hand, using the soft-Bellman equation for action value function, we obtain
\begin{align}
\nabla_\omega Q_\omega^\mathrm{soft}(s, b) = \phi(s, b) + \gamma \E_{s'}\left[ \E_{g_\omega^\star} [ \nabla_\omega Q_\omega^\mathrm{soft}(s', b') ]\right].
\end{align}
The solution to this recursive formula is equivalent to the expected discounted features, i.e., $\nabla_\omega Q_\omega^\mathrm{soft}(s, b) = \psi_\omega(s, b)$.
Therefore, we have
\begin{align}
\nabla_\omega \ell(s, b; \omega) = \psi_{\omega}(s, b) - \E_{g_\omega^\star}[\psi_\omega(s, b) ].
\end{align}
Moreover, the Hessian is
\begin{align}
\nabla_\omega^2 \ell(s, b; \omega)
&= \nabla_\omega \psi_{\omega}(s, b) - \E_{b' \sim g_\omega^\star}[\psi_\omega(s, b') \nabla \ell(s, b'; \omega)^\top] - \E_{b' \sim g_\omega^\star}[\nabla_\omega \psi_\omega(s, b') ]
\\
&= \nabla_\omega \psi_{\omega}(s, b) - \E_{b' \sim g_\omega^\star}[\nabla_\omega \psi_\omega(s, b') ] - \Cov_{b' \sim g_\omega^\star}[\psi_\omega(s, b') ].
\end{align}

Let $L(\omega) = \E_{d_{\gamma_F}^{\theta,g^{\theta\dagger}}}\left[\E_{g^{\theta\dagger}}\left[-\ell(s, a; \omega)\right]\right]$. Noting that $g^{\theta\dagger} = g_{\omega^\star}^\star$, the Hessian of this loss function at $\omega = \omega^\star$ is
\begin{align}
\nabla_\omega^2 L(\omega^\star)
&= -\E_{d_{\gamma_F}^{\theta,g^{\theta\dagger}}}\left[\E_{g^{\theta\dagger}}\left[\nabla_\omega^2 \ell(s, b; \omega^\star)\right]\right]
\\
&= -\E_{d_{\gamma_F}^{\theta,g^{\theta\dagger}}}\left[\E_{g^{\theta\dagger}}\left[\nabla_\omega \psi_{\omega^\star}(s, b) - \E_{g_{\omega^\star}^\star}[\nabla_\omega \psi_{\omega^\star}(s, b) ] - \Cov_{g_{\omega^\star}^\star}[\psi_{\omega^\star}(s, b) ]\right]\right]
\\
&= -\E_{d_{\gamma_F}^{\theta,g_{\omega^\star}}}\left[\E_{g_{\omega^\star}}\left[\nabla_\omega \psi_{\omega^\star}(s, b)\right] - \E_{g_{\omega^\star}^\star}[\nabla_\omega \psi_{\omega^\star}(s, b) ] - \Cov_{g_{\omega^\star}^\star}[\psi_{\omega^\star}(s, b) ]\right]
\\
&= \E_{d_{\gamma_F}^{\theta,g_{\omega^\star}^\star}}\left[\Cov_{g_{\omega^\star}^\star}[\psi_{\omega^\star}(s, b) ]\right].
\end{align}
Hence, it is non-negative definite.
If two reward functions $r_0$ and $r_1$ admit the same optimal policy, then their convex combination $r_\lambda = \lambda r_0 + (1 - \lambda) r_1$ for $\lambda \in [0, 1]$ also admits the same optimal policy.
This implies that all the optimal reward parameter must lie in the kernel of $\nabla_\omega^2 L(\omega^\star)$ centered at $\omega^\star$.
This completes the proof.
\end{proof}

\subsection{\Cref{thm:identifiability}}\label{apdx:thm:identifiability}
\begin{proof}
From \Cref{prop:irl}, we have $\Sigma_k (\omega_k - \omega_\star) = 0$ for each $k \in [K]$. Then, we have
\begin{align}
 \bar{\Sigma}_K \bar{\omega}_{K}
 = \sum_{k=1}^{K} \Sigma_k \omega_k
 = \sum_{k=1}^{K} \Sigma_k \omega^\star
 = \bar{\Sigma}_K \omega^\star.
\end{align}
Therefore, we have $\bar{\Sigma}_K (\bar{\omega}_K - \omega^\star) = 0$, which indicates that $\bar{\omega} - \omega^\star$ lies in the kernel of $\bar{\Sigma}_K$. Because $\Sigma_k$ are all non-negative definite, the kernel of $\bar{\Sigma}_K$ is the intersection of the kernels of $\Sigma_k$. Hence, we have $\bar{\omega}_K \in \textsc{IRL}_{\theta_k}(g^{\theta_k \dagger})$ for all $k \in [K]$.

The rest is straightforward. If $\Delta_\mathrm{const} = 0$ and $\textsc{Rank}(\bar{\Sigma}_K) = d$, there exists a unique solution to $\bar{\Sigma}_K \omega = \sum_{k=1}^{K} \Sigma_k \omega_k$ and it must agree with $\omega^\star$. If $\Delta_\mathrm{const} \neq 0$ and $\textsc{Rank}(\bar{\Sigma}_K) = d - 1$, then $\Delta_\mathrm{const}$ must lie in the kernel of $\bar{\Sigma}_K$ because $\omega + c \Delta_\textrm{const}$ results in a constant offset of the reward function as
\begin{align}
r_{\omega + c \Delta_\mathrm{const}}(s, b)
= \inner{\omega + c \Delta_\mathrm{const}}{\phi(s, b)}
= \inner{\omega}{\phi(s, b)} + c,
\end{align}
and hence the loss function $L$ must be constant in the direction of $\Delta_\mathrm{const}$. That is, the kernel of $\bar{\Sigma}_K$ is spanned by $\Delta_\mathrm{const}$. Therefore, we have $\bar{\omega}_K = \omega^\star + c \Delta_\mathrm{const}$. This completes the proof.
\end{proof}

\subsection{\Cref{cor:conv}}\label{apdx:cor:conv}
\begin{proof}
In this proof, we consider $K \geq \bar{K}$. 
Then, from the assumption $\textsc{Rank}(\bar{U}_K) = d - \ind{\Delta_\mathrm{const} \neq 0}$, the kernel of $\bar{U}_K$ must admit the kernel of $W$. Therefore, $W = \bar{U}_K \bar{U}_K^{-} = \bar{U}_K^{-} \bar{U}_K$. Moreover, $W \bar{U}_K^{-} = \bar{U}_K^{-}$ and $W \bar{U}_K = \bar{U}_K$ because of the property of pseudo-inverse.

Note first that $\bar{\omega}_K$ is a solution to $\bar{U}_K \omega = \sum_{k=1}^{K} U_k \widehat{\omega}_k$. 
The solution can be represented as 
$$\bar{\omega}_K = \sum_{k=1}^{K} U_k \widehat{\omega}_k + (I - W) \delta$$
for some $\delta \in \R^d$. 
Because 
$W = I - \norm{\Delta_\mathrm{const}}_2^{-2} \Delta_\mathrm{const}\Delta_\mathrm{const}^\top$,
we have 
$$\bar{\omega}_K = \sum_{k=1}^{K} U_k \widehat{\omega}_k + \norm{\Delta_\mathrm{const}}_2^{-2}\inner{\delta}{\Delta_\mathrm{const}} \Delta_\mathrm{const}.$$
Letting $c = \norm{\Delta_\mathrm{const}}_2^{-2}\inner{\delta}{\Delta_\mathrm{const}}$, we have
RE$\bar{\omega}_K = \sum_{k=1}^{K} U_k \widehat{\omega}_k + c \Delta_\mathrm{const}$.
It indicates that 
\begin{align*}
    \Delta_\omega^K 
    &= W(\bar{\omega}_K - \omega^\star) \\
    &= W \bar{U}_K^{-} \sum_{k=1}^{K} U_k \widehat{\omega}_k - W \omega^\star \\
    &= \bar{U}_K^{-} \sum_{k=1}^{K} U_k \widehat{\omega}_k - \bar{U}_K^{-} \bar{U}_K \omega^\star \\
    &= \bar{U}_K^{-} \sum_{k=1}^{K} U_k (\widehat{\omega}_k - \omega^\star).
\end{align*}

From \Cref{prop:irl}, we have $\Sigma_k (\omega_k - \omega^\star) = 0$, which is equivalent to $U_k( \omega_k - \omega^\star) = 0$ because $\Sigma_k$ and $U_k$ admit the same kernel by construction. Therefore, $\E[U_k \widehat{\omega}_k] = U_k \omega_k$ implies $\E[U_k \widehat{\omega}_k] = U_k \omega^\star$. 
Then, we have 
\begin{align*}
\E[\Delta_\omega^K] 
&= \bar{U}_K^{-} \sum_{k=1}^{K} U_k (\widehat{\omega}_k - \omega^\star) \\
&= \bar{U}_K^{-} \sum_{k=1}^{K} \E[U_k \widehat{\omega}_k] - U_k \omega^\star \\
&= 0.
\end{align*}

Because $\widehat{\omega}_k$ are independently distributed, we have 
\begin{align*}
W \Cov[\bar{\omega}_K]W 
&= \E[\Delta_\omega^K (\Delta_\omega^K)^\top] \\
&= \bar{U}_K^{-}  \sum_{k=1}^{K} \E[U_k (\widehat{\omega}_k - \omega^\star)(\widehat{\omega}_k - \omega^\star)^\top U_k] \bar{U}_K^{-} \\
&\preccurlyeq C \cdot K (\bar{U}_K^{-})^2.
\end{align*}

In light of the multivariate Chebyschev's inequality, we have
\begin{equation}
\Pr[ \norm{\Delta_\omega^K}_{\Cov[\Delta_\omega^K]^{-}} > t ] \leq \frac{d}{t^2}
\end{equation}
for any $t > 0$. 
Let $\sigma_{\max}(\bar{U}_K^{-})$ be the greatest eigenvalue of $\bar{U}_K^{-}$. Then, we have $\sigma_{\min} ( \Cov[\Delta_\omega^K]^{-} ) \geq (C K \sigma_{\max}(\bar{U}_K^{-})^2)^{-1}$, where $\sigma_{\min}$ represents the smallest non-zero eigenvalue. Then, we have
$\norm{\Delta_\omega^K}_{\Cov[\Delta_\omega^K]^{-}} \geq \sqrt{\sigma_{\min}(\Cov[\Delta_\omega^K]^{-}) }\norm{\Delta_\omega^K}_{2}$, and hence,
\begin{equation}
\Pr[ \norm{\Delta_\omega^K}_{2} > t ] \leq \frac{d }{t^2 \sigma_{\min}(\Cov[\Delta_\omega^K]^{-})} \leq \frac{d C K \sigma_{\max}(\bar{U}_K^{-})^2}{ t^2} \xrightarrow{K \to \infty} 0 .
\end{equation}
It implies $\lim_{K \to \infty} \Delta_\omega^K = 0$ in probability. This completes the proof.
\end{proof}

\subsection{Adaptive Variant of \Cref{cor:conv}}\label{apdx:adaptive}
$\Sigma_k$ is estimated from the batch $\mathcal{D}_k$ collected under $\theta_k$, so $U_k$ is not $\mathcal{F}_{k-1}$-measurable, and $U_k$ and $\widehat{\omega}_k$, being functions of the same $\mathcal{D}_k$, are dependent.

\Cref{cor:conv} treats $\{U_k\}_{k=1}^{K}$ as a deterministic sequence and assumes the estimates $\widehat{\omega}_k$ to be independent. Neither holds in \Cref{algo:proposal}: $\Sigma_k$ is estimated from the batch $\mathcal{D}_k$ collected under $\theta_k$, so $U_k$ is not $\mathcal{F}_{k-1}$-measurable, and $U_k$ and $\widehat{\omega}_k$, being functions of the same $\mathcal{D}_k$, are dependent. 

We state below the assumptions under which the conclusion of \Cref{cor:conv} survives, and note in \Cref{rem:measurability} that one of them holds for a lagged or sample-split version of \Cref{algo:proposal} rather than for the algorithm as implemented.
 
\begin{corollary}\label{cor:adaptive}
Let $\{U_k\}_{k=1}^{K}$ and $\{\widehat{\omega}_k\}_{k=1}^{K}$ be the
projection matrices and the estimated reward vectors generated by the closed
loop of \Cref{algo:proposal}, let $\bar{U}_K = \sum_{k=1}^{K} U_k$, and let
$\mathcal{F}_k$ be the $\sigma$-algebra generated by the history
$(\theta_1,\mathcal{D}_1,\widehat{\omega}_1,\ldots,\theta_k,\mathcal{D}_k,\widehat{\omega}_k)$,
where $\mathcal{D}_k$ denotes the trajectories collected under $\theta_k$.
Let $b \geq 0$ and $C \geq 0$ be constants and suppose that, for every $k$,
\begin{enumerate}
\item[(i)] $U_k$ is $\mathcal{F}_{k-1}$-measurable;
\item[(ii)] $\E[U_k\widehat{\omega}_k \mid \mathcal{F}_{k-1}]
 = U_k\omega_k + U_k b_k$ for some $b_k \in \R^d$ with $\norm{b_k}_2 \leq b$;
\item[(iii)] $\Cov[U_k\widehat{\omega}_k \mid \mathcal{F}_{k-1}]
 \preccurlyeq C \cdot I$;
\end{enumerate}
and that
\begin{enumerate}
\item[(iv)] $\textsc{Rank}(\bar{U}_K) = d - \ind{\Delta_\mathrm{const} \neq 0}$
 for all $K \geq \bar{K}$, for some $\bar{K} \geq 1$.
\end{enumerate}
Let $W$ and $\Delta_\omega^K$ be as in \Cref{cor:conv}, let
$\eta_k = U_k\widehat{\omega}_k - \E[U_k\widehat{\omega}_k \mid \mathcal{F}_{k-1}]$,
and let $\sigma_{\min}$ denote the smallest non-zero eigenvalue. Then, for
every $K \geq \bar{K}$, the output of \Cref{alg:bilevel_reward_identification}
decomposes as
$\Delta_\omega^K = \Delta_\omega^{K,\mathrm{bias}} + \Delta_\omega^{K,\mathrm{noise}}$
with $\Delta_\omega^{K,\mathrm{bias}} = \bar{U}_K^{-}\sum_{k=1}^{K} U_k b_k$
and $\Delta_\omega^{K,\mathrm{noise}} = \bar{U}_K^{-}\sum_{k=1}^{K} \eta_k$,
and these satisfy
\begin{align}
\E\left[ \norm{\textstyle\sum_{k=1}^{K} \eta_k}_2^2 \right] &\leq C d K,
 \label{eq:adaptive-noise} \\
\norm{\Delta_\omega^{K,\mathrm{bias}}}_2
 &\leq \frac{K b}{\sigma_{\min}(\bar{U}_K)} . \label{eq:adaptive-bias}
\end{align}
Consequently, if $\Pr[\sigma_{\min}(\bar{U}_K) \geq \sqrt{K} a_K] \to 1$ for
some deterministic sequence $a_K \to \infty$, then
$\Delta_\omega^{K,\mathrm{noise}} \to 0$ in probability; and if
$\Pr[\sigma_{\min}(\bar{U}_K) \geq \alpha K] \to 1$ for some $\alpha > 0$,
then $\Pr[\norm{\Delta_\omega^{K,\mathrm{bias}}}_2 > b/\alpha] \to 0$. In
particular, if $b = 0$ and the first spectral condition holds, then
$\Delta_\omega^K \to 0$ in probability, which is the conclusion of
\Cref{cor:conv} without the independence assumption.
\end{corollary}
 
\begin{proof}
Throughout we take $K \geq \bar{K}$. Line~3 of
\Cref{alg:bilevel_reward_identification} sets $U_k = E I_{\Lambda \succ \epsilon} E^\top$
with $E$ orthogonal and $I_{\Lambda \succ \epsilon}$ diagonal with entries in
$\{0,1\}$, so each $U_k$ is an orthogonal projection; we use this in Steps~2
and~4.
 
\emph{Step 1.} By (i), $U_k$ is $\mathcal{F}_{k-1}$-measurable,
so $\E[U_k\widehat{\omega}_k \mid \mathcal{F}_{k-1}] = U_k\E[\widehat{\omega}_k \mid \mathcal{F}_{k-1}]$
and hence
\begin{equation}\label{eq:eta-form}
\eta_k = U_k\left(\widehat{\omega}_k - \E[\widehat{\omega}_k \mid \mathcal{F}_{k-1}]\right) \quad \text{and}
\qquad \E[\eta_k \mid \mathcal{F}_{k-1}] = 0 .
\end{equation}
Let $\xi_k = U_k\widehat{\omega}_k - U_k\omega_k$.
By (ii), we have $\xi_k = U_k b_k + \eta_k$. 
\Cref{prop:irl} gives $\Sigma_k(\omega_k - \omega^\star) = 0$, and $\Sigma_k$ and $U_k$ have the same kernel by construction.
Therefore, we have $U_k\omega_k = U_k\omega^\star$, and hence $\xi_k = U_k(\widehat{\omega}_k - \omega^\star)$. 
The derivation in \Cref{apdx:cor:conv}, which uses only (iv), applies verbatim and yields $\Delta_\omega^K = \bar{U}_K^{-} M_K$ with $M_K = \sum_{k=1}^{K} \xi_k$.
Because $\xi_k = U_k b_k + \eta_k$, we have
\begin{align*}
\Delta_\omega^K = \bar{U}_K^{-} M_K = \bar{U}_K^{-} \sum_{k=1}^{K} \xi_k = \sum_{k=1}^{K} \bar{U}_K^{-} U_k b_k + \bar{U}_K^{-} \eta_k .
\end{align*}
 
\emph{Step 2.}
Since $U_k$ is an orthogonal projection, they are all symmetric and positive semi-definite. 
Therefore, so is $\bar{U}_K = \sum_{k=1}^{K} U_k$, and we have $\textsc{Range}(U_k) \subseteq \textsc{Range}(\bar{U}_K)$.
It indicates that $U_k x \in \textsc{Range}(\bar{U}_K)$ holds for any $x \in \R^d$.
Equivalently, for $x \in \textsc{Range}(\bar{U}_K)$, we have
\begin{equation}\label{eq:pinv-bound}
\norm{\bar{U}_K^{-} x}_2 \leq \frac{\norm{x}_2}{\sigma_{\min}(\bar{U}_K)} .
\end{equation}

\emph{Step 3.} 
The sequence $(\eta_k)_k$ defined in \eqref{eq:eta-form} is by definition a Martingale difference sequence under the filtration $(\mathcal{F}_k)_k$. 
That is, by the tower property, $\eta_k$ and $\eta_j$ ($j < k$) are uncorrelated. 
Then,
\begin{align*}
\E\left[\norm*{\sum_{k=1}^{K} \eta_k}_2^2 \right] 
&= \E\left[\sum_{k=1}^{K} \sum_{j=1}^{K} \eta_j^\top \eta_k\right]  
= \sum_{k=1}^{K} \sum_{j=1}^{K} \E\left[\eta_j^\top \eta_k\right]  
= \sum_{k=1}^{K} \E\left[\eta_k^\top \eta_k\right]  \\
&= \sum_{k=1}^{K} \E\left[\mathrm{Tr}\left(\Cov[U_k\widehat{\omega}_k \mid \mathcal{F}_{k-1}]\right)\right]
\leq C d K ,
\end{align*}
which proves \eqref{eq:adaptive-noise}.
 
\emph{Step 4.} Since $U_k$ is an orthogonal projection, Condition (ii) gives $\norm{U_k b_k}_2 \leq \norm{b_k}_2 \leq b$.
Applying the triangle inequality, we have $\norm{\sum_{k=1}^{K} U_k b_k}_2 \leq K b$.
Because $U_k b_k \in \textsc{Range}(\bar{U}_K)$ and so is their sum, i.e.\ $\sum_{k=1}^{K} U_k b_k\in \textsc{Range}(\bar{U}_K)$, 
applying \eqref{eq:pinv-bound} yields \eqref{eq:adaptive-bias}.
 
\emph{Step 5.} Let
$A_K = \{\sigma_{\min}(\bar{U}_K) \geq \sqrt{K} a_K\}$, so $\Pr[A_K] \to 1$ by
assumption. On $A_K$, \eqref{eq:pinv-bound} bounds
$\norm{\Delta_\omega^{K,\mathrm{noise}}}_2$ pointwise by
$\norm{\sum_k \eta_k}_2 / (\sqrt{K} a_K)$, so for every $\varepsilon > 0$,
\begin{equation*}
\Pr\left[ \norm{\Delta_\omega^{K,\mathrm{noise}}}_2 > \varepsilon \right]
\leq \Pr[A_K^{\mathrm{c}}]
 + \Pr\left[ \norm{\textstyle\sum_k \eta_k}_2 > \varepsilon\sqrt{K} a_K \right]
\leq \Pr[A_K^{\mathrm{c}}] + \frac{C d}{\varepsilon^2 a_K^2}
\longrightarrow 0 ,
\end{equation*}
by Markov's inequality and \eqref{eq:adaptive-noise}, since $\Pr[A_K] \to 1$
and $a_K \to \infty$. Likewise, on $\{\sigma_{\min}(\bar{U}_K) \geq \alpha K\}$
the bound \eqref{eq:adaptive-bias} gives
$\norm{\Delta_\omega^{K,\mathrm{bias}}}_2 \leq b/\alpha$ pointwise, so
$\Pr[\norm{\Delta_\omega^{K,\mathrm{bias}}}_2 > b/\alpha]
\leq \Pr[\sigma_{\min}(\bar{U}_K) < \alpha K] \to 0$. Finally, if $b = 0$ then
$\Delta_\omega^K = \Delta_\omega^{K,\mathrm{noise}}$, and the last claim
follows from the first limit.
\end{proof}
 
\begin{remark}
Independence enters \Cref{cor:conv} only through the cross terms of Step~3.
What the argument needs is that the centered terms are pairwise uncorrelated,
and the martingale difference property \eqref{eq:eta-form} supplies this
without independence. The two proofs differ in one further respect.
\Cref{cor:conv} treats $\{\theta_k\}$ as a given sequence, so $\bar{U}_K$ is
deterministic and the covariance bound
$\Cov[\Delta_\omega^K] \preccurlyeq C K (\bar{U}_K^{-})^2$ can be stated
directly. Here $\bar{U}_K$ is random and statistically coupled with $M_K$,
which is why Step~5 conditions on a spectral event instead.
\end{remark}
 
\begin{remark}\label{rem:measurability}
Condition (i) is what restricts the scope of \Cref{cor:adaptive}. In \Cref{cor:conv}, $\{\theta_k\}$ is a given sequence and $U_k$ is deterministic, so the question of which follower model $\Sigma_k$ is evaluated at does not arise, and our implementation uses the most direct approximation of $\Sigma^{\theta_k\dagger}$, namely the surrogate model built from $\widehat{\omega}_k$. Once the sequence is adaptively generated, (i) requires $\Sigma_k$ to be computed from $\theta_k$ together with the reward estimate available at the end of iteration $k-1$, for instance the surrogate model built from $\bar{\omega}_{k-1}$, or from a trajectory subset disjoint from the one used for $\widehat{\omega}_k$. 
\Cref{cor:adaptive} therefore covers these variants rather than \Cref{algo:proposal} as implemented. Otherwise a term $\Cov(U_k, \widehat{\omega}_k - \omega_k \mid \mathcal{F}_{k-1})$ survives even when $b_k = 0$. 
This is a requirement introduced by the stronger analysis rather than a defect of the original formulation, and since $U_k$ depends only on the eigenspace partition of $\Sigma_k$ at the threshold $\epsilon$, the substitution is expected to leave $U_k$ unchanged whenever the spectral gap at $\epsilon$ exceeds the estimation perturbation.
\end{remark}
 
\begin{remark}
The spectral hypotheses also differ from that of \Cref{cor:conv}, which asks
for $\sigma_{\max}(\bar{U}_K^{-}) = o(K^{-1/2})$. The condition
$\sigma_{\min}(\bar{U}_K) \geq \alpha K$ needed for the bias term is strictly
stronger, and coincides with the persistent excitation condition familiar from
adaptive control and system identification, where it is either assumed of the
input signal or enforced by injecting a dither signal. Empirically,
$\mathrm{cond}(\bar{U}_K)$ remains bounded in $[7.3, 103.1]$ across three seeds
over $400$ outer iterations, which together with
$\sigma_{\max}(\bar{U}_K) = \Theta(K)$ implies linear growth of
$\sigma_{\min}(\bar{U}_K)$.
\end{remark}

\subsection{\Cref{prop:bound}}\label{apdx:prop:bound}
\begin{proof}
First, we show that $\norm{r_{\widehat{\omega}} - r_{\omega^\star} - c}_\infty \leq L \norm{\Delta_\omega}_2$. 
It is straightforward from the linear reward structure: for any $(s, b) \in \mathcal{S} \times \mathcal{B}$,
\begin{align}
\abs{r_{\widehat{\omega}}(s, b) - r_{\omega^\star}(s, b) - c}
&= 
\abs{\inner{\widehat{\omega}}{\phi(s, b)} - \inner{\omega^\star}{\phi(s, b)} - c \inner{\Delta_\mathrm{const}}{\phi(s, b)}}
\\
&= 
\abs{\inner{\widehat{\omega} - (\omega^\star + c \Delta_\mathrm{const})}{\phi(s, b)}}
\\
&= 
\abs{\inner{\Delta_\omega}{\phi(s, b)}}
\\
&\leq 
\norm{\Delta_\omega}_{2} \norm{\phi(s, b)}_{2}
\\
&\leq 
L \norm{\Delta_\omega}_{2} .
\end{align}

Next, we show that $\norm{\widehat{V} - V_c^\star}_{\infty} \leq (1 - \gamma_F)^{-1}\norm{r_{\widehat{\omega}} - (r_{\omega^\star} + c)}_\infty$, where $V_c^\star$ is the optimal value function corresponding to $r_{\omega^\star} + c$.
Let $\mathcal{T}_{r}$ be the soft Bellman operator corresponding to reward function $r$, defined as
\begin{equation}
\mathcal{T}_r V = \log \left( \int \exp\left( r(s, b) + \gamma_F \E_{s'}[ V(s') ] \right) \mathrm{d}b \right),
\end{equation}
where $\E_{s'}$ is the expectation taken for the next state $s'$ given the current state $s$ and the action $b$. It has been shown that the soft Bellman operator is $\gamma_F$ contraction \citep{softQ}. 

Because $r_{\widehat{\omega}}(s, b) \leq r_{\omega^\star}(s, b) + c + \norm{r_{\widehat{\omega}} - (r_{\omega^\star} + c)}_\infty$ for all $(s, b) \in \mathcal{S} \times \mathcal{B}$, we have
\begin{align}
\mathcal{T}_{r_{\widehat{\omega}}} V 
&= \log \left( \int \exp\left( r_{\widehat{\omega}}(s, b) + \gamma_F \E_{s'}[ V(s') ] \right) \mathrm{d}b \right) \\
&\leq \log \left( \int \exp\left( r^\star(s, b) + c + \norm{r_{\widehat{\omega}} - (r_{\omega^\star} + c)}_\infty +\gamma_F \E_{s'}[ V(s') ] \right) \mathrm{d}b \right) \\
&= \norm{r_{\widehat{\omega}} - (r_{\omega^\star} + c)}_\infty + \log \left( \int \exp\left( r^\star(s, b) + c + \gamma_F \E_{s'}[ V(s') ] \right) \mathrm{d}b \right) \\
&= \norm{r_{\widehat{\omega}} - (r_{\omega^\star} + c)}_\infty + \mathcal{T}_{r^\star + c} V .
\end{align}
Analogously, by using the other side of the inequality, $r_{\widehat{\omega}}(s, b) \geq r_{\omega^\star}(s, b) + c - \norm{r_{\widehat{\omega}} - (r_{\omega^\star} + c)}_\infty$, we have $\mathcal{T}_{r_{\widehat{\omega}}} V \geq -\norm{r_{\widehat{\omega}} - (r_{\omega^\star} + c)}_\infty + \mathcal{T}_{r^\star + c} V $. Therefore, for any $V$, we have 
\begin{equation}
\norm{
\mathcal{T}_{r_{\widehat{\omega}}} V - 
\mathcal{T}_{r^\star + c} V}_{\infty}
\leq \norm{r_{\widehat{\omega}} - (r_{\omega^\star} + c)}_\infty.
\end{equation}

Because soft Bellman operators are contraction maps, $V_c^\star$ and $\widehat{V}$ are the unique fixed points of $\mathcal{T}_{r_{\omega^\star} + c}$ and $\mathcal{T}_{r_{\widehat{\omega}}}$, respectively. 
Therefore, we have
\begin{align}
\norm{\widehat{V} - V_c^\star}_{\infty}
&=
\norm{\mathcal{T}_{r_{\widehat{\omega}}}\widehat{V} - \mathcal{T}_{r_{\omega^\star}+c} V_c^\star}_{\infty} \\
&=
\norm{(\mathcal{T}_{r_{\widehat{\omega}}}\widehat{V} - \mathcal{T}_{r_{\omega^\star + c}}\widehat{V}) + (\mathcal{T}_{r_{\omega^\star + c}}\widehat{V} - \mathcal{T}_{r_{\omega^\star}+c} V_c^\star)}_{\infty} \\
&\leq
\norm{\mathcal{T}_{r_{\widehat{\omega}}}\widehat{V} - \mathcal{T}_{r_{\omega^\star + c}}\widehat{V}}_\infty + \norm{\mathcal{T}_{r_{\omega^\star + c}}\widehat{V} - \mathcal{T}_{r_{\omega^\star}+c} V_c^\star}_{\infty} \\
&\leq
\norm{r_{\widehat{\omega}} - (r_{\omega^\star} + c)}_\infty 
+ \gamma_F \norm{\widehat{V} -  V_c^\star}_{\infty} .
\end{align}
Rearranging the right-most side, we have
\begin{align}
\norm{\widehat{V} - V_c^\star}_{\infty}
\leq (1 - \gamma_F)^{-1}\norm{r_{\widehat{\omega}} - (r_{\omega^\star} + c)}_\infty
.
\end{align}

The right-hand side of the above inequality is derived by using $\norm{r_{\widehat{\omega}} - r_{\omega^\star} - c}_\infty \leq L \norm{\Delta_\omega}_2$. 
The left-hand side is derived by using the fact that if the optimal soft value function corresponding to $r$ is $V_r$, then the optimal soft value function corresponding to $r + c$ is $V_{r + c} = V_{r} + c (1 - \gamma_F)^{-1}$. Hence, we have $\norm{\widehat{V} - V^\star - c(1 - \gamma_F)^{-1}}_{\infty} \leq L(1 - \gamma_F)^{-1} \norm{\Delta_\omega}_2$. 

Finally, we bound the HG. 
First, we derive the bound for the Conf-MDP setting. 
Let $\nabla_\theta J_L(\theta, g^{\theta\dagger})$ be the HG derived in \eqref{eq:hg_configurable}, where the ground-truth value function of the follower is used. 
Let $\widehat{\nabla_\theta J_L(\theta, g^{\theta\dagger})}$ be the HG computed using the estimated value function of the follower, $\widehat{V}$. Then, it is easy to see that only the term $\nabla_\theta Q_F^{\theta\dagger}$ depends on the follower's value function. Therefore, 
\begin{align}
\norm*{\widehat{\nabla_\theta J_L(\theta, g^{\theta\dagger})} - \nabla_\theta J_L(\theta, g^{\theta\dagger})}
\leq \mathbb{E}_\tau \Bigg[ \sum_{t=0}^\infty \gamma_L^t \frac{\abs{B_L^{{\theta}\dagger}(s_t, b_t)}}{\beta} \norm*{ \nabla_{\theta} Q_F^{{\theta} \dagger}(s_t, b_t) - \widehat{\nabla_{\theta} Q_F^{{\theta} \dagger}(s_t, b_t)}} \Bigg],
\end{align}
where $\nabla_{\theta} Q_F^{{\theta} \dagger}(s_t, b_t)$ and $\widehat{\nabla_{\theta} Q_F^{{\theta} \dagger}(s_t, b_t)}$ are as defined in \eqref{eq:nablaQF} with the ground-truth value function $V^\star$ and its estimate $\widehat{V}$. 
Noting that for any constant $C \in \R$, 
\begin{align*}
\MoveEqLeft[2]
\mathbb{E}_\tau \Bigg[ \sum_{t=0}^\infty \gamma_F^{t+1} C \nabla_{{\theta}} \log p^{{\theta}}(s_{t+1} \mid s_t, b_t) \Bigg|\begin{array}{l}s_0=s\\b_0=b\end{array} \Bigg] 
\\
&= \sum_{t=0}^\infty \gamma_F^{t+1} \mathbb{E}_\tau \Bigg[ C \nabla_{{\theta}} \log p^{{\theta}}(s_{t+1} \mid s_t, b_t) \Bigg|\begin{array}{l}s_0=s\\b_0=b\end{array} \Bigg]
\\
&= \sum_{t=0}^\infty \gamma_F^{t+1} \mathbb{E}_\tau \Bigg[ \E_{t}[ C \nabla_{{\theta}} \log p^{{\theta}}(s_{t+1} \mid s_t, b_t)] \Bigg|\begin{array}{l}s_0=s\\b_0=b\end{array} \Bigg]
\\
&= \sum_{t=0}^\infty \gamma_F^{t+1} \mathbb{E}_\tau \Bigg[ \nabla_{{\theta}} \E_{t}[ C ] \Bigg|\begin{array}{l}s_0=s\\b_0=b\end{array} \Bigg]
\\
&= 0,
\end{align*}
where $\E_t$ denotes the conditional expectation under the natural filtration generated by $\{(s_i, b_i)\}_{i \leq t}$, we have 
\begin{equation}
\nabla_{{\theta}}Q_F^{{\theta}\dagger}(s,b) = \mathbb{E}_\tau \Bigg[ \sum_{t=0}^\infty \gamma_F^{t+1} (V_F^{{\theta}\dagger}(s_{t+1}) + C) \nabla_{{\theta}} \log p^{{\theta}}(s_{t+1} \mid s_t, b_t) \Bigg|\begin{array}{l}s_0=s\\b_0=b\end{array} \Bigg].
\end{equation}
With $C = c(1 - \gamma_F)^{-1}$, we have
\begin{align*}
\MoveEqLeft[2]
\norm*{ \nabla_{\theta} Q_F^{{\theta} \dagger}(s, b) - \widehat{\nabla_{\theta} Q_F^{{\theta} \dagger}(s, b)}}
\\
&=
\norm*{\mathbb{E}_\tau \Bigg[ \sum_{t=0}^\infty \gamma_F^{t+1} (V^\star(s_{t+1}) + c(1 - \gamma_F)^{-1} - \widehat{V}(s_{t+1})) \nabla_{{\theta}} \log p^{{\theta}}(s_{t+1} \mid s_t, b_t) \Bigg|\begin{array}{l}s_0=s\\b_0=b\end{array} \Bigg]}
\\
&\leq
\mathbb{E}_\tau \Bigg[ \sum_{t=0}^\infty \gamma_F^{t+1} \abs*{V^\star(s_{t+1}) + c(1 - \gamma_F)^{-1} - \widehat{V}(s_{t+1})} \norm*{\nabla_{{\theta}} \log p^{{\theta}}(s_{t+1} \mid s_t, b_t) }\Bigg|\begin{array}{l}s_0=s\\b_0=b\end{array} \Bigg]
\\
&\leq \norm{\widehat{V} - V_c^\star}_{\infty}
\mathbb{E}_\tau \Bigg[ \sum_{t=0}^\infty \gamma_F^{t+1} \norm*{\nabla_{{\theta}} \log p^{{\theta}}(s_{t+1} \mid s_t, b_t) }\Bigg|\begin{array}{l}s_0=s\\b_0=b\end{array} \Bigg]
\\
&\leq L(1 - \gamma_F)^{-1} \norm{\Delta_\omega}_2
\mathbb{E}_\tau \Bigg[ \sum_{t=0}^\infty \gamma_F^{t+1} \norm*{\nabla_{{\theta}} \log p^{{\theta}}(s_{t+1} \mid s_t, b_t) }\Bigg|\begin{array}{l}s_0=s\\b_0=b\end{array} \Bigg].
\end{align*}
Hence, we have
\begin{multline*}
\norm*{\widehat{\nabla_\theta J_L(\theta, g^{\theta\dagger})} - \nabla_\theta J_L(\theta, g^{\theta\dagger})}
\\
\leq 
\frac{L \norm{\Delta_\omega}_2}{1 - \gamma_F} 
\mathbb{E}_\tau \Bigg[ \sum_{t=0}^\infty \gamma_L^t \frac{\abs{B_L^{{\theta}\dagger}(s_t, b_t)}}{\beta} \mathbb{E}_{\tilde{\tau}} \Bigg[ \sum_{\tilde{t}=0}^\infty \gamma_F^{\tilde{t}+1} \norm*{\nabla_{{\theta}} \log p^{{\theta}}(\tilde{s}_{\tilde{t}+1} \mid \tilde{s}_{\tilde{t}}, \tilde{b}_{\tilde{t}}) }\Bigg|\begin{array}{l}\tilde{s}_0=s_t\\\tilde{b}_0=b_t\end{array} \Bigg]\Bigg].
\end{multline*}
This completes the proof for the Conf-MDP case.

Next, we derive the bound for 2-player MGs.
Similarly to the above argument, we have from \eqref{eq:hg_MG}
\begin{multline}
\norm*{\widehat{\nabla_\theta J_L(\theta, g^{\theta\dagger})} - \nabla_\theta J_L(\theta, g^{\theta\dagger})}
\\
\leq \frac{1}{1 - \gamma_L}\norm*{\mathbb{E} \Bigg[  \frac{B_L^{{\theta}\dagger}(s, a, b)}{\beta} \left(\nabla_{\theta} Q_F^{{\theta} \dagger}(s, a, b) - \widehat{\nabla_{\theta} Q_F^{{\theta} \dagger}(s, a, b)}\right) \Bigg]},
\end{multline}
where $\nabla_{\theta} Q_F^{{\theta} \dagger}(s, a, b)$ and $\widehat{\nabla_{\theta} Q_F^{{\theta} \dagger}(s, a, b)}$ are as defined in \eqref{eq:nablaQF_MG} with the ground-truth value function $V^\star$ and its estimate $\widehat{V}$, and the expectation is taken for $(s,a,b) \sim g^{{\theta}\dagger}(b\mid s,a)f_{{\theta}}(a\mid s) d_{\gamma_L}^{\theta\dagger}(s)$. 
Noting that for any constant $C \in \R$, 
\begin{align*}
\MoveEqLeft[2]
\mathbb{E}_\tau \Bigg[ \sum_{t=0}^\infty \gamma_F^t C\nabla_{{\theta}}\log f_{{\theta}}(a_t\mid s_t)\Bigg|\begin{array}{l}s_0=s\\a_0=a\\b_0=b\end{array} \Bigg] 
\\
&= \sum_{t=0}^\infty \gamma_F^{t+1} \mathbb{E}_\tau \Bigg[ C\nabla_{{\theta}}\log f_{{\theta}}(a_t\mid s_t)\Bigg|\begin{array}{l}s_0=s\\a_0=a\\b_0=b\end{array} \Bigg]
\\
&= C\nabla_{{\theta}}\log f_{{\theta}}(a\mid s) + \sum_{t=1}^\infty \gamma_F^{t+1} \mathbb{E}_\tau \Bigg[ \E_{t}[ C\nabla_{{\theta}}\nabla_\theta\log f_{{\theta}}(a_t\mid s_t)\Bigg|\begin{array}{l}s_0=s\\a_0=a\\b_0=b\end{array}  \Bigg]
\\
&= C\nabla_{{\theta}}\log f_{{\theta}}(a\mid s) +  \sum_{t=1}^\infty \gamma_F^{t+1} \mathbb{E}_\tau \Bigg[ \nabla_{{\theta}} \E_{t}[ C ] \Bigg|\begin{array}{l}s_0=s\\a_0=a\\b_0=b\end{array} \Bigg]
\\
&= C\nabla_{{\theta}}\log f_{{\theta}}(a\mid s) ,
\end{align*}
where $\E_t$ denotes the conditional expectation under the natural filtration generated by $\{(s_i, a_i, b_i)\}_{i \leq t}$, we have 
\begin{equation}
\nabla_{{\theta}}Q_F^{{\theta}\dagger}(s,a,b) = \mathbb{E}_\tau \Bigg[ \sum_{t=0}^\infty \gamma_F^{t+1} (V_F^{{\theta}\dagger}(s_{t+1}, a_{t+1}) + C) \nabla_\theta\log f_{{\theta}}(a_t\mid s_t)\Bigg|\begin{array}{l}s_0=s\\a_0=a\\b_0=b\end{array} \Bigg] - C\nabla_{{\theta}}\log f_{{\theta}}(a\mid s) .
\end{equation}
Noting that
\begin{align}
\mathbb{E} \Bigg[  \frac{B_L^{{\theta}\dagger}(s, a, b)}{\beta} C\nabla_{{\theta}}\log f_{{\theta}}(a\mid s) \Bigg]
= \mathbb{E} \Bigg[  \frac{\E_{b \sim g^{\theta\dagger}} [B_L^{{\theta}\dagger}(s, a, b)]}{\beta} C\nabla_{{\theta}}\log f_{{\theta}}(a\mid s) \Bigg]
= 0,
\end{align}
we have with $C = c(1 - \gamma_F)^{-1}$, 
\begin{align*}
\MoveEqLeft[0]
\norm*{\mathbb{E} \Bigg[  \frac{B_L^{{\theta}\dagger}(s, a, b)}{\beta} \left(\nabla_{\theta} Q_F^{{\theta} \dagger}(s, a, b) - \widehat{\nabla_{\theta} Q_F^{{\theta} \dagger}(s, a, b)}\right) \Bigg]}
\\
&=
\norm*{\mathbb{E} \Bigg[  \frac{B_L^{{\theta}\dagger}(s, a, b)}{\beta} \mathbb{E}_\tau \Bigg[ \sum_{t=0}^\infty \gamma_F^{t+1} (V^\star(s_{t+1}) + \frac{c}{1 - \gamma_F} - \widehat{V}(s_{t+1})) \nabla_\theta\log f_{{\theta}}(a_t\mid s_t)\Bigg|\begin{array}{l}s_0=s\\a_0=a\\b_0=b\end{array}\Bigg] \Bigg]}
\\
&\leq
\mathbb{E} \Bigg[  \frac{\abs{B_L^{{\theta}\dagger}(s, a, b)}}{\beta} \mathbb{E}_\tau \Bigg[ \sum_{t=0}^\infty \gamma_F^{t+1} \abs{V^\star(s_{t+1}) + \frac{c}{1 - \gamma_F} - \widehat{V}(s_{t+1})} \norm{\nabla_\theta\log f_{{\theta}}(a_t\mid s_t)} \Bigg|\begin{array}{l}s_0=s\\a_0=a\\b_0=b\end{array}\Bigg] \Bigg]
\\
&\leq
\norm{\widehat{V} - V_c^\star}_\infty 
\mathbb{E} \Bigg[  \frac{\abs{B_L^{{\theta}\dagger}(s, a, b)}}{\beta} \mathbb{E}_\tau \Bigg[ \sum_{t=0}^\infty \gamma_F^{t+1} \norm{\nabla_\theta\log f_{{\theta}}(a_t\mid s_t)} \Bigg|\begin{array}{l}s_0=s\\a_0=a\\b_0=b\end{array}\Bigg] \Bigg].
\end{align*}
The rest follows the same argument as the Conf-MDP case. This completes the proof.
\end{proof}

\section{Detail Implementation}

\subsection{Hypergradient derived in \citet{Kudo2026icaps}}\label{apdx:hg}

The HG in Conf-MDP settings can be explicitly written as
\begin{multline}
    \nabla_{{\theta}}J_L({\theta}, g^{\theta\dagger})
    = \mathbb{E}_\tau \Bigg[ \sum_{t=0}^\infty \gamma_L^t \bigg( \nabla_{\theta} r_L^{\theta}(s_t, b_t)
    + V_L^{{\theta}\dagger}(s_t) \nabla_{\theta} \log p^{\theta} (s_{t} \mid s_{t-1}, b_{t-1}) \\
    + \frac{B_L^{{\theta}\dagger}(s_t, b_t)}{\beta} \nabla_{\theta} Q_F^{{\theta} \dagger}(s_t, b_t) \bigg) \Bigg],
\label{eq:hg_configurable}
\end{multline}
where $p^{\theta}(s_0 \mid s_{-1}, b_{-1})$ is interpreted as the initial distribution $\rho_0^{\theta}(s_0)$. 

In the context of 2-player MGs, the benefit of the leader $f_{{\theta}}$, given that the follower takes action $b$ at state $s$ under the leader's action $a$, is defined as:
\begin{equation*}
    B_L^{{\theta}\dagger}(s, a, b) := Q_L^{{\theta}\dagger}(s,a,b)-\E_{b\sim g^{{\theta}\dagger}(\cdot\mid s,a)}\left[Q_L^{{\theta}\dagger}(s,a,b)\right].
\end{equation*}

The HG for 2-player MG with a stochastic leader policy is derived as follows. In the following, $d_{\gamma_L}^{\theta \dag}(s)$ denotes the discounted state visitation distribution under the leader's policy $f_\theta$ and its corresponding best response follower's policy $g^{\theta\dagger}$.

\citet{Kudo2026icaps} show that the HG of \eqref{eq:leader-utility-MG} with respect to ${\theta}$ is given by
\begin{equation}
    \nabla_{{\theta}}J_L({\theta}, g^{\theta\dagger})
    =\dfrac{1}{1-\gamma_L}\E\bigg[Q_L^{{\theta}\dagger}(s,a,b)\nabla_{{\theta}}\log f_{{\theta}}(a\mid s)
   +\dfrac{B_L^{{\theta}\dagger}(s,a,b)}{\beta}\nabla_{\theta}Q_F^{\theta\dagger}(s,a,b)\bigg],\label{eq:hg_MG}
\end{equation}
where the expectation $\E$ is taken for $(s,a,b) \sim g^{{\theta}\dagger}(b\mid s,a)f_{{\theta}}(a\mid s) d_{\gamma_L}^{\theta\dagger}(s)$, and
\begin{equation}
    \nabla_{\theta}Q_F^{\theta\dagger}(s,a,b) =\E_{\tau}\Bigg[\sum_{t=0}^\infty \gamma_F^t V_F^{{\theta}\dag}(s_t,a_t)\nabla_{{\theta}}\log f_{{\theta}}(a_t\mid s_t)\Bigg|\begin{array}{l}s_0=s\\a_0=a\\b_0=b\end{array}\Bigg].\label{eq:nablaQF_MG}
\end{equation}

\subsection{Hypergradient Baseline}\label{apdx:hgbl}
\subsubsection{Hypergradient Baselines for Conf-MDP}\label{apdx:hgbl_conf}

Based on the HG formulation derived in Appendix \ref{apdx:hg}, we introduce baseline functions to reduce the variance of the gradient estimators. For the Conf-MDP setting in Equation \eqref{eq:hg_configurable}, we subtract $\nabla_{\theta} V_F^{{\theta} \dagger}(s_t, b_t)$ from $\nabla_{\theta} Q_F^{{\theta} \dagger}(s_t, b_t)$. Because $\nabla_{\theta} V_F$ is independent of the action $b$, it effectively serves as an unbiased baseline when marginalized over the action distribution:
\begin{multline}
 \nabla_{{\theta}}J_L({\theta}, g^{\theta\dagger})
 = \mathbb{E}_\tau \Bigg[ \sum_{t=0}^\infty \gamma_L^t \bigg( \nabla_{\theta} r_L^{\theta}(s_t, b_t)
 + V_L^{{\theta}\dagger}(s_t) \nabla_{\theta} \log p^{\theta} (s_{t} \mid s_{t-1}, b_{t-1}) \\
 + \frac{B_L^{{\theta}\dagger}(s_t, b_t)}{\beta}( \nabla_{\theta} Q_F^{{\theta} \dagger}(s_t, b_t)-\nabla_{\theta} V_F^{{\theta} \dagger}(s_t) )\bigg) \Bigg],
\label{eq:hg_configurable_baseline}
\end{multline}
where we have $\nabla_{\theta} V_F^{{\theta} \dagger}(s_t) = \E_{g^{\theta\dagger}}[\nabla_{\theta} Q_F^{{\theta} \dagger}(s_t, b_t) \mid s_t]$. 
Note that this formulation is not used in \citet{Kudo2026icaps} because estimating both $\nabla_{\theta} Q_F^{{\theta} \dagger}(s_t, b_t)$ and its expectation $\nabla_{\theta} V_F^{{\theta} \dagger}(s_t)$ requires multiple trajectories starting from the same state $s_t$. In our setting, we can easily approximate $\nabla_{\theta} V_F^{{\theta} \dagger}(s_t)$ by using the surrogate follower $\widehat{g}$. 

To further reduce variance, an alternative formulation incorporating a temporal difference-like baseline can be employed by subtracting $V_L^{{\theta}\dagger}(s_{t+1})$ from $V_L^{{\theta}\dagger}(s_t)$:
\begin{multline}
 \nabla_{{\theta}}J_L({\theta}, g^{\theta\dagger})
\approx \mathbb{E}_\tau \Bigg[ \sum_{t=0}^\infty \gamma_L^t \bigg( \nabla_{\theta} r_L^{\theta}(s_t, b_t)
+ (V_L^{{\theta}\dagger}(s_t) - \mathbb{E} [V_L^{{\theta}\dagger}(s_{t+1}))] \nabla_{\theta} \log p^{\theta} (s_{t} \mid s_{t-1}, b_{t-1}) \\
+ \frac{B_L^{{\theta}\dagger}(s_t, b_t)}{\beta}( \nabla_{\theta} Q_F^{{\theta} \dagger}(s_t, b_t)-\nabla_{\theta} V_F^{{\theta} \dagger}(s_t) )\bigg) \Bigg].
\label{eq:hg_configurable_biased}
\end{multline}
However, it is important to note that this specific TD-like term does not act as a strict baseline, making the resulting gradient estimator biased rather than unbiased.

\subsubsection{Hypergradient Baselines for 2-player MG}\label{apdx:hgbl_mg}

Similarly, for the 2-player MG setting derived in \eqref{eq:hg_MG}, we apply a baseline to reduce variance by subtracting $\nabla_{\theta}V_F^{\theta\dagger}(s,a)$ from $\nabla_{\theta}Q_F^{\theta\dagger}(s,a,b)$:
\begin{multline}
  \nabla_{{\theta}}J_L({\theta}, g^{\theta\dagger})
 \\
 =\dfrac{1}{1-\gamma_L}\mathbb{E}\bigg[Q_L^{{\theta}\dagger}(s,a,b)\nabla_{{\theta}}\log f_{{\theta}}(a\mid s)
+\dfrac{B_L^{{\theta}\dagger}(s,a,b)}{\beta}(\nabla_{\theta}Q_F^{\theta\dagger}(s,a,b)-\nabla_{\theta}V_F^{\theta\dagger}(s,a))\bigg].\label{eq:hg_MG_baseline}
\end{multline}
The baseline formulation utilized in this setting is strictly unbiased, meaning it yields an unbiased gradient estimator.

\subsection{Reward Identification Algorithm}

\begin{algorithm}%
\caption{Reward Identification}
\label{alg:bilevel_reward_identification}
\begin{algorithmic}[1]
\REQUIRE $\omega_k$, $\Sigma_k$,$v_{k-1}$, $\bar{U}_{k-1}$, threshold parameter $\epsilon \geq 0$.
\STATE Perform eigen decomposition $E \Lambda E^\top = \Sigma_k$
\STATE Compute a diagonal matrix $I_{\Lambda\succ \epsilon}$, where $[I_{\Lambda\succ \epsilon}]_{dd} = 1$ if $[\Lambda]_{dd} > \epsilon$, $0$ otherwise
\STATE Compute the scaled covariance matrix $U_k = E I_{\Lambda\succ \epsilon} E^\top$
\STATE $\bar{U}_{k} \gets \bar{U}_{k-1} + U_k$, 
$v_{k} \gets v_{k-1} + U_k \omega_k$
\STATE Solve $\bar{U}_k \omega = v_{k}$ to obtain $\bar{\omega}_k$
\ENSURE $\bar{\omega}_k$, $v_k$, $\bar{U}_k$
\end{algorithmic}
\end{algorithm}

\subsection{Modules within Proposed Algorithm~\ref{algo:proposal}}
Note that $\bar{s}$ denotes state $s$ in Conf-MDP settings and denotes a pair of state $s$ and leader action $a$ in 2-player MGs.

\begin{algorithm}[H]
\caption{$\texttt{CriticUpdate}(Q_L^i, \mathcal{R})$ (Expected SARSA variant)}
\label{algo:critic_update_expected_sarsa}
\begin{algorithmic}[1]
\REQUIRE{Current critic $Q_L^i$, policies $f, \widehat{g}$, batch $\mathcal{R}$, learning rate $\alpha$, discount factor $\gamma_L$}
\FOR{$(\bar{s}, b, r_L, \bar{s}') \in \mathcal{R}$}
\STATE{Compute expected value: $V_L(\bar{s}') \gets \sum_{a'} f(a'|s') \sum_{b'} \widehat{g}(b'|\bar{s}') Q_L^i(\bar{s}', b')$}
\STATE{$y \gets r_L + \gamma_L V_L(\bar{s}')$}
\STATE{$Q_L^i(\bar{s}, b) \gets Q_L^i(\bar{s}, b) + \alpha \left( y - Q_L^i(\bar{s}, b) \right)$}
\ENDFOR
\STATE{$Q_L^{i+1} \gets Q_L^i$}
\ENSURE{$Q_L^{i+1}$}
\end{algorithmic}
\end{algorithm}

\begin{algorithm}[H]
\caption{$\texttt{CriticUpdate}$ (TD with target network)}
\label{algo:critic_update_td_target}
\begin{algorithmic}[1]
\REQUIRE
Online critic $Q_{\omega_L}$, target critic $Q_{\bar{\omega}_L}$,
batch $\mathcal{R}$,
discount $\gamma_L$, learning rate $\alpha$, Polyak factor $\tau$
\FOR{$(\bar{s},b,r_{L},\bar{s'},b',d)\in \mathcal{R}$}
  \STATE $y \gets r_{L} + \gamma_L(1-d)\,Q_{\bar{\omega}_L}(\bar{s'},b')$
\ENDFOR
\STATE
$\displaystyle
\omega_L \gets \omega_L - \alpha\nabla_{\omega_L}
\frac{1}{N}\sum_{j=1}^{N}\Bigl(Q_{\omega_L}(\bar{s_j},b_j)-y_j\Bigr)^2$
\STATE
$\displaystyle
\bar{\omega}_L \gets (1-\tau)\bar{\omega}_L + \tau\,\omega_L
\quad\text{(Polyak averaging)}$
\ENSURE Updated critics $Q_{\omega_L}, Q_{\bar{\omega}_L}$.
\end{algorithmic}
\end{algorithm}

\begin{algorithm}[H]
\caption{$\texttt{PartialDerivative}(\mathcal{R},\theta^i,V_L,\cdot)$ in Conf-MDP Settings}
\label{algo:estimate_partial_grad_Conf-MDP}
\begin{algorithmic}[1]
\REQUIRE{Trajectory batch $\mathcal{R}$, current leader parameter $\theta^i$, leader value function $V_L$, initial state distribution $\rho_0$}
\STATE{$\begin{aligned}
\widehat{\partial_{\theta}J_L} \gets \frac{1}{|\mathcal{R}|} \sum_{\tau\in \mathcal{R}} \Biggl( &V_L(s_0)\nabla_{\theta}\log\rho_0(s_0) + \\
&\sum_{t=0}^{T-1}\left( \gamma_L^{t}\nabla_{\theta}r_L^{\theta^i}(s_t,b_t) + \gamma_L^{t+1}V_L(s_{t+1}) \nabla_{\theta}\log p^{\theta^i}(s_{t+1}\mid s_t,b_t)\right) \Biggr)
\end{aligned}$}
\ENSURE{$\widehat{\partial_{\theta}J_L}$}
\end{algorithmic}
\end{algorithm}

\begin{algorithm}[H]
\caption{$\texttt{PartialDerivative}(\mathcal{R},\theta^i,\cdot,Q_L)$ in Markov Game Settings}
\label{algo:estimate_partial_grad_2mdp}
\begin{algorithmic}[1]
\REQUIRE{Transition sample batch $\mathcal{R}$, current leader parameter $\theta^i$, leader Q-function $Q_L$, leader policy $f_{\theta^i}$, surrogate follower policy $\widehat g$}
\STATE{$V_L(s)\gets \sum_{a'} f_{\theta^i}(a'|s)\sum_{b'} \widehat g(b'|s,a')\,Q_L(s,a',b')$}
\STATE{$\widehat{\partial_{\theta}J_L}\gets\dfrac{1}{|\mathcal{R}|}\sum_{(s,a,b)\in \mathcal{R}}\bigl(Q_L(s,a,b)-V_L(s)\bigr)\nabla_{\theta}\log f_{\theta^i}(a\mid s)$} \COMMENT{Empirical approximation of \cref{eq:hg_MG} without discount factor}
\ENSURE{$\widehat{\partial_{\theta}J_L}$}
\end{algorithmic}
\end{algorithm}

\begin{algorithm}[H]
\caption{$\texttt{FollowerQGrad}(s, a, b, \mathcal{R}, \theta^i, V_F)$ in Markov Game Settings}
\label{alg:QF-grad}
\begin{algorithmic}[1]
\REQUIRE{Target state $s$, leader action $a$, follower action $b$, full trajectory batch $\mathcal{R}$, current leader parameter $\theta^i$, follower soft value $V_F$}
\STATE{Extract segments: $\mathcal{R}_{sab} \gets \left\{\tau^{k:T} \mid (s_k, a_k, b_k) = (s,a,b) \text{ for } (s_k,a_k,b_k)\in\tau\in \mathcal{R}\right\}$}
\STATE{$\widehat{\nabla_{\theta}Q_F}(s,a,b)\!\gets\!\frac{1}{|\mathcal{R}_{sab}|}\!\sum_{\tau^{k:T}\in\mathcal{R}_{sab}}\sum_{t=k+1}^{T}\gamma_F^{\,t-k}\!~ V_F(s_t,a_t)$}
\ENSURE{$\widehat{\nabla_{\theta}Q_F}(s,a,b)$}
\end{algorithmic}
\end{algorithm}

\begin{algorithm}[H]
\caption{Influence Term}
\label{algo:influence}
\begin{algorithmic}[1]
    \STATE{$\widehat{\nabla_{\theta_k}J_L^k} \gets \widehat{\partial_{\theta_k}J_L^k} + \frac{1}{|\mathcal{R}_k|}\sum_{(s,a,b)\in \mathcal{R}_k} B_L^{k+1}(s,a,b) \widehat{\nabla_{\theta_k}Q_F^k}(s,a,b)$} 
\end{algorithmic}
\end{algorithm}

\begin{algorithm}[H]
\caption{Influence Term with Baseline in Markov Game Settings (Discrete Case)}
\label{algo:baseline_influence}
\begin{algorithmic}[1]
    \STATE{Compute baseline: $\widehat{\nabla_{\theta_k}V_F^k}(s, a) \gets \sum_{b'} \hat{g}_k(b'\mid s,a) \widehat{\nabla_{\theta_k}Q_F^k}(s, a, b')$}
    \STATE{$\widehat{\nabla_{\theta_k}J_L^k} \gets \widehat{\partial_{\theta_k}J_L^k} + \frac{1}{|\mathcal{R}_k|}\sum_{(s,a,b)\in \mathcal{R}_k} B_L^{k+1}(s,a,b) \left( \widehat{\nabla_{\theta_k}Q_F^k}(s,a,b) - \widehat{\nabla_{\theta_k}V_F^k}(s,a) \right)$} \COMMENT{Based on eq. \eqref{eq:hg_MG_baseline}}
\end{algorithmic}
\end{algorithm}

\section{Experiment}
\label{apdx:exp}

\subsection{Implementation Details}
Table~\ref{tab:implementation_details} summarizes the implementation choices used in each environment. While the overall algorithmic framework is shared across settings, several components are instantiated differently depending on whether the transition dynamics are known or unknown. In the Conf-MDP experiments, quantities such as $Q_F^{\theta\dagger}$, $Q_L^{\theta\dagger}$, $V_L^{\theta\dagger}$, and $\nabla Q_F^{\theta\dagger}$ are computed analytically from the known system dynamics using the Discrete Algebraic Riccati Equation (DARE) and discrete Lyapunov equations. In the 2-player Markov game experiments, surrogate policies and value-related terms are computed either by dynamic programming or by successor-feature-based estimation, depending on whether the transition model is available. For the continuous setting, the critic is first pre-trained with SARSA-TD updates from randomly initialized trajectories, and the transition dynamics are estimated by linear regression when unknown.

\begin{table}[H]
\centering
\caption{Implementation Details by Environment}
\label{tab:implementation_details}
\resizebox{\textwidth}{!}{
\begin{tabular}{llcp{8cm}}
\toprule
\textbf{Setting} & \textbf{Component} & \textbf{\Cref{algo:proposal} Line} & \textbf{Implementation Details} \\
\midrule
\textbf{Conf-MDP (Common)} 
& $Q_F^{\theta \dagger}$ & 7 & Since the transition dynamics are linear-Gaussian and the reward functions are quadratic, it is derived directly from the DARE solution. \\
& $Q_L^{\theta\dagger}, V_L^{\theta\dagger}$ & 9 & Analytically computed by solving the DARE. \\
& $\nabla Q_F^{\theta \dagger}$ & 13 & Obtained via the discrete Lyapunov equation. \\
\midrule
\textbf{Conf-MDP (Easy)} 
& HG Calculation & 15 & Computed analytically using the known transition dynamics and policies, which bypasses the need for empirical trajectory sampling. \\
\midrule
\textbf{Conf-MDP (Hard)} 
& HG Calculation & 15 & A variance reduction baseline is adopted to stabilize stochastic gradient estimates, using the expectation $\mathbb{E}[V_L^{\theta\dagger}(s_{t+1})]$ obtained via DARE as a baseline for the $\nabla_{\theta} \log p^{\theta}$ term. \\
\midrule
{\textbf{2-player MG (Discrete)}} 
& Surrogate Policy $\hat{g}_k$ & 5, 6 & Computed via Soft Value Iteration (SoftVI) if the transition probability (P) is known, or by incorporating Successor Features (SFs) to calculate expected feature values if unknown. \\
& Value functions \& gradients & 9, 10, 12, 13 & If P is known, computed using dynamic programming to eliminate sampling noise and improve learning efficiency. \\
\midrule
{\textbf{2-player MG (Continuous)}} 
& Critic ($Q_L$) Initialization & 1 & Prior to the main loop, pre-trained for 50 optimization epochs via SARSA-TD updates using 40 episodes (approx. 4,000 transitions) generated from a randomized initial distribution. \\
& Transition Dynamics & 5 & Estimated via linear regression if unknown. \\
& Benefit ($B_L$) & 12 & Calculated as the difference between the online leader's action-value function ($Q_L^{\text{online}}$) and the target leader's value function ($V_L^{\text{target}}$). \\
& $\nabla Q_F^{\theta \dagger}$ & 13 & Obtained via the discrete Lyapunov equation. \\
\bottomrule
\end{tabular}
}
\end{table}

\subsection{Synthetic task to evaluate the effect of shaped reward.}
\label{demo:syn}

We verify that a single IRL reward bias cannot achieve the true optimal objective in a simple environment due to dynamics mismatch. 

We consider a Conf-MDP $\mathcal{M}_\theta$, where the state space is $\mathcal{S} = \{s_1, s_2\}$ and the action space is $\mathcal{B} = \{b_1, b_2\}$. The true follower's reward is $r_F(s, b) = 0$ for all $(s, b)$. The transition probabilities are parameterized by $\theta \in (0, 1)$ such that $p(s_1\mid s_1, b_1) = p(s_1\mid s_2, b_2) = \theta$, $p(s_1\mid s_1, b_2) = 1$, and $p(s_1\mid s_2, b_1) = 0$.

Instead of a standard environment reward, we define a diagnostic meta-reward to explicitly evaluate the behavior of the surrogate policy.
Let the leader's reward be $r_L= -(0.5 - \hat{g}(b_t \mid s_t))^2$ and the regularization term be $\Omega(\theta)=-(\theta-0.5)^2$, where $\hat{g}$ is the estimated follower policy. The leader's 
diagnostic meta-objective is defined as:
\begin{equation}
J_L(\theta, g^{\theta\dagger}) := \mathbb{E}_\tau \left[ \sum_{t=0}^\infty \gamma_L^t r_L(s_t, b_t) \right] + \Omega(\theta).
\end{equation}

To simulate an estimated reward vector produced by IRL, we initialize the reward vector as $\mathcal{N}([a, a, 0, 0], I)$ with $a = 1.5$. By observing the gap between the estimated single reward and the aggregate rewards, we evaluate how this estimation error influences the optimization goal. The objective of this experiment is to drive the KL divergence to a uniform distribution to zero, and to converge $\theta$ to $0.5$.

Since the leader knows the true dynamics, the surrogate follower's policy is recovered using Soft Value Iteration. As shown in the~\Cref{fig:conf_discrete_rewards,fig:conf_discrete_kl_theta,fig:conf_discrete_terms} below, the estimation errors prevent the method from achieving the true optimal objective, confirming our observation.

\paragraph{Results}
The reward identifiability is not only theoretically but also practically important. 
The experimental results are shown in ~\Cref{fig:conf_discrete_rewards,fig:conf_discrete_kl_theta,fig:conf_discrete_terms}. \rehg effectively eliminates the potential shaping term, which is evidenced by the convergence of Terms 2, 3 in \eqref{eq:hg_configurable} and the estimated reward function to 0. Furthermore, the leader's parameter $\theta$ successfully converges to the target value of $0.5$. 
In contrast, the single reward baseline (Single) fails to achieve the optimal objective due to the influence of potential shaping. Specifically, the KL divergence, individual evaluation terms, and the parameters $\theta$ converge to sub-optimal values. These results empirically demonstrate that the inability to vanish the potential shaping term inevitably leads to estimation errors when a dynamics mismatch exists.

\begin{figure}[H]
  \centering
  \begin{minipage}[b]{0.49\textwidth}
    \centering
    \includegraphics[width=\textwidth]{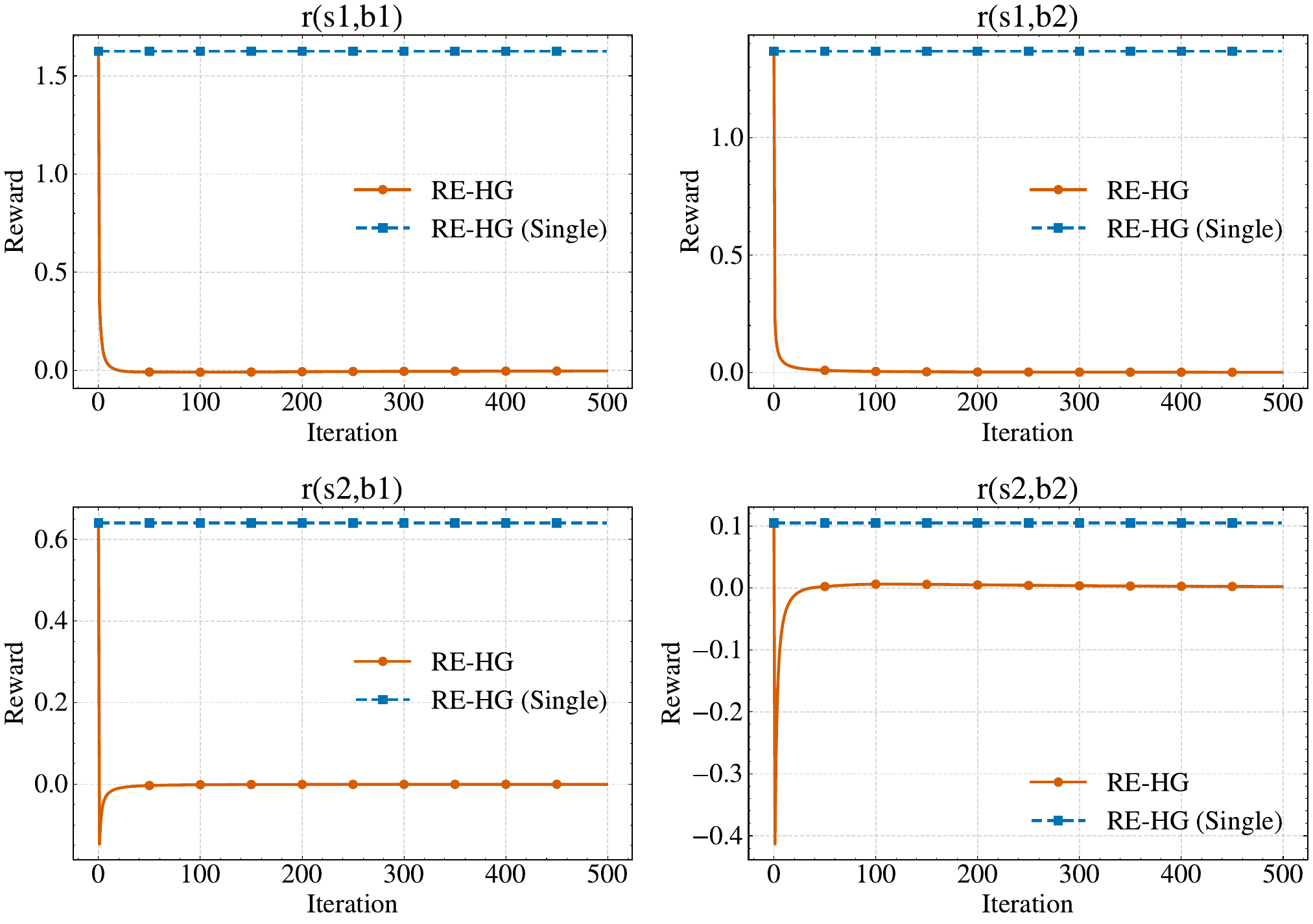}
    \caption{Estimated Rewards}
    \label{fig:conf_discrete_rewards}
  \end{minipage}
  \hfill
  \begin{minipage}[b]{0.49\textwidth}
    \centering
    \includegraphics[width=\textwidth]{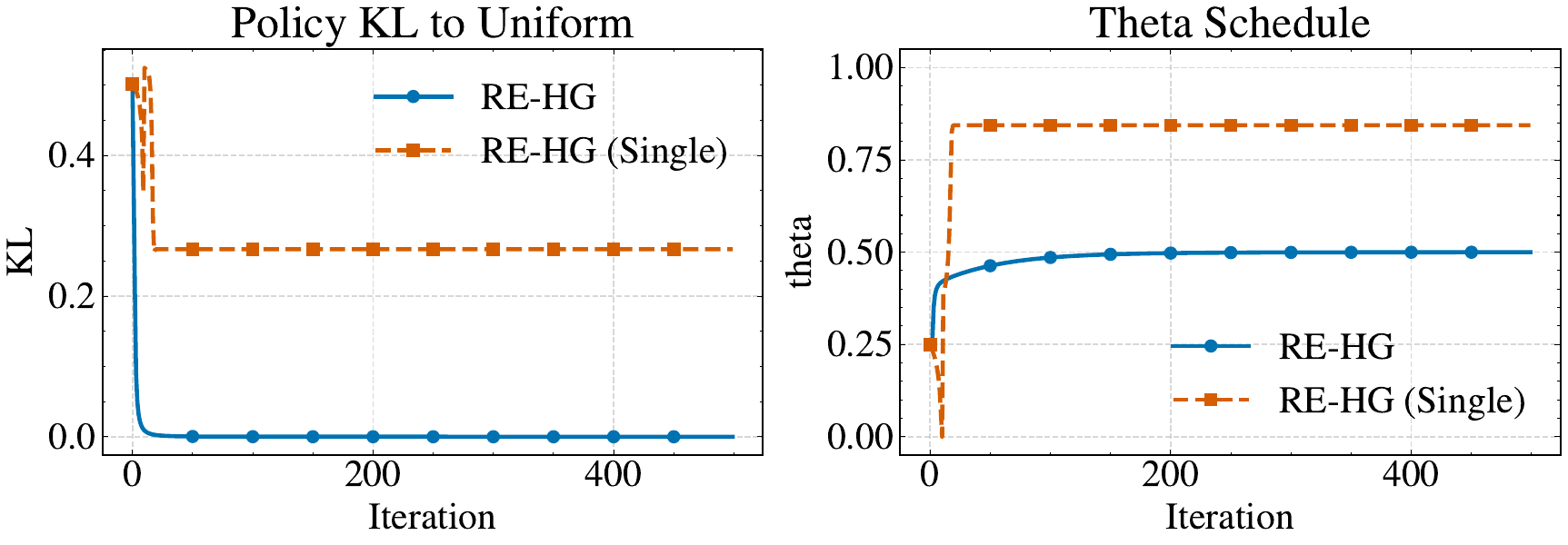}
    \caption{Summary of Diagnostics}
    \label{fig:conf_discrete_kl_theta}
  \end{minipage}

  \vspace{6pt}

  \begin{minipage}[b]{\textwidth}
    \centering
    \includegraphics[width=0.98\textwidth]{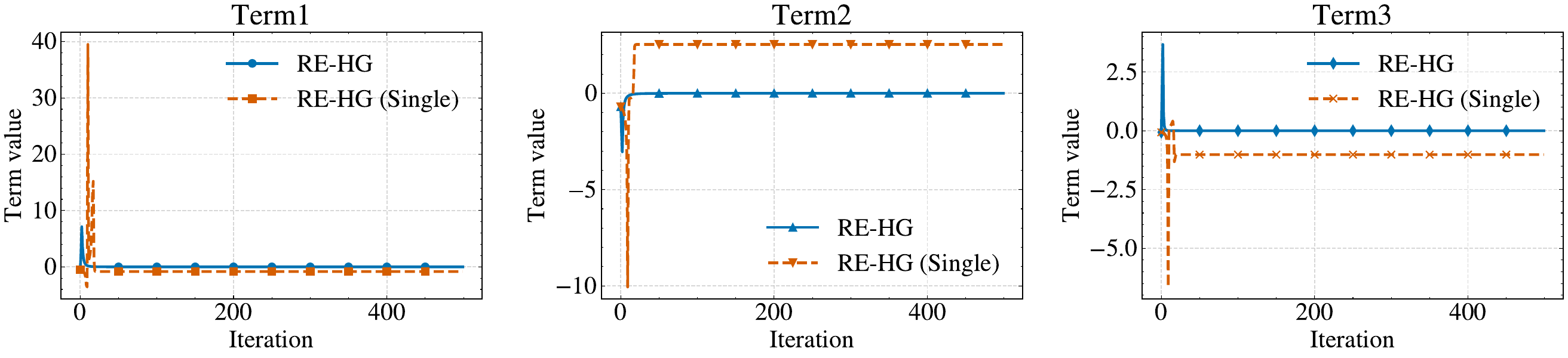}
    \caption{Summary of Terms}
    \label{fig:conf_discrete_terms}
  \end{minipage}
\end{figure}

\paragraph{Hyperparameters}
Hyperparameters used in this experiment are summarized in 
\Cref{tab:hyperparams_syn}.

\begin{table}[H]
    \centering
    \caption{Hyperparameters}
    \label{tab:hyperparams_syn}
    \begin{tabular}{lcc}
        \toprule
        Parameter & Symbol & Value \\
        \midrule
        Leader Discount Factor & $\gamma_L$ & 0.95 \\
        Follower Discount Factor & $\gamma_F$ & 0.95 \\
        Leader Parameter Learning Rate & $\alpha_{L,\text{actor}}$ & $1.0 \times 10^{-2}$ \\
        Initial Leader Parameter  & $\theta_0$ & $0.25$ \\
        Leader Iterations & - & 500 \\
        Temperature ($\beta$) & -  & $2.5 \times 10^{-1}$ \\
        \bottomrule
    \end{tabular}
\end{table}

\subsection{2-player Markov Game (Discrete Case)}
\label{apdx:2mg-discrete}

\paragraph{Implementation Details}
In Lines~\ref{line:irl0}, ~ \ref{line:irl1} of~\Cref{algo:proposal}, when the transition dynamics $P$ is known, the leader computes the surrogate follower's policy via Soft Value Iteration (SoftVI).  
At Lines~\ref{line:irl0},~\ref{line:irl1} of~\Cref{algo:proposal}, when the transition dynamics $P$ is not known, the leader computes the expected feature value under the initial distribution $\rho_0$ by incorporating Successor Features (SFs)~\citep{10.5555/3294996.3295161}. IRL is performed at iterations 0, 10, $\dots$, 90. 
At Lines~\ref{line:HG0},~\ref{line:for},~\ref{line:benefit},~\ref{line:FG} of~\Cref{algo:proposal}, when the transition dynamics $P$ is known, the leader computes the value functions and gradients via dynamic programming. This approach eliminates sampling noise and significantly improves learning efficiency by avoiding Monte-Carlo simulations. However, it should be noted that this is an approximation based on learned SFs rather than unbiased sampling; therefore, the estimates may inherit biases from TD errors during the SF training process. 
At Line ~\ref{line:HG1} of~\Cref{algo:proposal}, we utilize the \Cref{algo:baseline_influence}.

\begin{figure}[H]
  \centering
  \begin{subfigure}{0.5\textwidth}
    \centering
    \includegraphics[width=0.9\textwidth,clip,trim=0 0 0 0]{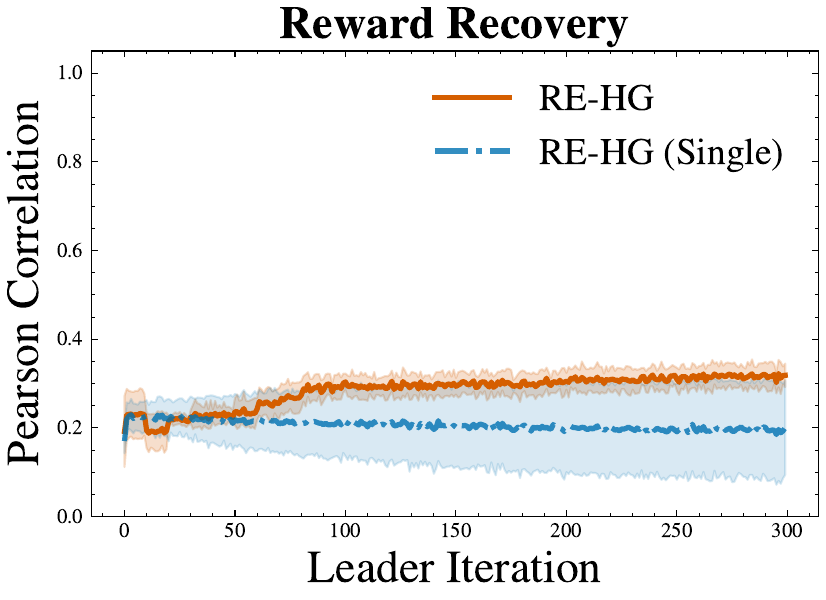}
    \caption{Reward Correlation}
    \label{fig:dr}
  \end{subfigure}%
  \begin{subfigure}{0.5\textwidth}
    \centering
    \includegraphics[width=0.9\textwidth,clip,trim=0 0 0 0]{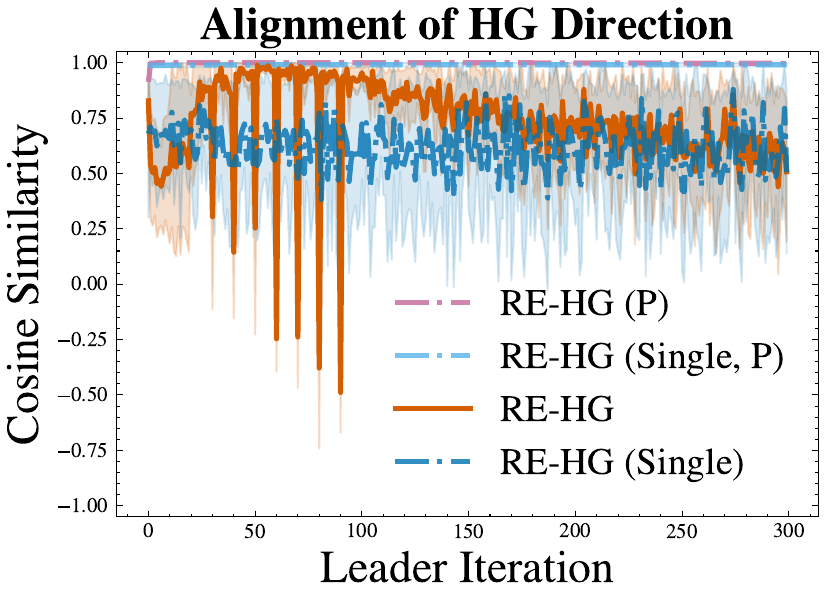}
    \caption{HG Similarity}
    \label{fig:dh}
  \end{subfigure}%
  \caption{Discrete toy task support material.``P'' indicates the known transition probability scenario. (a) Pearson correlation between the recovered and true rewards, weighted by the undiscounted state-action occupancy measure of the surrogate follower policy. For each run, performance is evaluated as the mean of 10 test rollouts. Avg.\ (line) and std.\ (shaded area) across the 3 independent training runs are displayed. (b) Cosine similarity of the HG between the estimated and Oracle models. For each run, performance is evaluated as the mean of 10 test rollouts. Avg.\ (line) and std.\ (shaded area) across the 3 independent training runs are displayed.}
  \label{fig:integrated_results_2}
\end{figure}

\paragraph{Result}
As discussed in \Cref{sec:2mg-dis}, \Cref{fig:integrated_results_2} provides the detailed trajectories of the reward correlation and the HG cosine similarity. \Cref{fig:dr} shows that the Pearson correlation remains relatively low. This is primarily due to estimation errors in regions where insufficient trajectories were collected, which limits the accuracy of reward recovery. However, as illustrated in \Cref{fig:dh}, the cosine similarity of the HG remains highly stable throughout the training process. This indicates that the estimation errors in those rarely visited regions do not negatively affect the overall gradient direction.

\paragraph{Hyperparameters}
Hyperparameters used in this experiment are summarized in 
\Cref{tab:hyperparams_commond}.
\begin{table}[H]
    \centering
    \caption{Hyperparameters}
    \label{tab:hyperparams_commond}
    \begin{tabular}{lcc}
        \toprule
        Parameter & Symbol & Value \\
        \midrule
        Leader Discount Factor & $\gamma_L$ & 0.99 \\
        Leader Actor Learning Rate & $\alpha_{L,\text{actor}}$ & $1.0 \times 10^{-2}$ \\
        Leader Critic Learning Rate & $\alpha_{L,\text{critic}}$ & $1.0 \times 10^{-3}$ \\
        Leader Iterations & - & 300 \\
        Critic Updates per Iteration & - & 10 \\
        Follower Discount Factor & $\gamma_F$ & 0.9 \\
        Follower Learning Rate (SoftQL) & $\alpha_F$ & 0.01 \\
        Temperature ($\beta$) & -  & $2.5 \times 10^{-1}$ \\
        Max Inner Iterations & - & 150 \\
        Inner Episodes & - & 50 \\
        Successor feature Update Episodes & - & 100 \\
        Expert Episodes per Iteration & - & 200 \\
        IRL Rollout Episodes & - & 100 \\
        $\omega$ Learning Rate  & - & $5 \times 10^{-3}$ \\
        IRL Timing & - &  Step 0,10, $\dots$ ,90 \\
        Successor feature Learning Rate & $\alpha_{\omega}$ & 0.05 \\
        Adam First Moment Coefficient & $\beta_1$ & 0.9 \\
        Adam Second Moment Coefficient & $\beta_2$ & 0.999 \\
        threshold  & $\epsilon$ & $10^{-23}$ \\
        \bottomrule
    \end{tabular}
\end{table}

\subsection{Configurable MDP Continuous}\label{apdx:Conf-MDP}

\paragraph{Analytical Derivation of the Leader's Value Functions}
First, we define the system dynamics, the leader's immediate reward, and the follower's best response policy. The state transition is given by:
\begin{equation} \label{eq:transition}
    s^\prime = A_\theta s + B b + w, \quad w \sim \mathcal{N}(\bm{0}, \Sigma_w).
\end{equation}
The leader's immediate reward is defined using the weighting matrices $W_L$ and $U_L$, and a configuration cost $c_L(\theta)$:
\begin{equation} \label{eq:leader_reward}
    r_L(s, b) = -s^\top W_L s - b^\top U_L b - c_L(\theta).
\end{equation}
The follower's best response policy is $b \sim \mathcal{N}(-K_s^\theta s, \Sigma_F^\theta)$, which can be reparameterized using the noise term $\epsilon_b$ as:
\begin{equation} \label{eq:follower_policy}
    b = -K^\theta s + \epsilon_b, \quad \epsilon_b \sim \mathcal{N}(\bm{0}, \Sigma_F^\theta).
\end{equation}

We assume the leader's optimal value function takes a quadratic form:
\begin{equation} \label{eq:v_ansatz}
    V_L^{\theta\dagger}(s) = -s^\top P_L^\theta s - v_L^\theta,
\end{equation}
where $P_L^\theta \ge 0$ is a symmetric matrix and $v_L^\theta$ is a scalar. The leader's action-value function $Q_L^{\theta\dagger}(s, b)$ and the state-value function $V_L^{\theta\dagger}(s)$ satisfy the following Bellman equations:
\begin{equation} \label{eq:q_bellman}
    Q_L^{\theta\dagger}(s, b) = r_L(s, b) + \gamma_Lt \mathbb{E}_{w}\left[ V_L^{\theta\dagger}(s^\prime) \right],
\end{equation}
\begin{equation} \label{eq:v_bellman}
    V_L^{\theta\dagger}(s) = \mathbb{E}_{\epsilon_b}\left[ Q_L^{\theta\dagger}(s, b) \right].
\end{equation}

To identify $P_L^\theta$ and $v_L^\theta$, we first evaluate the expectation over the unobservable disturbance $w$ in Eq.~\eqref{eq:q_bellman}. Substituting Eq.~\eqref{eq:transition} and Eq.~\eqref{eq:v_ansatz} into Eq.~\eqref{eq:q_bellman}, and utilizing the properties $\mathbb{E}_w[w]=\bm{0}$ and $\mathbb{E}_w[w w^\top]=\Sigma_w$, we obtain:
\begin{align}
    \mathbb{E}_w\left[ V_L^{\theta\dagger}(A_\theta s + B b + w) \right]
    &= -\mathbb{E}_w\left[ (A_\theta s + B b + w)^\top P_L^\theta (A_\theta s + B b + w) \right] - v_L^\theta \nonumber \\
    &= -(A_\theta s + B b)^\top P_L^\theta (A_\theta s + B b) - \mathbb{E}_w[\mathrm{Tr}(w^\top P_L^\theta w)] - v_L^\theta \nonumber \\
    &= -(A_\theta s + B b)^\top P_L^\theta (A_\theta s + B b) - \mathrm{Tr}(P_L^\theta \Sigma_w) - v_L^\theta. \label{eq:expected_v_w}
\end{align}
Substituting Eq.~\eqref{eq:leader_reward} and Eq.~\eqref{eq:expected_v_w} into Eq.~\eqref{eq:q_bellman} yields the closed quadratic form for $Q_L^{\theta\dagger}(s, b)$:
\begin{equation} \label{eq:q_expanded}
    Q_L^{\theta\dagger}(s, b) = -s^\top W_L s - b^\top U_L b - c_L(\theta) - \gamma_L((A_\theta s + B b)^\top P_L^\theta (A_\theta s + B b) +  \mathrm{Tr}(P_L^\theta \Sigma_w) +  v_L^\theta).
\end{equation}

Next, we evaluate Eq.~\eqref{eq:v_bellman} by taking the expectation of Eq.~\eqref{eq:q_expanded} over the follower's action noise $\epsilon_b$. Defining the closed-loop transition matrix as $\tilde{A}_\theta := A_\theta - B K_s^\theta$ and substituting Eq.~\eqref{eq:follower_policy}, we expand the state-action terms:
\begin{align}
    V_L^{\theta\dagger}(s) &= \mathbb{E}_{\epsilon_b} \bigg[ -s^\top W_L s - (-K_s^\theta s + \epsilon_b)^\top U_L (-K_s^\theta s + \epsilon_b) - c_L(\theta) \nonumber \\
    &\qquad\quad - \gamma_L((\tilde{A}_\theta s + B \epsilon_b)^\top P_L^\theta (\tilde{A}_\theta s + B \epsilon_b) +  \mathrm{Tr}(P_L^\theta \Sigma_w) +  v_L^\theta) \bigg]. \label{eq:v_expectation_expanded}
\end{align}
Since $\mathbb{E}_{\epsilon_b}[\epsilon_b] = \bm{0}$, the cross terms vanish. By applying $\mathbb{E}_{\epsilon_b}[\epsilon_b \epsilon_b^\top] = \Sigma_F^\theta$, the quadratic terms in $\epsilon_b$ evaluate to traces:
\begin{align}
    V_L^{\theta\dagger}(s) &= -s^\top \left( W_L + (K_s^\theta)^\top U_L K_s^\theta + \gamma_L \tilde{A}_\theta^\top P_L^\theta \tilde{A}_\theta \right) s \nonumber \\
    &\quad - \mathrm{Tr}(U_L \Sigma_F^\theta)- c_L(\theta) - \gamma_L(\mathrm{Tr}(P_L^\theta B \Sigma_F^\theta B^\top)  +  \mathrm{Tr}(P_L^\theta \Sigma_w) + v_L^\theta). \label{eq:v_final_form}
\end{align}

Finally, by matching the quadratic and scalar coefficients of Eq.~\eqref{eq:v_final_form} with the ansatz in Eq.~\eqref{eq:v_ansatz}, we conclude that $P_L^\theta$ is the unique positive semi-definite solution to the discrete-time Lyapunov equation:
\begin{equation} \label{eq:p_solution}
    P_L^\theta = W_L + (K_s^\theta)^\top U_L K_s^\theta + \gamma_L \tilde{A}_\theta^\top P_L^\theta \tilde{A}_\theta.
\end{equation}
The scalar constant $v_L^\theta$ is subsequently obtained as:
\begin{equation} \label{eq:v_solution}
    v_L^\theta = \frac{1}{1-\gamma_L} \left( \mathrm{Tr}(U_L \Sigma_F^\theta) + c_L(\theta)+ \gamma_L(\mathrm{Tr}(P_L^\theta B \Sigma_F^\theta B^\top)  +  \mathrm{Tr}(P_L^\theta \Sigma_w))   \right).
\end{equation}
Thus, the leader's value functions are fully determined analytically without requiring neural network approximations.

\subsubsection{Easy Setting}
\paragraph{Task Description}
The task is defined as an $n$-zone building thermal control problem with $n=4$. The state $s_t \in \mathbb{R}^4$ represents temperature deviations, and the follower's action $b_t \in \mathbb{R}^2$ represents HVAC actuation.
As defined previously, the system dynamics follow $s_{t+1} = A_\theta s_t + B b_t + w_t$ with $w_t \sim \mathcal{N}(\bm{0}, \Sigma_w)$, and the leader's immediate reward is $r_L(s_t, b_t) = -(s_t^\top W_L s_t + b_t^\top U_L b_t + c_L(\theta))$.

For the leader's immediate reward, the specific weighting matrices $W_L$, $U_L$, and the configuration cost $c_L(\theta)$ are defined as:
\begin{equation}
    W_L = \frac{1}{4} \left( I_4 - \frac{1}{4}\mathbf{1}\mathbf{1}^\top \right), \quad U_L = 0.5 \cdot I_2, \quad c_L(\theta) = 0.1 \cdot \left(\|\alpha\|_2^2 + \|\nu\|_2^2\right),
\end{equation}
where $I_n$ is the identity matrix, and $\mathbf{1} \in \mathbb{R}^n$ is a vector of ones.

The configurable state transition matrix $A_\theta \in \mathbb{R}^{4 \times 4}$ is parameterized by the leader's decision variables, insulation levels $\alpha \in [0, 1]^4$ and airflow levels $\nu \in [0, 1]^4$, as follows:
\begin{equation}
    \resizebox{\textwidth}{!}{
    $A_\theta = \begin{bmatrix}
    1 + \Delta(\alpha_1) - h_{12}\nu_1 - h_{41}\nu_4 & h_{12}\nu_1 & 0 & h_{41}\nu_4 \\
    h_{21}\nu_1 & 1 + \Delta(\alpha_2) - h_{23}\nu_2 - h_{21}\nu_1 & h_{23}\nu_2 & 0 \\
    0 & h_{32}\nu_2 & 1 + \Delta(\alpha_3) - h_{34}\nu_3 - h_{32}\nu_2 & h_{34}\nu_3 \\
    h_{41}\nu_4 & 0 & h_{43}\nu_3 & 1 + \Delta(\alpha_4) - h_{41}\nu_4 - h_{43}\nu_3
    \end{bmatrix}$
    }
\end{equation}
where $\Delta(\alpha_i) := k_i(1 - \alpha_i)$. The positive coefficients $k_i$ and $h_{ij}$ representing specific physical constants are set as:
\begin{equation}
\begin{aligned}
    k_1 &= 0.04, & h_{12} &= 0.05, & h_{41} &= 0.05 \\
    k_2 &= 0.03, & h_{23} &= 0.04, & h_{21} &= 0.03 \\
    k_3 &= 0.06, & h_{34} &= 0.06, & h_{32} &= 0.04 \\
    k_4 &= 0.05, & h_{41} &= 0.05, & h_{43} &= 0.03
\end{aligned}
\end{equation}

The follower's immediate reward is defined as a quadratic cost:
\begin{equation}
    r_F(s, b) = -s^\top W_F s - b^\top U_F b.
\end{equation}

The specific weighting matrices for the follower are given by:
\begin{equation}
    W_F = \begin{bmatrix} 8 & 0 & 0 & 0 \\ 0 & 1 & 0 & 0 \\ 0 & 0 & 5 & 0 \\ 0 & 0 & 0 & 6 \end{bmatrix}, \quad U_F = 0.01\cdot I_2.
\end{equation}

The control input matrix $B$ and the unobservable disturbance covariance $\Sigma_w$ are defined as:
\begin{equation}
    B = \begin{bmatrix} 0.10 & 0 \\ 0.6 & 0 \\ 0 & 0.55 \\ 0 & 0.30 \end{bmatrix}, \quad \Sigma_w = 0.02\cdot I_4.
\end{equation}

\paragraph{Implementation Details}
In Line~\ref{line:QF} of~\Cref{algo:proposal}, since the transition dynamics are linear-Gaussian and the reward functions are quadratic, we obtain $Q_F^{\theta \dagger}$ directly from the Discrete Algebraic Riccati Equation (DARE) solution for the follower's policy.
For the same reason, in Line~\ref{line:HG0} of~\Cref{algo:proposal}, we analytically compute $Q_L^{\theta\dagger}$ and $V_L^{\theta\dagger}$ by solving the DARE.
In Line~\ref{line:FG} of~\Cref{algo:proposal}, we obtain $\nabla Q_F^{\theta \dagger}$ via the discrete Lyapunov equation.
Finally, in Line~\ref{line:infterm} of \Cref{algo:proposal}, the HG components are computed analytically using the known transition dynamics and policies, which bypasses the need for empirical trajectory sampling in these steps.

\paragraph{Result}

\begin{figure}[H]
    \centering
    \includegraphics[width=0.7\textwidth]{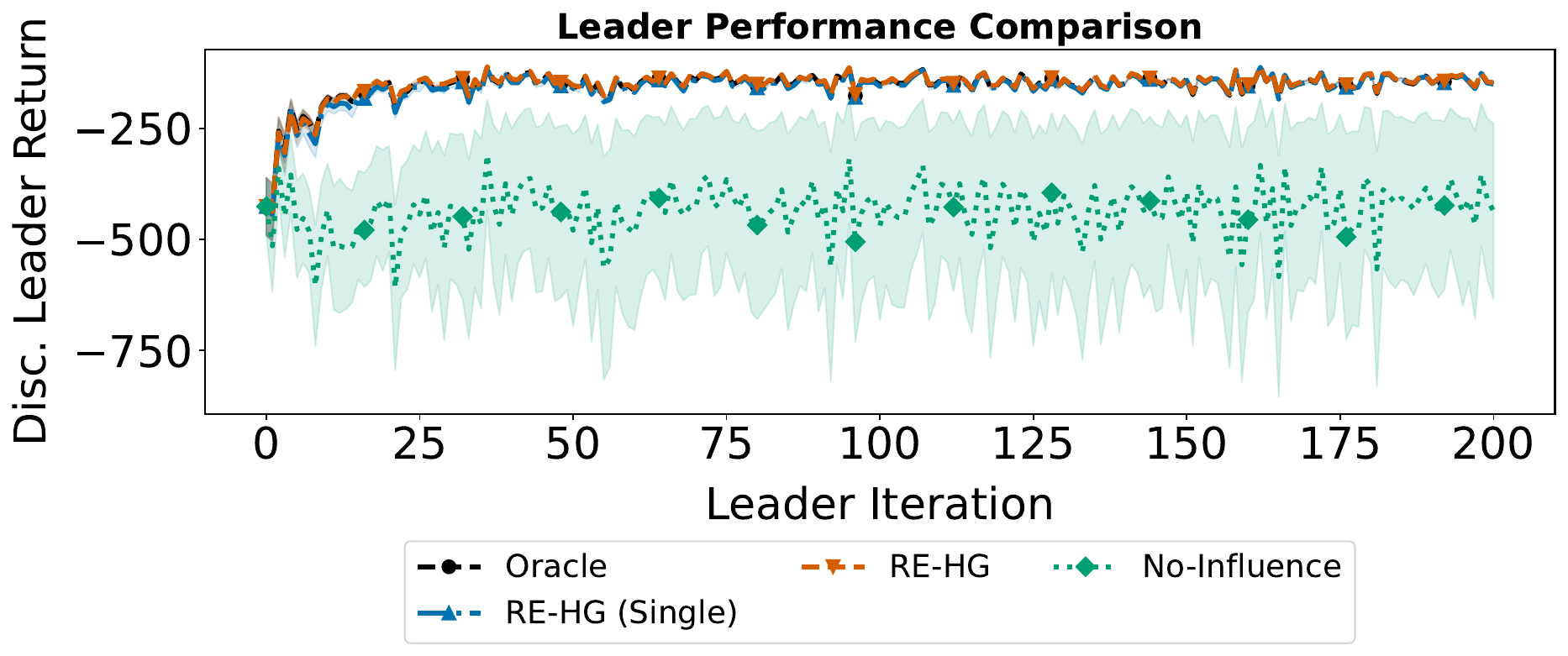}
    \caption{Results on Building Thermal Control. For each run, performance is evaluated as the mean of 50 test rollouts. Avg.\ (line) and std.\ (shaded area) across the 3 independent training runs are displayed.}
    \label{fig:0.3godcomparison_curves_leader_only.pdf}
\end{figure}
\Cref{fig:0.3godcomparison_curves_leader_only.pdf} presents the performance comparison in the Building Thermal Control task. In our setup, the initial state is stochastic, and the leader's initial parameters $\theta$ are sampled from a uniform distribution $\mathcal{U}(0, 0.3)$. To ensure the statistical reliability of the results despite this randomness, we evaluated the methods across three different environments, each with ten independent trials using distinct random seeds.

The results clearly demonstrate that \rehg significantly outperforms \noinf. Notably, the performance of \rehg is comparable to that of \oracle. This indicates that our approach effectively recovers the necessary information to achieve near-optimal leader strategies.

\paragraph{Hyperparameters}
Hyperparameters used in this experiment are summarized in 
\Cref{tab:confmdp_hyperparamsease}.

\begin{table}[H]
\centering
\caption{Hyperparameters}
\label{tab:confmdp_hyperparamsease}
\begin{tabular}{llc}
\toprule
\textbf{Name} & \textbf{Symbol} & \textbf{Value} \\
\midrule
Leader iterations & $K$ & $200$ \\
Episodes per iteration & - & $100$ \\
Evaluation episodes & - & $50$ \\
Leader discount factor & $\gamma_L$ & $0.9$ \\
Follower discount factor & $\gamma_F$ & $0.9$ \\
Leader learning rate & $\alpha_L$ & $0.1$ \\
Temperature & $\beta$ & $0.1$ \\
Gradient clip max norm & - & $1.0$ \\
Local IRL iterations & - & $200$ \\
Local IRL learning rate & - & $0.01$ \\
threshold & $\epsilon$ & $10^{-4}$ \\
\bottomrule
\end{tabular}
\end{table}

\subsubsection{Hard Setting (8-Zone Central Corridor Topology)}
\label{apdx:hard-setting}

\paragraph{Task Description}
The task is extended to an $n$-zone building thermal control problem where $n=8$. The state $s_t \in \mathbb{R}^8$ represents temperature deviations, and the follower's action $b_t \in \mathbb{R}^4$ represents HVAC actuation for the directly controlled zones.
As defined previously, the system dynamics follow $s_{t+1} = A_\theta s_t + B b_t + w_t$ with $w_t \sim \mathcal{N}(\bm{0}, \Sigma_w)$, and the leader's immediate reward is $r_L(s_t, b_t) = -(s_t^\top W_L s_t + b_t^\top U_L b_t + c_L(\theta))$.

For the leader's immediate reward, the specific weighting matrices $W_L$, $U_L$, and the configuration cost $c_L(\theta)$ are defined as:
\begin{equation}
    W_L = \frac{1}{8} \left( I_8 - \frac{1}{8}\mathbf{1}\mathbf{1}^\top \right), \quad U_L = 0.5 \cdot I_4, \quad c_L(\theta) = \|\alpha\|_2^2 + \|\nu\|_2^2,
\end{equation}
where $I_n$ is the identity matrix, and $\mathbf{1} \in \mathbb{R}^n$ is a vector of ones.

The configurable state transition matrix $A_\theta \in \mathbb{R}^{8 \times 8}$ is parameterized by the leader's decision variables, insulation levels $\alpha \in [0, 1]^8$ and airflow levels $\nu \in [0, 1]^8$. The topology consists of two core corridors (Zones 1 and 2) and six perimeter offices (Zones 3--8). The off-diagonal entries $A_{i,j}$ (representing heat transfer from zone $j$ to zone $i$) are:
\begin{equation}
A_{i,j} = \begin{cases}
h_{cc} \nu_1 & \text{if } (i,j) \in \{(1,2), (2,1)\} \\
h_{cr} \nu_{k} & \text{if } i \in \{1,2\} \text{ and } j \text{ is an adjacent office} \\
h_{rc} \nu_{k} & \text{if } j \in \{1,2\} \text{ and } i \text{ is an adjacent office} \\
0 & \text{otherwise}
\end{cases}
\end{equation}
where $\nu_k$ corresponds to the specific airflow parameter for the edge. Specifically, the edges connected to the West corridor (Zone 1) are controlled by $\nu_3$ (Zone 3), $\nu_4$ (Zone 4), and $\nu_6$ (Zone 6). The edges connected to the East corridor (Zone 2) are controlled by $\nu_5$ (Zone 5), $\nu_7$ (Zone 7), and $\nu_8$ (Zone 8). The parameter $\nu_2$ is kept redundant to maintain the unified dimensionality $\theta \in [0,1]^{16}$.
The diagonal entries are defined to preserve energy balance:
\begin{equation}
    A_{i,i} = 1 + \Delta(\alpha_i) - \sum_{j \neq i} A_{j,i}
\end{equation}
where $\Delta(\alpha_i) := k_i(1 - \alpha_i)$. The physical heat transfer coefficients are set to $h_{cc} = 0.08$, $h_{cr} = 0.05$, $h_{rc} = 0.03$, and the insulation effects $k_i$ are:
\begin{equation}
\begin{aligned}
    k_1 &= 0.01, & k_2 &= 0.01, & k_3 &= 0.06, & k_4 &= 0.08 \\
    k_5 &= 0.06, & k_6 &= 0.03, & k_7 &= 0.04, & k_8 &= 0.03
\end{aligned}
\end{equation}

The follower's immediate reward is defined as a quadratic cost:
\begin{equation}
    r_F(s, b) = -s^\top W_F s - b^\top U_F b.
\end{equation}

The specific weighting matrices for the follower are given by:
\begin{equation}
    W_F = \text{diag}(0.1, 0.1, 8.0, 30.0, 8.0, 3.0, 5.0, 3.0), \quad U_F = 0.01\cdot I_4.
\end{equation}
Note that Zone 4 (VIP office) is assigned the highest priority weight ($W_{F,44} = 30.0$) despite having no direct HVAC control, creating a challenging under-actuated control task for the follower.

The control input matrix $B \in \mathbb{R}^{8 \times 4}$ and the unobservable disturbance covariance $\Sigma_w$ are defined as:
\begin{equation}
    B = \begin{bmatrix} 
    0 & 0 & 0 & 0 \\ 
    0 & 0 & 0 & 0 \\ 
    0.8 & 0 & 0 & 0 \\ 
    0 & 0 & 0 & 0 \\ 
    0 & 0.8 & 0 & 0 \\ 
    0 & 0 & 0.8 & 0 \\ 
    0 & 0 & 0 & 0 \\ 
    0 & 0 & 0 & 0.8 
    \end{bmatrix}, \quad \Sigma_w = 0.02\cdot I_8.
\end{equation}

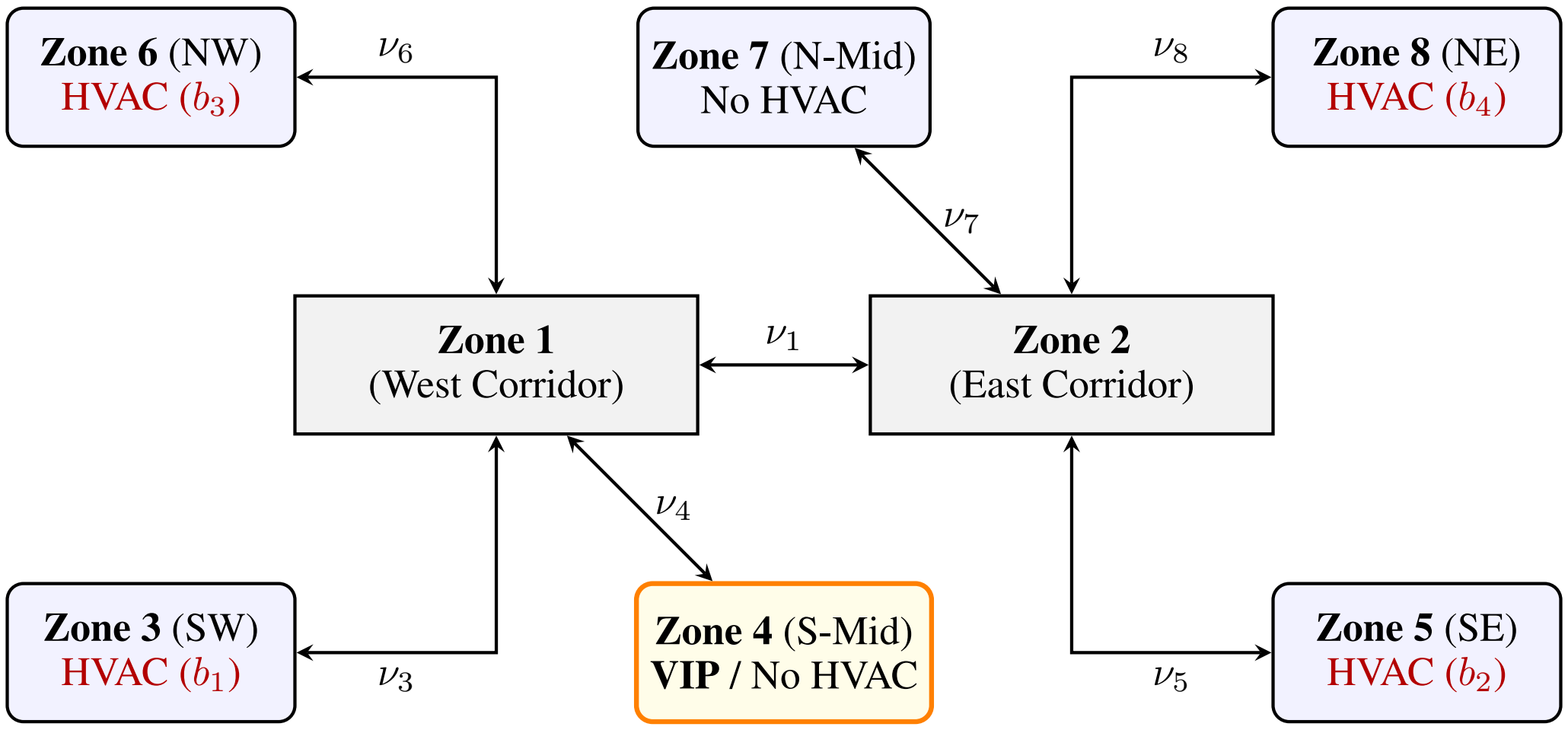
\begin{figure}[H]
    \centering
    \begin{tikzpicture}[
        zone/.style={draw, thick, minimum width=2.5cm, minimum height=1.2cm, align=center, fill=blue!5, rounded corners},
        vip/.style={draw=orange, very thick, minimum width=2.5cm, minimum height=1.2cm, align=center, fill=yellow!10, rounded corners},
        corridor/.style={draw, thick, minimum width=3.5cm, minimum height=1.2cm, align=center, fill=gray!10},
        edge/.style={<->, thick, >=stealth}
    ]
        
        \node[corridor] (z1) at (-2.5, 0) {\textbf{Zone 1}\\(West Corridor)};
        \node[corridor] (z2) at (2.5, 0) {\textbf{Zone 2}\\(East Corridor)};
        
        \node[zone] (z6) at (-5.5, 2.5) {\textbf{Zone 6} (NW)\\ \textcolor{red!70!black}{HVAC ($b_3$)}};
        \node[zone] (z3) at (-5.5, -2.5) {\textbf{Zone 3} (SW)\\ \textcolor{red!70!black}{HVAC ($b_1$)}};
        \node[vip]  (z4) at (0.0, -2.5) {\textbf{Zone 4} (S-Mid)\\ \textbf{VIP} / No HVAC};
        
        \node[zone] (z7) at (0.0, 2.5) {\textbf{Zone 7} (N-Mid)\\ No HVAC};
        \node[zone] (z8) at (5.5, 2.5) {\textbf{Zone 8} (NE)\\ \textcolor{red!70!black}{HVAC ($b_4$)}};
        \node[zone] (z5) at (5.5, -2.5) {\textbf{Zone 5} (SE)\\ \textcolor{red!70!black}{HVAC ($b_2$)}};
        
        \draw[edge] (z1) -- (z2) node[midway, above] {$\nu_1$};
        
        \draw[edge] (z1) |- (z6) node[pos=0.75, above] {$\nu_6$};
        \draw[edge] (z1) |- (z3) node[pos=0.75, below] {$\nu_3$};
        \draw[edge] (z1) -- (z4) node[midway, right] {$\nu_4$};
        
        \draw[edge] (z2) -- (z7) node[midway, right] {$\nu_7$};
        \draw[edge] (z2) |- (z8) node[pos=0.75, above] {$\nu_8$};
        \draw[edge] (z2) |- (z5) node[pos=0.75, below] {$\nu_5$};

    \end{tikzpicture}
    \caption{Topology of the 8-Zone Building Thermal Control Environment. Nodes represent the zones, with their respective HVAC control inputs. Edges represent the heat transfer between adjacent zones, parameterized by the leader's airflow configuration $\nu_k$. Note that the VIP office (Zone 4) has no direct HVAC, highlighting the under-actuated nature of the follower's control task.}
    \label{fig:8zone_topology}
\end{figure}

\paragraph{Implementation Details}
In Line~\ref{line:QF} of~\Cref{algo:proposal}, since the transition dynamics are linear-Gaussian and the reward functions are quadratic, we obtain $Q_F^{\theta \dagger}$ directly from the DARE solution for the follower's policy.
For the same reason, in Line~\ref{line:HG0} of~\Cref{algo:proposal}, we analytically compute $Q_L^{\theta\dagger}$ and $V_L^{\theta\dagger}$ by solving the DARE.
At line~\ref{line:infterm} in~\Cref{algo:proposal}, we adopt the variance reduction baseline as formulated in \eqref{eq:hg_configurable_biased}. 
We utilize the relation $\nabla_{\theta} Q_F^{\theta \dagger} - \nabla_{\theta} V_F^{\theta \dagger} = \nabla_{\theta} \log g^{\theta\dagger}$, which is computed by the discrete Lyapunov equation. 
Furthermore, we use the expectation $\mathbb{E}[V_L^{\theta\dagger}(s_{t+1})]$ obtained via DARE as a baseline for the $\nabla_{\theta} \log p^{\theta}$ term to stabilize the stochastic gradient estimates. Finally, in Line~\ref{line:infterm} of \Cref{algo:proposal}, the HG components are computed analytically using the known transition dynamics and policies, which bypasses the need for empirical trajectory sampling in these steps.
To ensure stable gradient-based optimization, we apply gradient norm clipping 
with a maximum norm of $1.0$. If the L2-norm of the gradient exceeds this threshold, 
we rescale it to maintain numerical stability during policy updates.

\paragraph{Result}

\begin{figure}[H]
  \centering
  \begin{subfigure}{0.44\textwidth}
    \centering
    \includegraphics[width=\textwidth,clip,trim=0 0 0 0]{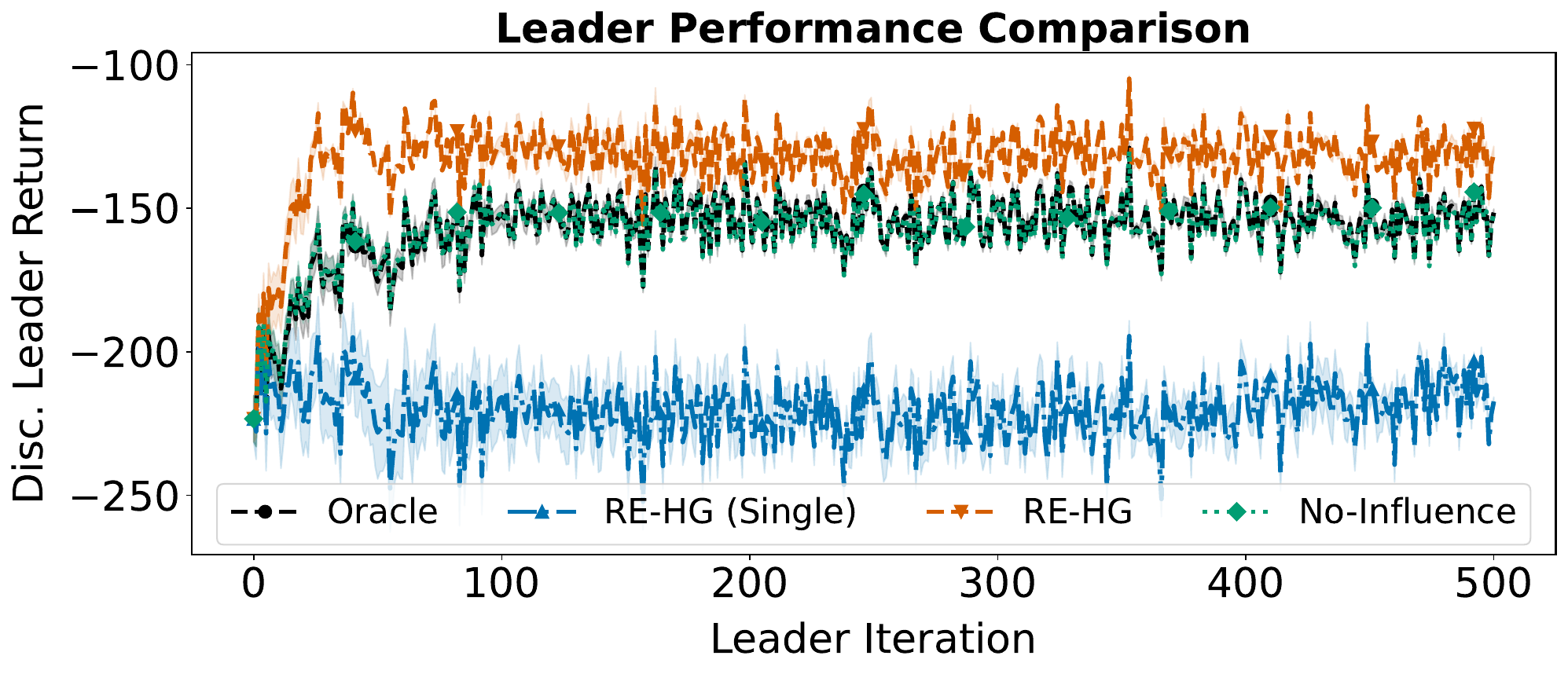}
    \caption{Leader's discounted return}
    \label{fig:1clrsub}
  \end{subfigure}\hfill
  \begin{subfigure}{0.28\textwidth}
    \centering
    \includegraphics[width=\textwidth,clip,trim=0 0 0 0]{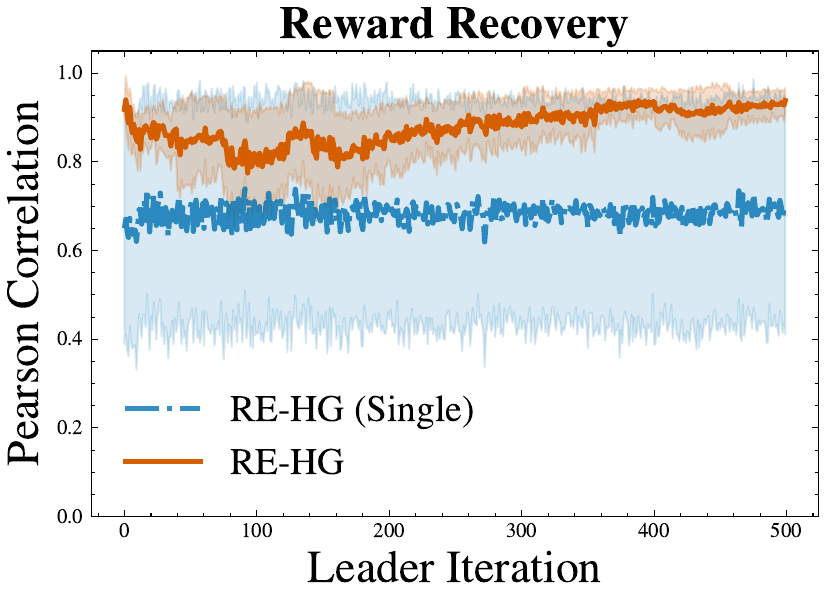}
    \caption{Reward Correlation}
    \label{fig:102_reward_pearson}
  \end{subfigure}\hfill
  \begin{subfigure}{0.28\textwidth}
    \centering
    \includegraphics[width=\textwidth,clip,trim=0 0 0 0]{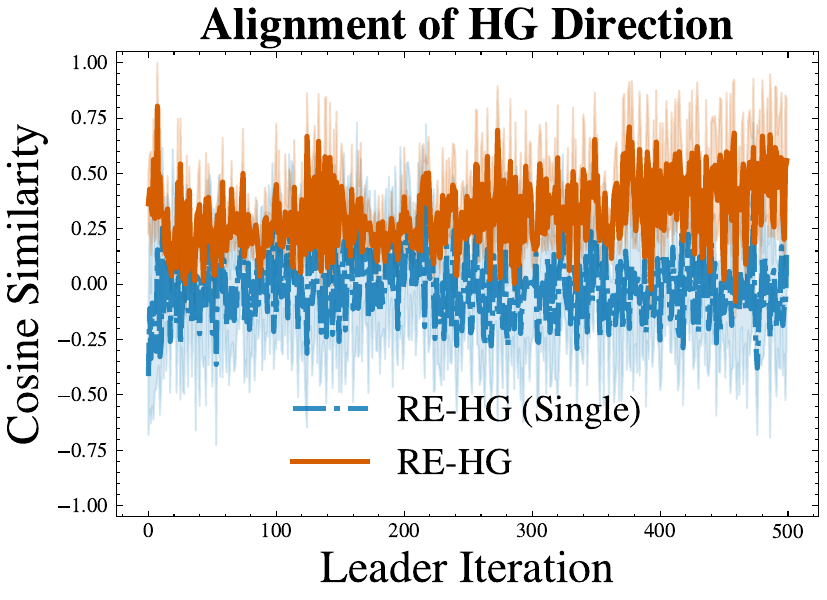}
    \caption{HG CosineSimilarity}
    \label{fig:104_hg_cosine_similarity.pdf}
  \end{subfigure}
  \caption{Results for the relaxed setting ($W_{F,44} = 10.0$). (a) Leader's discounted return. For each run, performance is evaluated as the mean of 50 test rollouts. Avg.\ (line) and std.\ (shaded area) across the 5 independent training runs are displayed. Solid lines and shaded areas denote the avg.\ and std., respectively. (b) Pearson correlation between the recovered and true rewards, weighted by the undiscounted state-action occupancy measure of the surrogate follower policy. (c) Cosine similarity of the HG between the estimated and \oracle}
  \label{fig:smconf10}
\end{figure}

\begin{figure}[H]
  \centering
  \begin{subfigure}{0.44\textwidth}
    \centering
    \includegraphics[width=\textwidth,clip,trim=0 0 0 0]{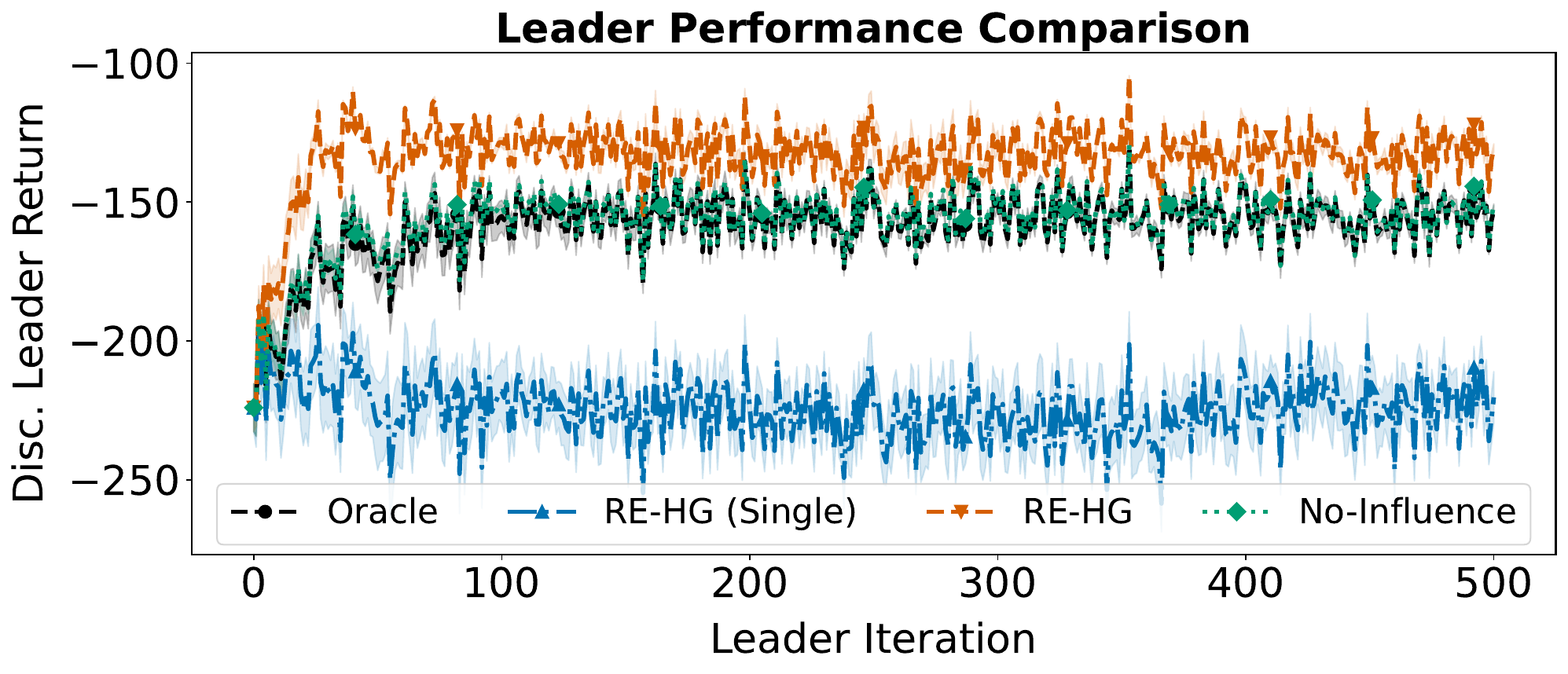}
    \caption{Leader's discounted return}
    \label{fig:2clrsub}
  \end{subfigure}\hfill
  \begin{subfigure}{0.28\textwidth}
    \centering
    \includegraphics[width=\textwidth,clip,trim=0 0 0 0]{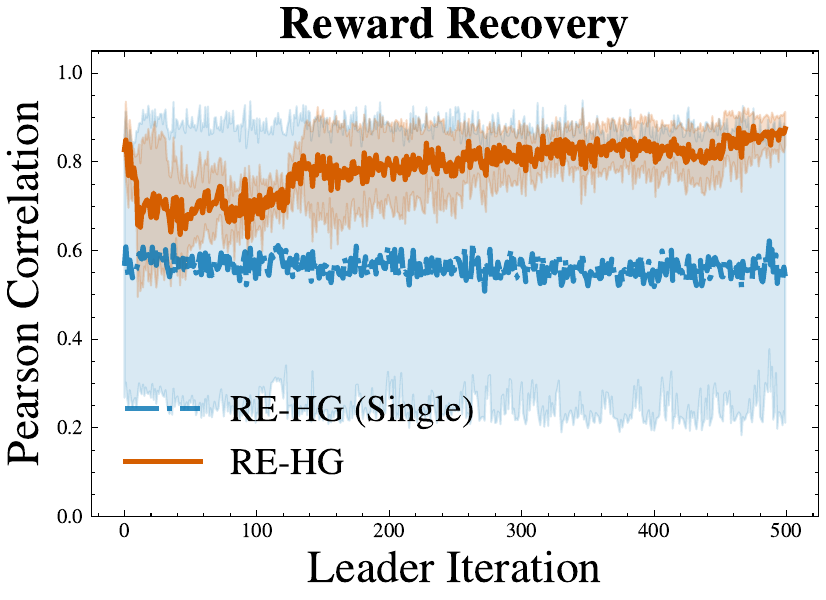}
    \caption{Reward Correlation}
    \label{fig:201_gradient_norm_log.pdf}
  \end{subfigure}\hfill
  \begin{subfigure}{0.28\textwidth}
    \centering
    \includegraphics[width=\textwidth,clip,trim=0 0 0 0]{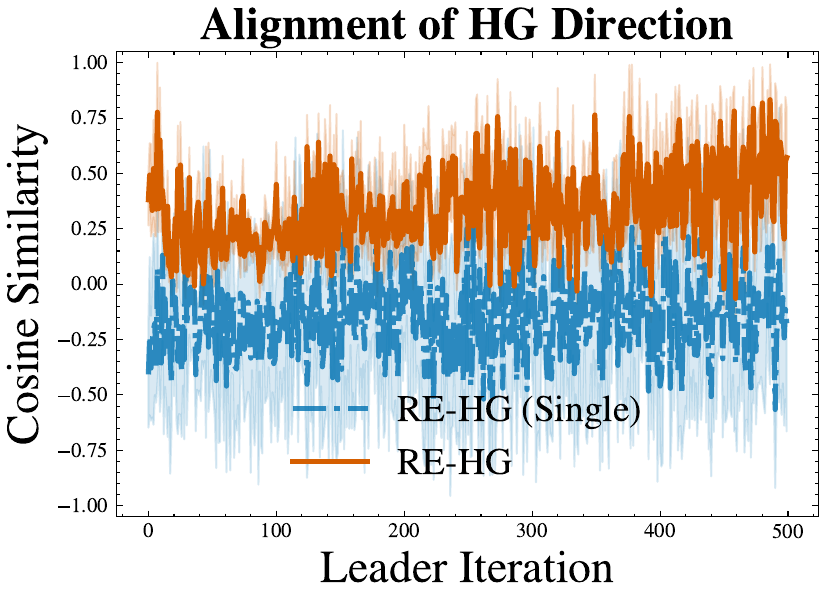}
    \caption{HG CosineSimilarity}
    \label{fig:204_hg_cosine_similarity.pdf}
  \end{subfigure}
  \caption{Results for the relaxed setting ($W_{F,44} = 20.0$). (a) Leader's discounted return. For each run, performance is evaluated as the mean of 50 test rollouts. Avg.\ (line) and std.\ (shaded area) across the 5 independent training runs are displayed. Solid lines and shaded areas denote the avg.\ and std., respectively. (b) Pearson correlation between the recovered and true rewards, weighted by the undiscounted state-action occupancy measure of the surrogate follower policy. (c) Cosine similarity of the HG between the estimated and \oracle.}
  \label{fig:smconf20}
\end{figure}

As discussed in the ablation study in \Cref{sec:c-continuous}, we investigate the impact of the follower's reward landscape steepness on the optimization process. In the original Hard Setting, the highly skewed penalty for the VIP zone ($W_{F,44} = 30.0$) caused gradient explosions, trapping the \oracle in a suboptimal local minimum. 

\Cref{fig:smconf10} and \Cref{fig:smconf20} present the results when this steep penalty is relaxed to $W_{F,44} = 10.0$ and $W_{F,44} = 20.0$, respectively. As shown in the \Cref{fig:smconf10,fig:smconf20}, relaxing the reward steepness mitigates the gradient explosion. Consequently, the performance of the \oracle significantly improves compared to the original $W_{F,44} = 30.0$ case, successfully finding superior leader policies. Meanwhile, \rehg maintains robust and highly stable performance across all variations, consistently achieving high cosine similarity and optimal returns. This confirms that while the \oracle is highly sensitive to sharp reward landscapes, the implicit smoothing mechanism of \rehg inherently generates a manageable gradient landscape, reliably extracting effective HG regardless of the underlying reward steepness.

\paragraph{Hyperparameters}
Hyperparameters used in this experiment are summarized in \Cref{tab:confmdp_hyperparams}.

\begin{table}[h]
\centering
\caption{Hyperparameters}
\label{tab:confmdp_hyperparams}
\begin{tabular}{llc}
\toprule
\textbf{Name} & \textbf{Symbol} & \textbf{Value} \\
\midrule
Leader iterations & $K$ & $500$ \\
Episodes per iteration & - & $100$ \\
Evaluation episodes & - & $50$ \\
Leader discount factor & $\gamma_L$ & $0.9$ \\
Follower discount factor & $\gamma_F$ & $0.9$ \\
Leader learning rate & $\alpha_L$ & $0.1$ \\
Temperature & $\beta$ & $0.1$ \\
Gradient clip max norm & - & $1.0$ \\
Local IRL iterations & - & $200$ \\
Local IRL learning rate & - & $0.01$ \\
threshold & $\epsilon$ & $10^{-2}$ \\
\bottomrule
\end{tabular}
\end{table}

\subsection{2-player MG (Continuous)}
\label{apdx:2mg-continuous}
\paragraph{Task Description}
The environment is modeled as a discretized linear system with a time step $\Delta t = 0.1$. The state $s_t = [p_t, v_t]^\top \in \mathbb{R}^2$ represents the position and velocity, and its evolution is governed by $s_{t+1} = A s_t + B b_t + C a_t$, where $a_t$ and $b_t$ are the control inputs of the leader and the follower, respectively. The system matrices and the initial state $s_0$ are defined as:
\begin{equation}
    A = \begin{bmatrix} 1 & \Delta t \\ 0 & 1 \end{bmatrix}, \quad
    B = \begin{bmatrix} 0 \\ \Delta t \end{bmatrix}, \quad
    C = \begin{bmatrix} \Delta t \\ 0 \end{bmatrix}, \quad
    s_0 = \begin{bmatrix} 1 \\ 0 \end{bmatrix}.
\end{equation}

\paragraph{Reward Functions} 
The follower maximizes the quadratic reward $r_F(s_t, b_t) = -s_t^\top W_F s_t - b_t^\top U_F b_t$ to stabilize at the origin, with $W_F = I_2$ and $U_F = [1]$. The leader aims to induce oscillation around the target $s^\star_L = [-0.8, 0]^\top$ by maximizing a Gaussian kernel reward:
\begin{equation}
    r_L(s_t) = \exp\left( -\frac{1}{2} (s_t - s^\star_L)^\top M (s_t - s^\star_L) \right),
\end{equation}
where $M = 0.25 I_2$. Note that while $r_L$ depends only on the state $s_t$, it is implicitly coupled with the actions $a_t$ and $b_t$ through the system dynamics.

\paragraph{Implementation Details}
Before the main training loop, the leader's critic $Q_L$ is pre-trained using 40 episodes generated from a randomized initial distribution where $s_0[0] \sim \mathcal{U}(-1.5, 1.5)$ and $s_0[1] \sim \mathcal{U}(-1.0, 1.0)$. This dataset comprises approximately 4,000 transitions based on a 100-step limit per episode, which are used to perform 50 optimization epochs via SARSA-TD updates to stabilize the critic initialization.
In Line~\ref{line:irl0} of~\Cref{algo:proposal}, if the transition dynamics are unknown, we estimate them via linear regression.
In Line~\ref{line:FG} of~\Cref{algo:proposal}, we obtain $\nabla Q_F^{\theta \dagger}$ via the discrete Lyapunov equation.
In Line~\ref{line:benefit} of~\Cref{algo:proposal}, the benefit $B_L$ is calculated as the difference between the online leader's action-value function ($Q_L^{\text{online}}$) and the target leader's value function ($V_L^{\text{target}}$).
To ensure stable gradient-based optimization, we apply gradient norm clipping 
with a maximum norm of 1.0. If the L2-norm of the gradient exceeds this threshold, 
we rescale it to maintain numerical stability during policy updates.

\begin{figure}[H]
\centering
  \begin{subfigure}{0.48\textwidth}
    \centering
    \includegraphics[width=\textwidth]{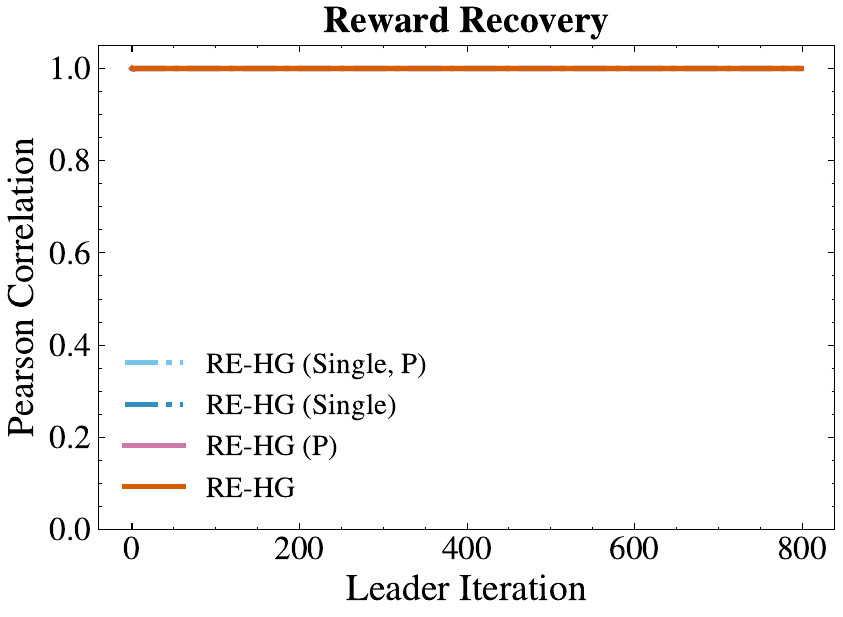}
    \caption{Reward Correlation}
    \label{fig:cp2_reward_pearson}
  \end{subfigure}
  \hfill 
  \begin{subfigure}{0.48\textwidth}
    \centering
    \includegraphics[width=\textwidth]{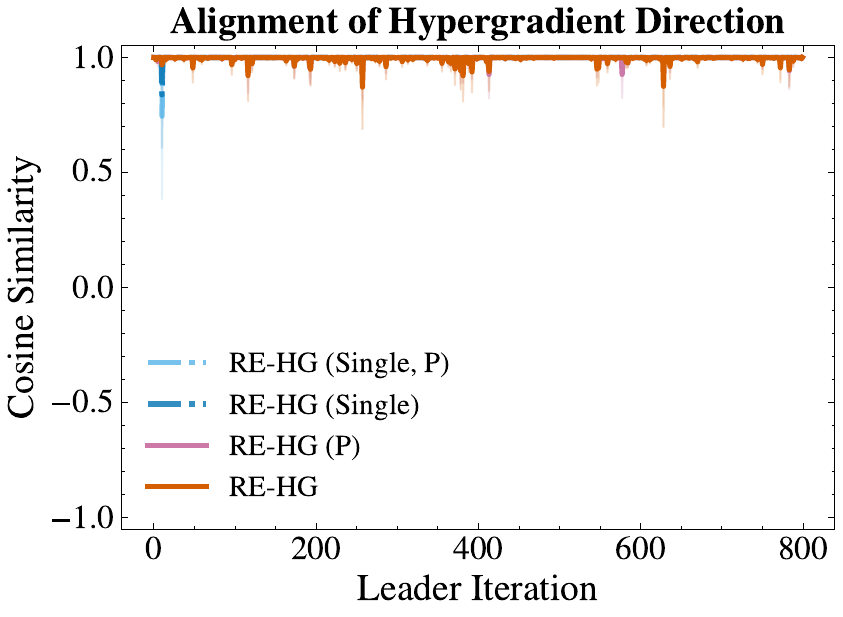}
    \caption{HG Cosine Similarity}
    \label{fig:cphg_cosine_similarity}
  \end{subfigure}
  \caption{(a) Pearson correlation between the recovered and true rewards, weighted by the undiscounted state-action occupancy measure of the surrogate follower policy. (b) Cosine similarity of the HG between the estimated and \oracle.}  \label{fig:smcp}
\end{figure}

\begin{figure}[H]
    \centering
    \includegraphics[width=0.6\textwidth]{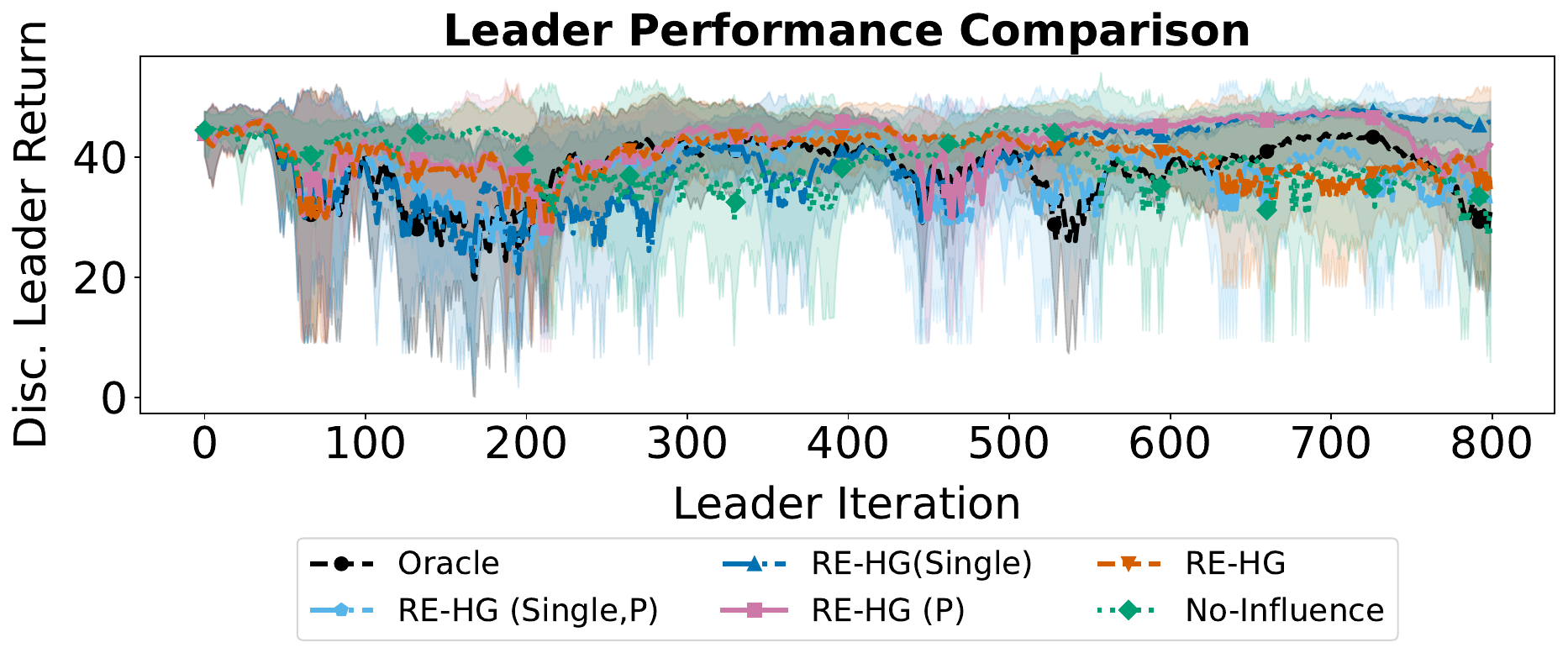}
    \caption{This figure illustrates the instability caused by calculating the benefit using the online network for both functions. Specifically, the benefit is calculated as the difference between the online leader's action-value function ($Q_L^{\text{online}}$) and the online leader's value function ($V_L^{\text{online}}$). For each run, performance is evaluated as the mean of 10 test rollouts. Avg.\ (line) and std.\ (shaded area) across the 3 independent training runs are displayed. ``P'' indicates the known transition probability scenario.}
    \label{fig:oleader_performance}
\end{figure}

\paragraph{Result}
As discussed in \Cref{sec:2mg-continuous}, we present the detailed evaluation metrics for the continuous 2-player Markov Game task. \Cref{fig:cp2_reward_pearson} shows the highly stable weighted Pearson correlation between the recovered and true rewards. \Cref{fig:cphg_cosine_similarity} illustrates the cosine similarity of the hypergradient direction against \oracle. As shown in \Cref{fig:oleader_performance}, calculating the benefit using the online network for both functions (i.e., $Q_L^{\text{online}} - V_L^{\text{online}}$) fails to achieve stability. This instability arises because, near the optimal policy, the influence term practically vanishes. While $Q_L$ learns and generalizes at this point, the leader's policy continues to change due to other gradient terms. Consequently, the previously generalized $Q_L$ values become inaccurate for the shifted policy, leading to performance degradation.

\paragraph{Hyperparameters} 
Hyperparameters used in this experiment are summarized in \Cref{tab:algorithm_hyperparams}.

\begin{table}[H]
\centering
\caption{Hyperparameters}
\label{tab:algorithm_hyperparams}
\begin{tabular}{llc}
\toprule
\textbf{Name} & \textbf{Symbol} & \textbf{Value} \\
\midrule
Time step & $\Delta t$ & $0.1$ \\
Max steps per episode  & $T$ & $100$ \\
Leader iterations & $K$ & $800$ \\
Episodes per iteration & - & $4$ \\
Leader discount factor & $\gamma_L$ & $0.99$ \\
Follower discount factor & $\gamma_F$ & $0.99$ \\
Leader learning rate & $\alpha_L$ & $0.05$ \\
Temperature & $\beta$ & $0.01$ \\
Leader Critic Update times & $-$ & $5$ \\
Leader sigma & $\Sigma_L$ & $1.0 \times 10 ^{-3}$ \\
threshold  & $\epsilon$ & $10^{-4}$ \\
\midrule
Pre-training episodes & $n_{\text{pre}}$ & $40$ \\
Pre-training epochs & - & $50$ \\
\bottomrule
\end{tabular}
\end{table}

\subsection{Comparing Oracle}\label{apdx:oracle-ablation}
 
In the Hard Setting of \Cref{apdx:hard-setting}, \rehg attains a higher leader
return than \oracle. Since \oracle computes the exact hypergradient, the gap
may simply reflect the smaller effective update of \rehg. We test this by
retuning the \oracle update in three ways: (a) a tuned learning rate,
(b) gradient clipping, and (c) a scaled influence term. Modification (c) is the
closest \oracle-side analogue of the regularization in \rehg, since the
eigenvalue truncation in \Cref{alg:bilevel_reward_identification} has no
counterpart when the true reward is available.
 
\begin{figure}[H]
    \centering
    \includegraphics[width=0.95\textwidth]{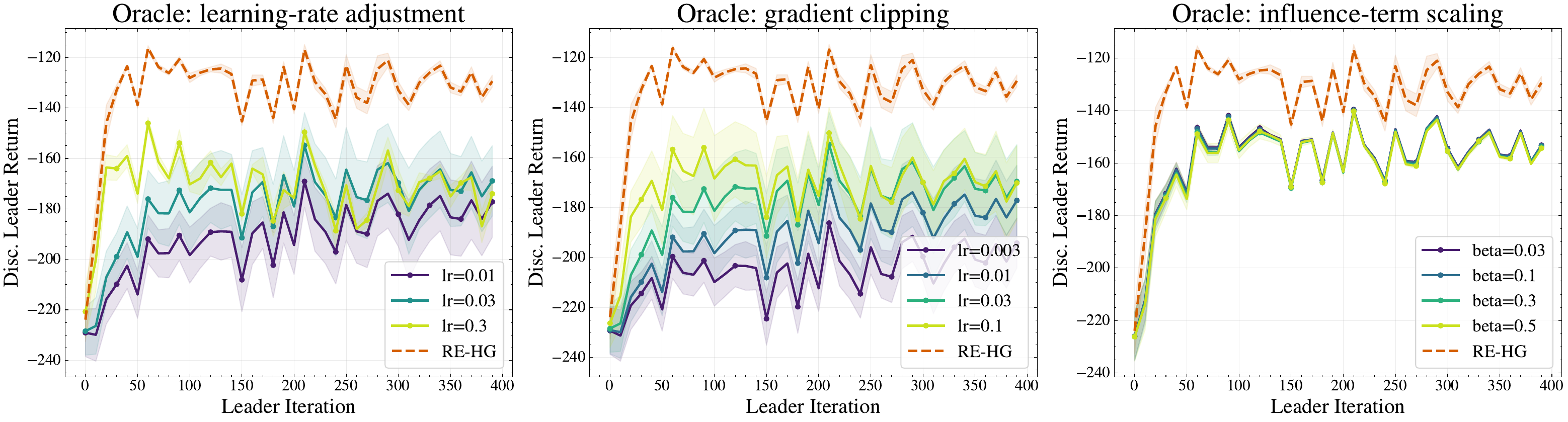}
    \caption{Learning curves for three \oracle-side modifications. Lines
    represent the mean discounted leader return across five random seeds, and
    shaded regions represent one standard deviation.
    \textbf{(a) Tuned learning rate:} the \oracle learning rate is varied.
    \textbf{(b) Gradient clipping:} the hypergradient is clipped and the
    learning rate is varied.
    \textbf{(c) Influence-term scaling:} the influence term is multiplied by a
    global factor while the direct term is left unchanged.
    The dashed \rehg curve is included as the baseline in each panel.}
    \label{fig:oracle_comparison}
\end{figure}

All ablations use the continuous Conf-MDP environment with the same base
configuration as the Hard Setting of \Cref{apdx:hard-setting}, and each
\oracle and \rehg variant is evaluated over five random seeds.
\Cref{fig:oracle_comparison} shows the learning curves.
 
Tuning the learning rate does not help. The \oracle uses $\alpha_L = 0.1$ in
the Hard Setting, and \Cref{fig:oracle_comparison}(a) shows that none of the
alternatives $0.01$, $0.03$, and $0.3$ improved on it, the best being
$-170.54 \pm 14.07$. Clipping the hypergradient as $g \leftarrow g/\norm{g}$
fares no better across learning rates $0.003$, $0.01$, $0.03$, and $0.1$, the
last of which is the original value, with a best of $-171.12 \pm 14.73$ in
\Cref{fig:oracle_comparison}(b).
 
Scaling the influence term, the second term of the hypergradient, by factors
between $0.03$ and $0.5$ while leaving the direct term unchanged does help.
\Cref{fig:oracle_comparison}(c) shows it raises the \oracle return to between
$-155.28 \pm 0.72$ and $-154.15 \pm 0.00$, still far below the
$-129.59 \pm 0.88$ of \rehg.
 
Rescaling the update therefore does not reproduce the effect of eigenvalue
truncation. Modifications (a) and (b) change only the magnitude of the update,
and (c) changes the ratio between the direct and influence terms by a single
global factor. Eigenvalue truncation instead suppresses poorly excited
directions of the influence term selectively, so it changes the direction of
the update rather than its scale.

\subsection{Multi-environment IRL}\label{apdx:multi-env-irl}
 
A natural alternative to the sequential aggregation of
\Cref{alg:bilevel_reward_identification} is to estimate one shared reward
jointly from the demonstrations accumulated across all environments visited so
far. We ask whether sequential aggregation gives anything up.
 
\begin{figure}[H]
    \centering
    \includegraphics[width=0.75\textwidth]{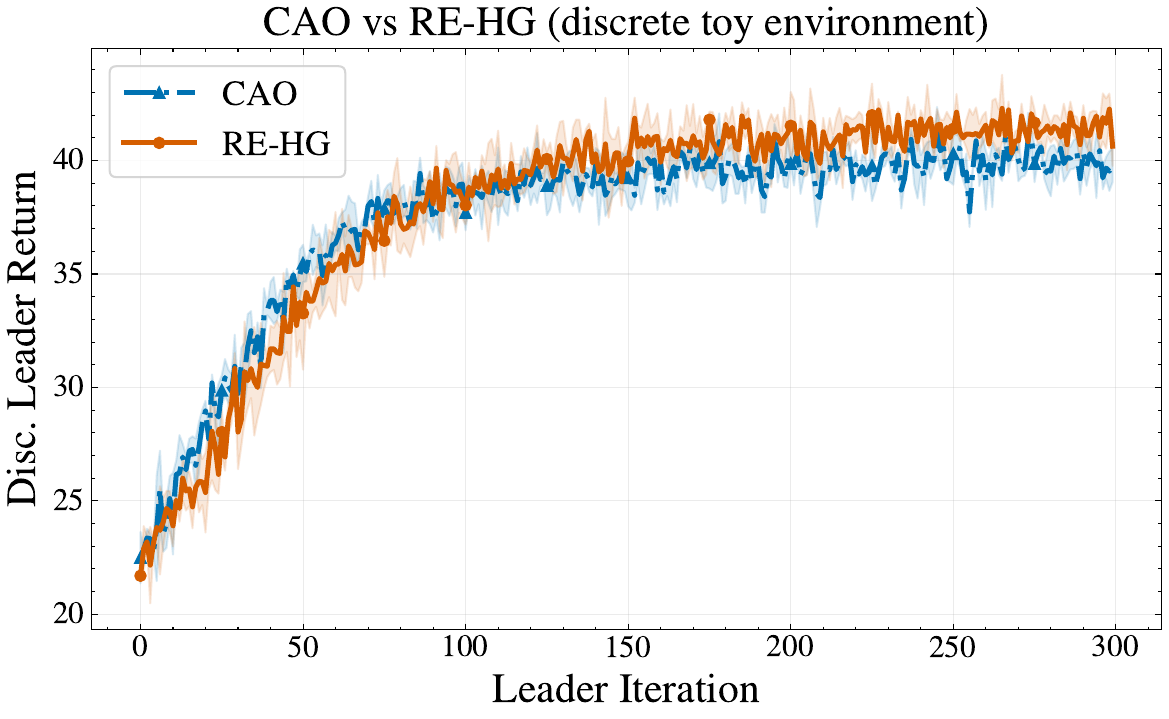}
    \caption{Comparison between the joint multi-environment IRL baseline
    (CAO) and \rehg in the discrete toy environment with known transition
    dynamics. Lines represent the mean discounted leader return across three
    random seeds, and shaded regions represent one standard deviation.}
    \label{fig:multi_environment_irl}
\end{figure}

We implemented a joint multi-environment IRL baseline from Section~7.2 of
\citet{Cao2021neurips}. That formulation assumes access to the transition
dynamics, so we evaluated both methods in a discrete environment with known
dynamics. \Cref{fig:multi_environment_irl} shows the learning curves.
 
The two methods match throughout training and both converge close to the
optimum, reaching discounted leader returns of $21.46$ and $20.75$
respectively, averaged over the last 30 leader iterations with the same three
random seeds. Sequential aggregation is therefore no better in final
performance, which is what our analysis predicts: both approaches resolve
reward ambiguity under their respective idealized assumptions. Its advantages
are algorithmic, and are threefold.
 
First, \rehg integrates reward estimation into the standard bilevel
optimization procedure. Each leader update presents the follower with a new
environment, and \rehg re-estimates the reward from trajectories collected in
that environment and immediately computes the hypergradient. Joint
multi-environment IRL instead estimates a shared reward from demonstrations
accumulated across environments, which requires storing past trajectories and
deciding how many of them to keep.
 
Second, the formulation of \citet{Cao2021neurips} assumes known transition
dynamics, whereas \rehg also applies to the two-player Markov game setting of
this paper, where they are unknown.
 
Third, \Cref{thm:identifiability} and \Cref{prop:bound} are developed for
sequential reward aggregation and give identifiability together with an error
bound relating the reward-estimation error to the hypergradient error. We are
not aware of a joint multi-environment IRL method that applies directly to our
setting and offers a comparable guarantee.

\subsection{Misspecified reward features}\label{apdx:misspec}
 
\begin{figure}[H]
    \centering
    \includegraphics[width=0.95\textwidth]{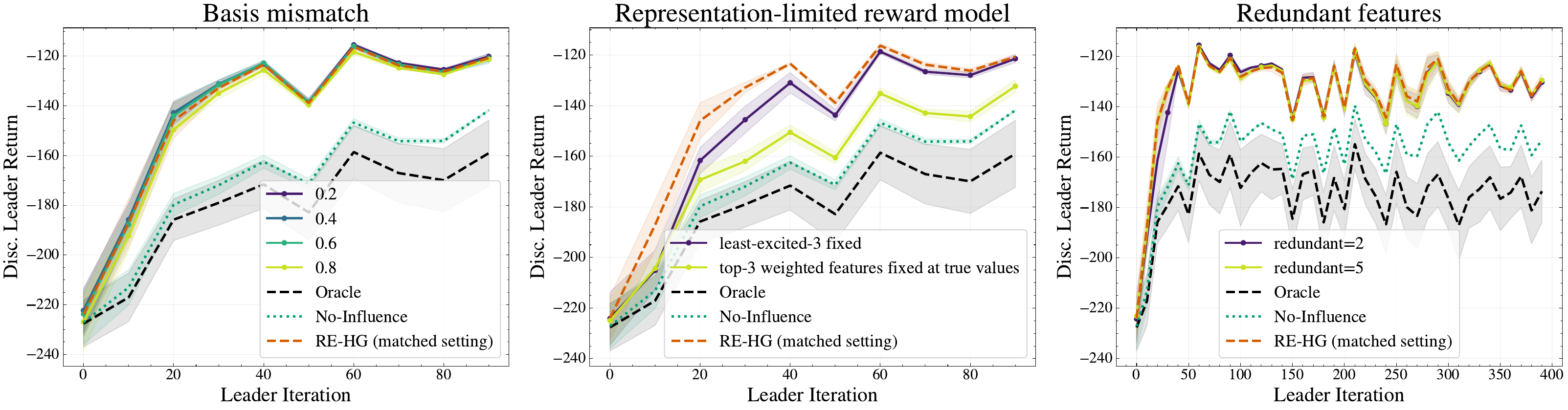}
    \caption{
    Learning curves under reward-feature misspecification in the continuous Conf-MDP.
    Lines show mean discounted leader return and shaded regions show one standard deviation.
    The misspecified \rehg{} runs use three random seeds, while the Oracle,
    No-Influence, and matched-setting \rehg{} reference curves use five seeds.
    \textbf{(a) Basis mismatch:} the true reward contains off-diagonal interaction
    terms such as $s_i s_j$, whereas the IRL model is restricted to diagonal
    quadratic features.
    \textbf{(b) Limited update capacity and representation misspecification:} the three state-cost features with the largest true weights, accounting for approximately $80.4\%$ of the total state-cost weight, are fixed at their true values to isolate an update-capacity restriction. A complementary curve fixes the three least-excited features at incorrect values, introducing genuine representation misspecification.
    \textbf{(c) Redundant-feature over-specification:} two or five cross-term
    features are added to the IRL model even though their true coefficients are
    zero.
    The \rehg{} reference curve is included in every panel.
    }
    \label{fig:misspecified_reward_features}
\end{figure}

\Cref{asm:linear} assumes a known feature extractor. One-hot features make this
harmless in discrete environments, but in continuous ones the recovered reward
is only as good as the chosen basis, so we test three forms of
misspecification: basis mismatch, limited update capacity, and redundant
features. All experiments use the continuous Conf-MDP (LQR) environment with
the same base configuration as the Hard Setting of \Cref{apdx:hard-setting};
each condition is evaluated over five random seeds. \Cref{fig:misspecified_reward_features} shows the learning curves.
 
Under basis mismatch, \Cref{fig:misspecified_reward_features}(a) shows that the
exact analytic \oracle becomes unstable because its HG is
unregularized, yielding leader returns between $-171.38 \pm 10.72$ and
$-155.81 \pm 0.00$. \rehg stays between $-128.40 \pm 1.68$ and
$-129.45 \pm 2.33$, consistent with the smoothing effect of eigenvalue
truncation.
 
\Cref{fig:misspecified_reward_features}(b) separates a restriction on update
capacity from genuine representation error. Freezing the three state-cost
features with the largest true weights, which account for
$46/(0.1+0.1+8+30+8+3+5+3) \approx 80.4\%$ of the total state-cost weight, at
their true values introduces no representation error and isolates the effect of
restricting which directions can be updated; \rehg still reaches
$-141.43 \pm 2.50$, well above the \noinf baseline of $-153.21$. Freezing the
three least-excited features at incorrect values instead, $[1,1,1]$ against
true values $[8,3,8]$, costs almost nothing: \rehg reaches
$-129.94 \pm 1.94$, because eigenvalue truncation already suppresses these
directions.
 
With redundant features, \Cref{fig:misspecified_reward_features}(c) shows that
\rehg again beats the \noinf baseline, reaching $-131.84 \pm 1.60$ with two
redundant features and $-130.92 \pm 0.99$ with five.

\subsection{Optimal follower assumption}\label{apdx:suboptimal-follower}
 
\Cref{prop:irl} characterizes the IRL solution only when the follower plays its
exact best response, and a real follower is at best approximately optimal. We
therefore test two suboptimal followers: a myopic one that runs a limited
number of Riccati iterations, $T \in \{12,16,20,30\}$, and an under-reactive
one whose feedback gain is scaled as $(1-\alpha)K^*$ with
$\alpha \in \{0.05,0.1,0.2,0.35\}$. Smaller $T$ and larger $\alpha$ mean
larger deviations from the optimal follower.
 
\begin{figure}[H]
    \centering
    \includegraphics[width=0.95\textwidth]{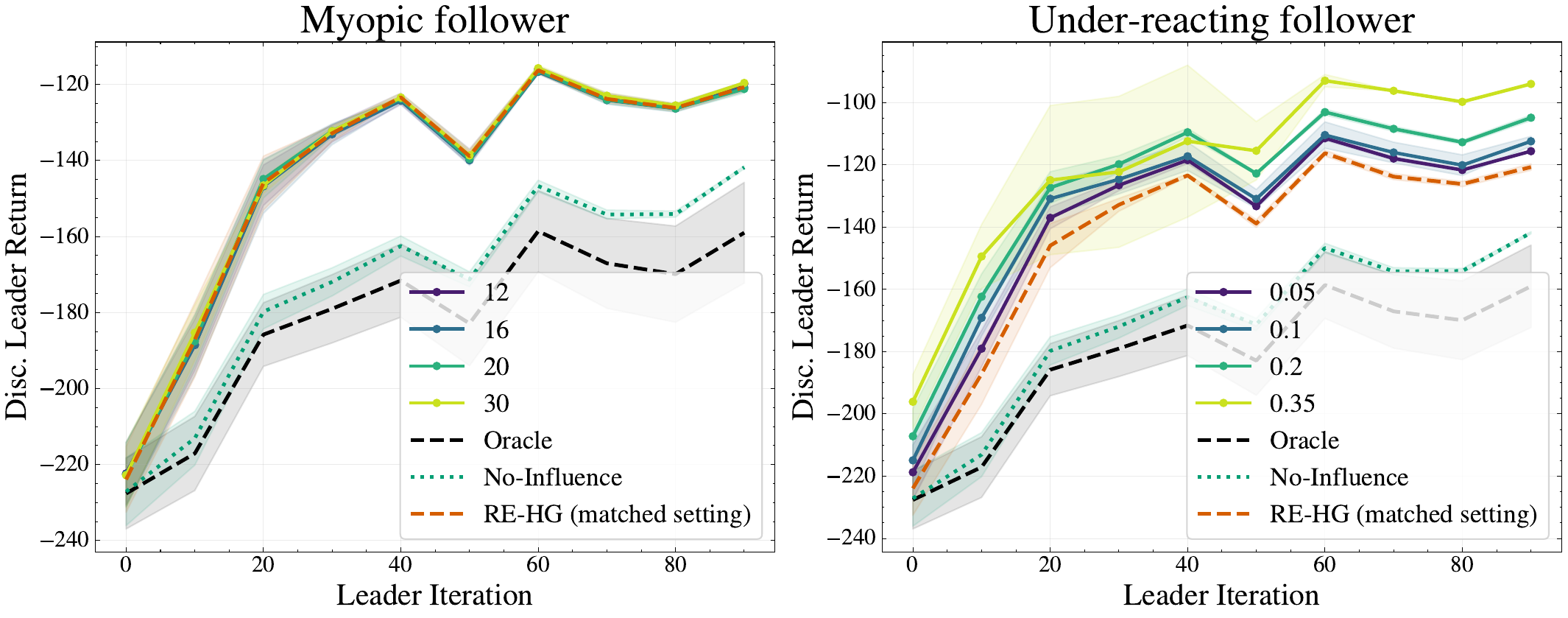}
    \caption{Learning curves under suboptimal follower models in the continuous
    Conf-MDP. Lines represent mean discounted leader return across three random
    seeds (five for the \rehg baseline), and shaded regions represent one
    standard deviation.
    \textbf{(a) Myopic follower:} varying Riccati iteration budget $T$.
    \textbf{(b) Under-reactive follower:} varying gain reduction parameter
    $\alpha$.
    The \rehg curve is included as the baseline in each panel.}
    \label{fig:suboptimal_follower}
\end{figure}

\begin{figure}[H]
    \centering
    \includegraphics[width=0.95\textwidth]{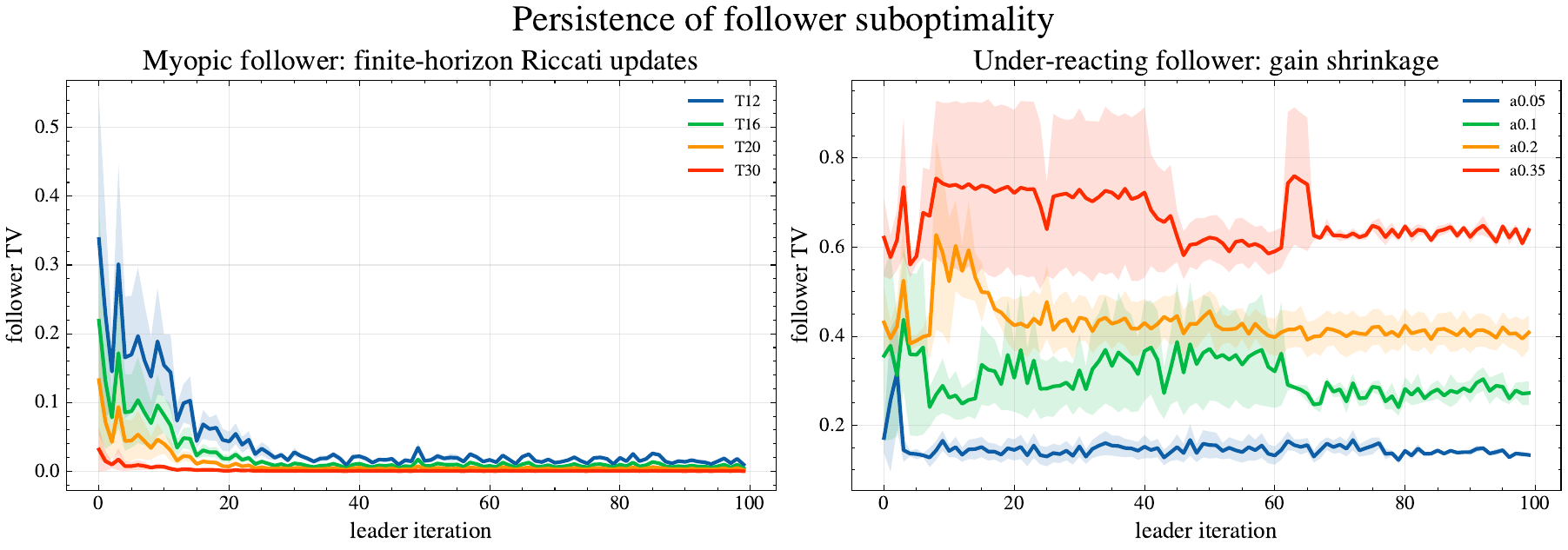}
    \caption{Total-variation distance between the suboptimal follower and the
    exact best response over the course of training, for the same two follower
    models and in the same panel order as \Cref{fig:suboptimal_follower}.
    \textbf{(a) Myopic follower.} \textbf{(b) Under-reactive follower.}
    Lines represent the mean across three random seeds, and shaded regions
    represent one standard deviation.}
    \label{fig:suboptimal_follower_tv}
\end{figure}

Both experiments use the continuous Conf-MDP (LQR) environment with the same
base configuration as the Hard Setting of \Cref{apdx:hard-setting}; each
condition is evaluated over three random seeds and the \rehg baseline over
five. \Cref{fig:suboptimal_follower} shows the learning curves and
\Cref{fig:suboptimal_follower_tv} the total-variation (TV) distance between the
suboptimal follower and the exact best response during training.
 
The myopic follower deviates only transiently. In
\Cref{fig:suboptimal_follower_tv}(a) the initial TV distance is about $0.34$
for $T=12$ but the average over training is $0.0443 \pm 0.0153$, and
\Cref{fig:suboptimal_follower}(a) shows the leader return is essentially
unaffected: \rehg reaches $-127.55 \pm 0.46$ for $T=30$ and
$-127.91 \pm 0.87$ for $T=12$.
 
The under-reactive follower deviates persistently, with TV distances between
$0.15$ and $0.66$ throughout training in
\Cref{fig:suboptimal_follower_tv}(b). Here \rehg adapts its reward estimate to
the observed trajectories and, as \Cref{fig:suboptimal_follower}(b) shows,
overtakes the \oracle: $-123.25 \pm 0.42$ for $\alpha=0.05$ and
$-99.74 \pm 0.17$ for $\alpha=0.35$, against $-154.28 \pm 2.35$ to
$-152.36 \pm 0.79$ for the \oracle across the tested under-reaction levels. The
reward correlations stay between $0.70$ and $0.82$. The comparison is not
symmetric, however: the \oracle uses the true reward but still assumes an exact
best response, which is violated here by construction.
 
Taken together, \rehg tolerates both transient and persistent follower
suboptimality by fitting its reward estimate to the behavior actually observed.

\section{Broader Impacts}\label{apdx:broader}
This research can positively impact the efficiency of real-world hierarchical systems, such as smart grids and traffic control, where the follower's reward is unknown. However, if applied to human-centric domains like recommendation systems, inferring unobserved rewards from user behavior could lead to unintended manipulation or privacy concerns. To mitigate these risks, such applications require transparency regarding data usage and strict safety constraints on the leader's action space.

\section{Compute Resources}\label{apdx:compute}
All experiments were conducted on a workstation equipped with dual Intel(R) Xeon(R) Gold 6348 CPUs @ 2.60GHz (totaling 56 physical cores and 112 logical threads) and 512 GB of RAM. The entire suite of experimental evaluations, including all baseline comparisons and ablations, was completed within 12 hours of total wall-clock time.

\end{document}